\documentclass{article}
\usepackage{arxiv}

\usepackage[normalem]{ulem} 

\usepackage[utf8]{inputenc} 
\usepackage[T1]{fontenc}    
\usepackage{hyperref}       
\usepackage{url}            
\usepackage{booktabs}       
\usepackage{amsfonts}       
\usepackage{microtype}      
\usepackage{color}
\usepackage{lipsum}
\usepackage{amsthm}
\usepackage{indentfirst}
\usepackage{graphicx}
\usepackage{epsfig,epsf,latexsym,subfigure}
\usepackage{float}
\usepackage{epstopdf}
\usepackage{fourier}
\usepackage{bm}
\usepackage{bbm} 
\usepackage{mathtools}
\usepackage{latexsym,enumerate}
\usepackage{tabularx}
\usepackage{capt-of}
\usepackage{cases}
\usepackage{enumitem} 
\usepackage{algorithm}
\usepackage{algpseudocode}
\usepackage{fourier}
\usepackage[usenames,dvipsnames]{xcolor}
\usepackage{mathrsfs}
\usepackage{subcaption}
\usepackage{changepage}
\usepackage{todonotes}

\DeclareMathAlphabet{\mathcal}{OMS}{cmsy}{m}{n}
\SetMathAlphabet{\mathcal}{bold}{OMS}{cmsy}{b}{n}

\def\bm{\boldsymbol}

\newcommand{\comment}[1]{}

\newcommand{\BEA}{\begin{eqnarray}}
\newcommand{\EEA}{\end{eqnarray}}

\newcommand{\td}{\text{d}}

\newcommand{\mbv}{\mathbf{v}}

\newcommand{\BR}{\mathbb{R}}

\newcommand{\mbX}{\mathbf{X}}
\newcommand{\mbY}{\mathbf{Y}}
\newcommand{\mbf}{\mathbf{f}}

\newcommand{\mbx}{\mathbf{x}}
\newcommand{\mby}{\mathbf{y}}
\newcommand{\mbW}{\mathbf{W}}
\newcommand{\mbF}{\mathbf{F}}

\newcommand{\mbu}{\mathbf{u}}
\newcommand{\mbK}{\mathbf{K}}
\newcommand{\mbV}{\mathbf{V}}
\newcommand{\mbA}{\mathbf{A}}
\newcommand{\mbP}{\mathbf{P}}
\newcommand{\mbH}{\mathbf{H}}

\newcommand{\mbG}{\mathbf{G}}
\newcommand{\mbS}{\mathbf{S}}
\newcommand{\mbE}{\mathbf{E}}
\newcommand{\mbD}{\mathbf{D}}
\newcommand{\mbJ}{\mathbf{J}}

\newcommand{\mbU}{\mathbf{U}}
\newcommand{\mbC}{\mathbf{C}}
\newcommand{\mbI}{\mathbf{I}}
\newcommand{\mbR}{\mathbf{R}}

\title{Learning dynamical systems from noisy data with Weak-form Kernel Ridge Regression}

\author{
  Max Kreider \\
  Department of Mathematics \\
  The Pennsylvania State University, University Park, PA 16802, USA\\
  \texttt{mbk6295@psu.edu} \\
  \And
 John Harlim \\
  Department of Mathematics, Institute for Computational and Data Sciences \\
  The Pennsylvania State University, University Park, PA 16802, USA\\
  \texttt{jharlim@psu.edu} \\
  \And
 Daning Huang\\
  Department of Aerospace Engineering \\
  The Pennsylvania State University, University Park, PA 16802, USA\\
  \texttt{daning@psu.edu}
}

\begin{document}

\maketitle

\begin{abstract}

Accurate prediction of complex dynamical systems from noisy measurements remains a significant challenge in scientific computing.
Kernel ridge regression learning strategies are often effective when applied to clean data, but have limited success with noisy data.
Recent work has observed that a weak formulation can act to filter noisy data, and different learning strategies have achieved increased noise robustness with a weak-form framework.
In this manuscript, we give an overview of the filtering mechanism behind the weak formulation and provide a bias-variance error decomposition.
Using these insights, we combine a weak formulation with a kernel learning strategy to propose Weak-form Kernel Ridge Regression (WKRR) for learning dynamical systems.
The proposed framework is simple to implement, effective for both clean and noisy data, and outperforms several baseline methods.
We demonstrate the performance of WKRR on chaotic benchmark systems in up to 64 dimensions, as well as 15,000-dimensional real-world fluid data.

\end{abstract}

\section{Introduction}\label{sec: introduction}

Many problems in scientific computing and engineering disciplines involve modeling and prediction of dynamical systems, with applications including weather \cite{christensen2019reliable, hussain2018dynamic, nino2021data}, environmental and ecological science \cite{gimenez2024theoretical, luo2011ecological, song2014application, ye2015equation}, biology \cite{gilpin2020learning, prokop2025data, xing2022reconstructing}, fluid dynamics \cite{agostini2020exploration, erge2022combining, long2018hybridnet, schmidt2026data}, finance \cite{castillo1995intelligent, waheed2026data}, and traffic \cite{antoniou2013dynamic, avila2020data, xu2016big}.
However, many physically relevant problems remain challenging due to high-dimensionality, complex or chaotic dynamics, lack of known underlying dynamics, and noisy or low-fidelity observational data.

Purely data-driven methods have emerged as a strong option for learning dynamical systems \cite{brunton2022data, ghadami2022data, north2023review, wang2005gaussian, williams2015data}.
Such methods circumvent the need to form dynamical equations and typically seek to represent unknown dynamics with a high-fidelity reduced-order model.
A variety of popular approaches have proven to be competitive in this context.
Dynamic mode decomposition (DMD) and variants leverage a Koopman framework that lifts finite-dimensional nonlinear data to an infinite-dimensional linear representation, often leading to a simplified low-dimensional surrogate  model \cite{colbrook2023mpedmd, colbrook2023residual, kutz2016dynamic, li2017extended, mezic2022numerical, williams2015data}. 
Sparse identification of nonlinear dynamical systems (SINDy) and variants discover dynamical equations from data by choosing a suitable, often sparse, linear combination of dictionary functions \cite{brunton2016discovering, brunton2016sparse, kaheman2020sindy, zhang2019convergence}.
Neural ordinary differential equations (NODEs) learning underlying dynamics by training a neural network to represent a continuous-time vector field \cite{chen2018neural, goyal2022neural, li2025weak, oh2025comprehensive, Yu2024gkbf, zhao2025accelerating}.
Various machine learning techniques such as Long Short-Term Memory \cite{hochreiter1997long, lindemann2021survey, vlachas2018data, yu2019review}, reservoir computing \cite{gauthier2021next, nakajima2021reservoir, tanaka2019recent, yan2024emerging}, and autoencoders \cite{floryan2022data} have also gained prominence.
Kernel-based approaches such as kernel ridge regression (KRR) mitigate the curse of dimensionality with the so-called ``kernel trick'' \cite{ahmed2022kernel, ali2020complete, exterkate2016nonlinear, song2025learning, vovk2013kernel}.
KRR is especially attractive because it does not require a dictionary of functions, and is straightforward to implement.
Recent work has shown that KRR outperforms multiple baseline methods in data-driven dynamical system learning and forecasting problems over a wide range of data sets \cite{song2025learning}.
While these approaches typically perform well when applied to clean data, their performance often degrades significantly in the presence of measurement or observational noise, especially when the underlying dynamics are chaotic.

Several methods have been proposed to mitigate lack of robustness in the presence of noise, including data assimilation and filtering methods \cite{cheng2023machine,gottwald2021combining, gottwald2021supervised}, and Gaussian process approaches \cite{girard2002gaussian, yan2009gaussian, yang2021inference}.
The approach developed in \cite{gottwald2021combining}, coined RAFDA, has proven to be competitive for low-dimensional systems, but does not scale well to high-dimensions.
A different approach is to combine a learning strategy with a weak formulation, which involves integrating data residuals over a family of test functions.
Classical (strong) approaches enforce pointwise consistency between given observational data, and typically perform poorly for noisy data because pointwise errors are magnified by erroneous fluctuations.
In contrast, weak approaches relax pointwise loss strategies in favor of orthogonality constraints between residuals and test functions.
Loss functions involving weak formulations have been observed to perform more robustly in the presence of noise, and have been incorporated successfully into both the SINDy framework \cite{bortz2024weak,messenger2021weak, messenger2025asymptotic, messenger2024weak} and the NODE framework \cite{li2025weak}.
Both \cite{li2025weak} and \cite{messenger2021weak} point out that a weak formulation acts as a filter for noisy data, suggesting a mechanism for its observed noise robustness.

In the present manuscript, we will build on these previous insights to show explicitly that the weak formulation acts to filter noisy data.
We will consider a family of test functions that arise from uniform translations of a generating function, and will interpret the resulting weak formulation as an orthogonal projection procedure which filters by projecting noisy data onto the subspace spanned by these test functions.
Moreover, we provide a brief bias-variance error decomposition of signal filtering via the weak form.
In particular, we provide an exact expression for the variance, i.e., the error that arises due to noise corruption.
This interpretation connects to well-known results in information theory and signal processing \cite{unser2002sampling,unser2002general, unser2002polynomial, unser2002generalized, wyner1998introduction}.
Shannon's sampling theorem, which states that an ideal band-limited function can be perfectly reconstructed with suitably sampled data, may be interpreted as an orthogonal projection onto a subspace of band-limited functions \cite{shannon1949communication}.
Subsequent work has extended this orthogonal projection analysis to classes of functions arising from integer shifts of generating functions \cite{unser2002sampling}.
While the error analysis that we provide is not novel, we include it to justify the use of a weak formulation and to clarify the mechanism by which it provides noise robustness.

Motivated by the simplicity and recent success of kernel-based learning methods and the noise robustness of weak formulations, we propose Weak-form Kernel Ridge Regression (WKRR) as a noise robust, data-driven learning framework.
We remark that WKRR does not require a suite of dictionary functions, in contrast to Weak SINDy approaches \cite{bortz2024weak,messenger2021weak, messenger2025asymptotic, messenger2024weak} that critically need the underlying functions or vector fields to be spanned by the dictionary functions.
Moreover, WKRR does not require extensive hyperparameter tuning as many machine learning frameworks, such as NODE, require \cite{li2025weak}.
We will show numerically that the forecasting horizon of WKRR is similar to that of a strong KRR formulation, but much less computationally expensive.
We will also demonstrate the effectiveness of WKRR with different choices of kernel function, including the standard Gaussian kernel and the Diffusion Maps (DM) kernel, which has recently received attention for its excellent performance across a wide range of chaotic and experimental datasets \cite{song2025learning}.
The success of WKRR with various kernel functions broadens the method's applicability and suggests that the practitioner may select kernel functions that favor speed or accuracy as the situation warrants.

The remainder of the manuscript is organized as follows.
In \S\ref{sec: preliminaries}, we review the classical KRR method and the Gaussian and DM kernel functions.
In \S\ref{sec: explaining the weak form}, we review the weak formulation, explain how it acts as a filter, and provide a brief bias-variance error decomposition for the filtering procedure.
The main contribution of this work is provided in \S\ref{sec: proposed approach}, where we propose WKRR as a noise-robust learning method for dynamical systems.
We propose a validation procedure to select appropriate model hyperparameters, and summarize the steps needed to implement WKRR in practice.
We apply WKRR to several numerical examples in 
\S\ref{sec: examples}, including chaotic baseline systems in up to 64 dimensions, and real-world turbulent fluid data made available by \cite{schmidt2026data} as part of a Community Challenge.
We conclude with a brief discussion in \S\ref{sec: discussion}.

\section{Kernel Ridge Regression Review}\label{sec: preliminaries}

In this section, we review the classical kernel ridge regression (KRR) framework for learning solution operators of dynamical systems.
We will refer to this approach as the ``strong approach'' throughout the manuscript.

\subsection{Kernel Ridge Regression}\label{subsec: KRR}

Suppose we are given noisy data $\mbu(t_i)\equiv\mbu_i = (u_i^{(1)},\dots,u_i^{(n)})\in \mathbb{R}^n$, $i=1,\dots,N$, sampled at times $t_i = (i-1)\cdot \Delta t$.
We assume that the data is of the form 
\begin{equation}\label{eq: data}
    \mbu_i = \mbx_i + \sigma \bm\xi_i,
\end{equation}
where $\mbx_i$ denotes clean data generated by an autonomous dynamical system of the form $\mbx'= \mbf(\mbx)$, and $\bm\xi_i$ is i.i.d.~Gaussian with zero mean and covariance matrix $\bm\Sigma = \operatorname{diag}(\eta_1^2,\dots,\eta_n^2)$.
We define $\eta_\ell$ to be the root-mean-square (RMS) of the $\ell$th component of the clean signal
\begin{equation}
    \eta_\ell = \sqrt{\frac{1}{N} \sum_{i=1}^N \left|x_i^{(\ell)}\right|^2}.
\end{equation}
The parameter $\sigma \geq 0$ represents the signal-to-noise ratio and will be varied in our numerical experiments.

In this section, we assume that $\sigma = 0$ so that the given data is not corrupted by noise.
Let $\mathcal{M}\subset \mathbb{R}^n$ denote the forward invariant set of the underlying system.
Define the flow map $\bm{\Phi}_\tau = (\Phi_{\tau}^{(1)},\dots,\Phi_{\tau}^{(n)}): \mathbb{R}^n \to \mathcal{M}$, which advances the system state over a time increment $\tau>0$, i.e., $\bm{\Phi}_\tau(\mbx(t)) = \mbx(t+\tau)$.

We will use KRR to learn solution operators with either a \textit{direct-connection} or a \textit{skip-connection} scheme.
Both frameworks are useful in practice; direct-connection often outperforms skip-connection when the underlying dynamics are stiff.
Otherwise, the skip-connection scheme typically yields superior performance \cite{song2025learning}.
We employ both frameworks in our numerical examples.
We now describe each learning framework in turn.

\subsubsection{Direct-Connection Scheme}

Let $k(\cdot,\cdot;\epsilon):\mathbb{R}^n\times \mathbb{R}^n \to \mathbb{R}$ denote a scalar-valued kernel function, where $\epsilon >0$ denotes a tunable bandwidth parameter. 
The key idea of the direct-connection scheme is to approximate each component of the flow map directly,
\begin{equation}\label{eq: KRR approx}
    \Phi^{(\ell)}_\tau(\mbx) \approx \sum_{i=1}^N k(\mbx,\mbu_i;\epsilon)\alpha_i^{(\ell)},
\end{equation}
for reconstruction coefficients $\{\alpha_{i}^{(\ell)}\}_{\ell=1,\ldots, n}$ to be determined.
To proceed, we collect the available data in snapshot pairs $S=(\mbX,\mbY)$.
The input matrix $\mbX\in \mathbb{R}^{ (N-1)\times n}$ is defined by
\begin{equation}\label{eq: X matrix clean}
    \mbX = \begin{bmatrix}
        \mbu_1 & \mbu_2 & \dots & \mbu_{N-1}
    \end{bmatrix}^\top,
\end{equation}
while the output matrix $\mbY \in \mathbb{R}^{(N-1)\times n}$ is defined by
\begin{equation}\label{eq: direct-connection matrix}
        \mbY = \begin{bmatrix}
            \mbu_2 & \mbu_3 & \dots & \mbu_{N}
        \end{bmatrix}^\top.
    \end{equation}

    Let $K(\cdot,\mbX;\epsilon) = [k(\cdot,\mbu_1;\epsilon),\dots,k(\cdot,\mbu_{N-1};\epsilon)]:\mathbb{R}^n \to \mathbb{R}^{N-1}$ be the kernel function evaluated over the input data $\mbX$.
    In this notation, our goal is to model the flow map
\begin{equation}\label{eq: kernel direct-connection}
    \mbu_{i+1} = \bm\Phi_{\Delta t}(\mbu_i) \approx K(\mbu_i,\mbX;\epsilon)\bm\alpha,
\end{equation}
where $\bm\alpha\in \mathbb{R}^{(N-1)\times n}$ collects the reconstruction coefficients.
With the goal of expressing \eqref{eq: kernel direct-connection} in matrix notation, define the Gram matrix $\mbK(\epsilon)\in \mathbb{R}^{(N-1)\times (N-1)}$ whose $(i,j)$ entry is $k(\mbu_i,\mbu_j;\epsilon)$ for $i,j=1,\dots,N-1$.
We then arrive at a linear least squares problem which enforces pointwise consistency between the snapshot data pairs
\begin{equation}\label{eq: least squares problem}
    \mbY = \mbK(\epsilon) \bm\alpha.
\end{equation}
We remark that \eqref{eq: least squares problem} is often ill-conditioned.
In practice, a solution $\hat{\bm\alpha}$ is determined by solving the related KRR problem
\begin{equation}\label{eq: KRR problem}
    \hat{\bm{\alpha}} = \arg\min_{\bm\alpha} \left(\|\mbK(\epsilon)\bm\alpha - \mbY\|_F^2 + \lambda \bm\alpha^\top \mbK(\epsilon)\bm\alpha \right),
\end{equation}
where $\lambda$ is a user-specified regularization parameter.
The problem \eqref{eq: KRR problem} admits a unique solution
\begin{equation}
    \hat{\bm{\alpha}} = \left(\mbK(\epsilon) + \lambda \mbI\right)^{-1}\mbY,
\end{equation}
where $\mbI$ denotes the $(N-1)\times (N-1)$ identity matrix.
Once $\hat{\bm\alpha}$ has been determined, out-of-sample model forecasting can be achieved via
\begin{equation}\label{eq: forecasting direct}
    \Tilde{\mbu}_{i+1} = K(\Tilde{\mbu}_i,\mbX;\epsilon)\hat{\bm\alpha},
\end{equation}
where $\Tilde{\mbu}_i$ is the kernel approximation to the given data.

\subsubsection{Skip-Connection Scheme}

Instead of directly modeling the flow map, a skip-connection scheme models each component of the residual map
\begin{equation}
    \Phi^{(\ell)}_\tau(\mbx) - \mbx^{(\ell)} \approx \sum_{i=1}^N k(\mbx,\mbu_i;\epsilon)\alpha_i^{(\ell)}.
\end{equation}
We define $\mbX$ according to \eqref{eq: X matrix clean} as before, but now define the output matrix to be
\begin{equation}\label{eq: skip-connection matrix}
        \mbY = \begin{bmatrix}
            \Delta\mbu_1 & \Delta\mbu_2 & \dots & \Delta\mbu_{N-1}
        \end{bmatrix}^\top, \quad \Delta \mbu_i = \mbu_{i+1} - \mbu_i.
    \end{equation}
    In this setup, our goal is to model
    \begin{equation}\label{eq: kernel skip-connection}
    \mbu_{i+1} - \mbu_i = \bm\Phi_{\Delta t}(\mbu_i) - \mbu_i \approx K(\mbu_i,\mbX;\epsilon)\bm\alpha.
\end{equation}
In the same notation as above, we arrive at a least squares problem \eqref{eq: least squares problem}, but with $\mbY$ now defined according to \eqref{eq: skip-connection matrix}.
Solving for the coefficients $\hat{\bm\alpha}$ proceeds as above, and forecasting is performed according to 
\begin{equation}\label{eq: forecasting skip}
    \Tilde{\mbu}_{i+1} = \Tilde{\mbu}_i + K(\Tilde{\mbu}_i,\mbX;\epsilon)\hat{\bm\alpha}.
\end{equation}

Notice that both the direct- and skip-connection schemes require the user to specify a kernel function, a bandwidth parameter $\epsilon$, and a regularization parameter $\lambda$.
We discuss appropriate choices for the kernel function below. 
We defer the discussion of the tunable parameters $\epsilon$ and $\lambda$ to \S\ref{subsec: validation}, where we  extend a previously developed validation strategy \cite{song2025learning} to handle noisy data using a weak formulation.

\subsection{Choice of Kernel Function}

The choice of kernel function in the KRR approach described above is crucial to the success of the method.
While any kernel function can be used, a specific choice of kernel may be
more, or less, appropriate for a given problem.

Suppose we are given a dataset $\mbU = \{\mbu_1,\dots,\mbu_N\}$ consisting of observations sampled from the forward invariant set $\mathcal{M}\subset \mathbb{R}^n$.
Unless otherwise stated, in this work we will consider a standard Gaussian kernel $k_{\text{RBF}}:\mathcal{M}\times \mathcal{M}\to \mathbb{R}$,
\begin{equation}\label{eq: Gaussian kernel}
    k_{\text{RBF}}(\mbx,\mby;\epsilon) = \exp\left(-\frac{\|\mbx-\mby\|_2^2}{4\epsilon} \right),
\end{equation}
for any $\mbx,\mby\in \mathcal{M}$.
The parameter $\epsilon>0$ is a scalar-valued bandwidth parameter that should be determined by the practitioner.
We describe an approach to choose $\epsilon$ in  \S\ref{subsec: validation} where we extend the KRR approach to noisy data.

We also consider the Diffusion Maps (DM) kernel, a data-driven kernel function based on the Diffusion Maps algorithm \cite{cl:2006} that has recently been employed with success over a wide range of datasets \cite{song2025learning}.
In particular, recent work has shown that the DM kernel can exhibit superior forecasting performance compared to the standard Gaussian kernel when applied to clean data, especially when the dimension of the invariant set $\mathcal{M}$ is much lower than its ambient dimension, $n$ \cite{song2025learning}.

We will compare the Gaussian and DM kernels over two baseline examples in \S\ref{sec: examples} when only noisy data is available.
We will show numerically that both kernels have nearly identical performance for a low-dimensional chaotic system, but that the DM kernel exhibits superior performance on a high-dimensional chaotic system with low intrinsic dimension.
These results suggest that systems with low intrinsic dimension or special geometry may benefit from the DM kernel, even when observational data is corrupted by noise.
Otherwise, the Gaussian kernel may be more appropriate due to lower computational complexity and ease of implementation.

We now describe the numerical implementation of the DM kernel.
Following \cite{song2025learning}, we begin with the standard Gaussian RBF kernel \eqref{eq: Gaussian kernel}.
We then normalize \eqref{eq: Gaussian kernel} in two stages to arrive at the DM kernel, $k_{\text{DM},N}:\mathcal{M}\times\mathcal{M} \to \BR$, by first writing
\begin{equation}
    k_{s,N}(\mbx, \mby;\epsilon) = \frac{k_{\text{RBF}}(\mbx,
    \mby;\epsilon)}{s_{N}(\mbx;\epsilon)  s_{N}(\mby;\epsilon)}, \qquad s_{N}(\mbx;\epsilon) = \frac{1}{N}\sum_{j = 1}^N  k_{\text{RBF}}(\mbx, \mbu_j;\epsilon),
\end{equation}
and then writing
\begin{equation}\label{eq: our discrete DM kernel}
k_{\text{DM},N}(\mbx,\mby;\epsilon) = \frac{{k}_{s,N}(\mbx,\mby;\epsilon)}{\sqrt{{q}_{N}(\mbx;\epsilon){q}_{N}(\mby;\epsilon)}},\qquad 
{q}_{N}(\mbx;\epsilon) = \frac{1}{N}\sum_{j = 1}^N 
k_{s,N}(\mbx, \mbu_j;\epsilon),
\end{equation}
for any $\mbx,\mby\in \mathcal{M}$.
Note that the DM kernel is not the same as the DM matrix \cite{cl:2006},
\begin{equation}\label{eq: DM matrix}
    P_{N} (\mbx,\mby;\epsilon) = \frac{
k_{s,N}(\mbx, \mby;\epsilon)}{\sum_{j=1}^N 
k_{s,N}(\mbx, \mbu_j;\epsilon)},
\end{equation}
that approximates a Markov transition kernel of a reversible Markov chain on $\mathcal{M}$. 
Observe that \eqref{eq: DM matrix} defines a Markov matrix because the normalizing denominator term here is a summation, not an average as in \eqref{eq: our discrete DM kernel}.

One can verify that
\begin{equation}\label{DMsymmetrickernel}
    \sqrt{N{q}_{N}(\mbx;\epsilon)} \frac{P_{N} (\mbx,\mby;\epsilon)}{\sqrt{N{q}_{N}(\mby;\epsilon)}} =  \frac{{k}_{s,N}(\mbx,
\mby;\epsilon)}{\sqrt{N{q}_{N}(\mbx;\epsilon) N{q}_{N}(\mby;\epsilon)}}= \frac{1}{N}k_{\text{DM},N}(\mbx,\mby;\epsilon),
\end{equation}
which suggests that the Gram matrix $\mathbf{P}(\epsilon)$ corresponding to $P_{N}$ is diagonally conjugate to the normalized Gram matrix $\mathbf{K}(\epsilon)/N$ corresponding to $k_{\text{DM},N}/N$. 
In particular, the eigenvalues of $\mathbf{P}(\epsilon)$ and $\mathbf{K}(\epsilon)/N$ are the same. 
We refer to \cite{harlim2026diffusion} for a theoretical discussion of the DM kernel in conjunction with KRR. 

\section{Filtering Noise Through a Weak Formulation}\label{sec: explaining the weak form}

The key learning mechanism underlying classical KRR is to enforce pointwise consistency across snapshot pairs of data.
While this strong formulation is often appropriate for clean data, its performance often noticeably worsens in the presence of noise because the regression may fit erroneous fluctuations instead of underlying dynamics.  

The weak formulation is a popular approach to address this lack of noise robustness.
Weak approaches integrate noisy data over a family of test functions, relaxing pointwise consistency in favor of orthogonality constraints.
While such approaches have recently been observed to offer increased robustness to noise \cite{bortz2024weak,li2025weak,messenger2021weak, messenger2021weak2, messenger2025asymptotic, messenger2024weak}, the connection to filtering was recently pointed out in \cite{li2025weak} and \cite{messenger2021weak}. 
Building on this intuition, the purpose of this section is to provide a brief overview of the filtering mechanism behind the weak formulation.

\subsection{Problem Statement and Notation}

A classical solution to the differential equation $\mbx'(t) = \mbf(\mbx(t))$ is a solution $\mbx(t)$ which satisfies 
\begin{equation}
    \mbR(\mbx(t)) \equiv \mbx'(t) - \mbf(\mbx(t)) = \bm0.
\end{equation}
In contrast, a weak formulation involves integrating the residual over smooth test functions $\varphi(t)$
\begin{equation}
    \int_{\BR} \td t \; \mbR(\mbx(t))\varphi(t) = \bm0,
\end{equation}
which may be interpreted as an $L^2$ inner product enforcing the orthogonality of the residual with the test functions.
This integral formulation, with appropriate choice of compactly supported test functions, has been observed to increase robustness to noise \cite{bortz2024weak,li2025weak,messenger2021weak, messenger2021weak2, messenger2025asymptotic, messenger2024weak}.

In the following, we will study weak formulations applied to an arbitrary noisy signal $\mbu(t)$,
\begin{equation}\label{eq: integral}
    \int_{\BR} \td t \; \mbu(t)\varphi_j(t) = \bm0,
\end{equation}
for an appropriate family of test functions $\{\varphi_j\}$. 
While we will ultimately employ the weak formulation in conjunction with KRR, the analysis in this section does not depend on a specific learning method.

Let $\varphi(t)$ be a smooth test function with compact support on $[0,L]$, and define $\varphi_j(t) = \varphi(t-jh)$ to be a family of uniformly translated test functions with compact support on $[L_{j}, L_{j+1}]$.
The parameter $h$ determines the distance between the adjacent test functions, and together with $L$, describes the extent to which adjacent test functions overlap.
For a dataset $\mbu_i\equiv\mbu(t_i)$ of finite length and fixed $h$, the indices $j$ for which the support of $\varphi_j(t)$ lies in the sampling time is restricted.
If the signal consists of $N$ points sampled uniformly with step $\Delta t$, we adhere to the convention that $j\in [1,2,\dots,k^*]$, where
\begin{equation}
    k^* = \left\lfloor \frac{N\Delta t-L}{h} \right\rfloor.
\end{equation}

In practice, the integral \eqref{eq: integral} must be computed with quadrature over the available data
\begin{equation}\label{eq: quadrature integral friend}
    c_j^{(\ell)} = \left\langle \varphi_j(t), u^{(\ell)}(t) \right\rangle = \int_{L_j}^{L_{j+1}} \td t\; \varphi_j(t) u^{(\ell)}(t) \approx \sum_{i=1}^{N} w_i \varphi_j(t_i) u_i^{(\ell)}, \quad \ell = 1,\dots,n, \quad j=1,\dots,k^*,
\end{equation}
where $w_i$ are appropriate quadrature weights.
We will find it convenient to express these quadrature approximations as a matrix product.
To that end, collect the translated test functions into the rows of a matrix $\bm\Psi\in \mathbb{R}^{k^* \times N}$ whose $(j,i)$ entry is $\varphi_j(t_i)$.
Note that $\bm\Psi$ is often sparse due to the compact support of the test functions.
Further define $\mbW=\operatorname{diag}(w_1,\dots,w_N)\in \mathbb{R}^{N\times N}$ to be a diagonal matrix of quadrature weights.
We assume that $\mbW$ is invertible throughout the manuscript.
Note that $\bm\Psi\mbW $ is the matrix whose rows are test functions scaled by the quadrature weights.

With this notation, we may discretize \eqref{eq: quadrature integral friend} as a linear system
\begin{equation}\label{eq: more coefficient nonsense}
    \mbC =  \bm\Psi \mbW \mbU,
\end{equation} 
where $\mbC\in \mathbb{R}^{k^*\times n}$ is the matrix of inner product coefficients $c_j^{(\ell)}$, and $\mbU = [\mbu_1,\dots,\mbu_N]^\top\in \mathbb{R}^{N\times n}$ collects the data at available sample times.
Notice that in the typical case that $k^*<N$, the coefficients $\mbC$ represent a compressed representation of the original signal $\mbU$.

\subsection{The Weak Formulation as a Filter}\label{subsec: weak form as a filter}

To see the weak formulation as a filter, we consider the process of reconstructing a signal from a set of coefficients $\mbC$.
Let ${\hat{\mbu}}(t)=(\hat u^{(1)}(t),\dots,\hat u^{(n)}(t))$ denote the reconstructed signal, which we now express as a linear combination of the available test functions
\begin{equation}\label{eq: expansion nonsense}
     \hat{u}^{(\ell)}(t) = \sum_{j=1}^{k^*} a_j^{(\ell)}\varphi_j(t), \quad \ell=1,\dots,n,
\end{equation}
where $a_j^{(\ell)}$ are unknown coefficients to be determined.

In the event that the $\varphi_j(t)$ are orthonormal, taking inner products on both sides of \eqref{eq: expansion nonsense} is sufficient to isolate the coefficients.
Without orthogonality, we may still recover the coefficients by requiring that the inner product of the reconstruction with test functions $\varphi_q(t)$ agree with the given coefficients,
\begin{equation}\label{eq: reconstruction nonsense}
    \mbC_{q,\ell}=\langle\hat{u}^{(\ell)}(t),\varphi_q(t) \rangle = \sum_{j=1}^{k^*} a_j^{(\ell)}\langle\varphi_j(t),\varphi_q(t)\rangle.
\end{equation}

 We will find it useful to express \eqref{eq: reconstruction nonsense} as a matrix equation.
Let $\mbA\in \mathbb{R}^{k^* \times n}$ be the collection of unknown coefficients whose $(j,\ell)$ entry is $\alpha_j^{(\ell)}$, and let $\mbG = \bm\Psi \mbW \bm\Psi^\top \in \mathbb{R}^{k^*\times k^*}$ be the matrix whose $(q,j)$ entry is a quadrature approximation of the inner product $\langle \varphi_q(t), \varphi_j(t)\rangle$.
Under the assumption that the test functions are linearly independent and overlapping, $\mbG$ is invertible.
We can then express \eqref{eq: reconstruction nonsense} as the linear system $\mbC = \mbG\mbA$.
It follows that
\begin{equation}\label{eq: coefficient nonsense}
    \mbA = \mbG^{-1}\mbC.
\end{equation}

Let $\hat{\mbU}= [\hat\mbu_1,\dots,\hat\mbu_N]^\top \in \mathbb{R}^{N\times n}$ denote the matrix collecting the reconstructed signal at the available discrete sample times.
We can now express \eqref{eq: expansion nonsense} in matrix notation to arrive at an expression for $\hat\mbU$, 
\begin{equation}
    \hat{\mbU} = \bm\Psi^\top \mbA = \bm\Psi^\top \mbG^{-1}\mbC = \bm\Psi^\top \mbG^{-1} \bm\Psi \mbW \mbU =  \bm\Psi^\top ( \bm\Psi\mbW\bm\Psi^\top)^{-1}\bm \Psi\mbW\mbU,
\end{equation}
where the second equality follows from \eqref{eq: coefficient nonsense} and the third follows from \eqref{eq: more coefficient nonsense}.
Define $\mbP \equiv \bm\Psi^\top ( \bm\Psi\mbW\bm\Psi^\top)^{-1}\bm \Psi\mbW \in \mathbb{R}^{N\times N}$.
Note that $\mbP$ is a projection matrix because $\mbP^2=\mbP$.
Moreover, we see that 
\begin{equation}
    \langle \mbP\mbu,\mbv\rangle_{\mbW}:= (\mbP\mbu)^\top\mbW \mbv = \mbu^\top \mbP^\top \mbW \mbv = \mbu^\top\mbW \bm\Psi^\top \mbG^{-1}\bm \Psi \mbW \mbv = \mbu^\top \mbW (\mbP\mbv)= \langle \mbu,\mbP\mbv\rangle_{\mbW},
\end{equation}
for any $\mbu,\mbv \in \BR^N$, which implies that $\mbP$ is self-adjoint with respect to the inner-product weighted by the quadrature weight matrix $\mbW$.
In particular, we have
\begin{equation}\label{eq: self-adjoint}
    \mbP^\top\mbW = \mbW \mbP.
\end{equation}
These observations imply that $\mbP$ is an orthogonal projection under the inner product $\|\mbP\|_W^2 = \text{tr}(\mbP^\top \mbW \mbP)$.

The formulation above suggests that the weak-form reconstruction procedure can be interpreted as an orthogonal projection onto the span of a family of test functions,  
\begin{equation}\label{eq: weak reconstruction formula}
    \hat{\mbU} = \mbP \mbU,
\end{equation}
which allows weak formulations to be interpreted in light of classical results in information theory and signal processing, such as Shannon's sampling theorem and related results which use the language of orthogonal projection to study filtering \cite{shannon1949communication, unser2002sampling, unser2002general}.

\subsubsection{Choice of Test Function}\label{subsec: choice of test function}

In this section, we consider a specific choice of test function that was found to be successful in the literature \cite{messenger2021weak},
\begin{equation}\label{eq: poly test function}
    \varphi(t) = \begin{cases}
        C_p (t-a)^p(b-t)^p, & a<t<b
        \\
        0, & \text{otherwise}
    \end{cases}, \qquad C_p = \left(\frac{2}{L}\right)^{2p},
\end{equation}
where $L=b-a$ is the support length and $p\in \mathbb{Z}^+$ is a parameter which specifies the degree of the polynomial.
Note that $\varphi(t)$ is smooth and compactly supported.
We will use \eqref{eq: poly test function} in all of our numerical examples in \S\ref{sec: examples}.

Recall that we construct the matrix $\bm \Psi\mbW$ by generating a family of linearly independent test functions, $\varphi_j(t) = \varphi(t - jh)$, where $h>0$ dictates the overlap between test functions.
Consequently, the polynomial test functions depend on three key parameters: (i) the polynomial degree $p$, (ii) the support length $L$, and (iii) the overlap $h$.
Appropriate values for the triple $(p,L,h)$ may vary significantly across datasets and noise levels.
In practice, an appropriate choice of these parameter values should balance speed and reconstruction accuracy.

There are many approaches to empirically measure the accuracy of the weak-form reconstruction. 
If clean or high-quality reference data is available, one could select parameters such that the reconstruction of the noisy data most closely aligns with the clean reference data under a suitable metric, e.g., RMSE.
If no clean reference data is available, a common approach is to define a score metric that can be evaluated from only the observed data.
Such metrics commonly measure whiteness, variance ratio, or smoothness ratio \cite{box2015time,kalman1960new,li2025weak,ljung1978measure, maybeck1982stochastic, savitzky1964smoothing}.
We refer to \cite{li2025weak} for a discussion and numerical implementation of this approach.
Meanwhile, the number of test functions $k^*$ is inversely proportional to the overlap parameter $h$.
We will show in \S\ref{sec: proposed approach} that taking $h$ large has the potential to greatly reduce computational complexity requirements.

\begin{figure}[htbp]
    \centering
    \begin{minipage}[b]{.98\textwidth}
        \includegraphics[width=\textwidth]{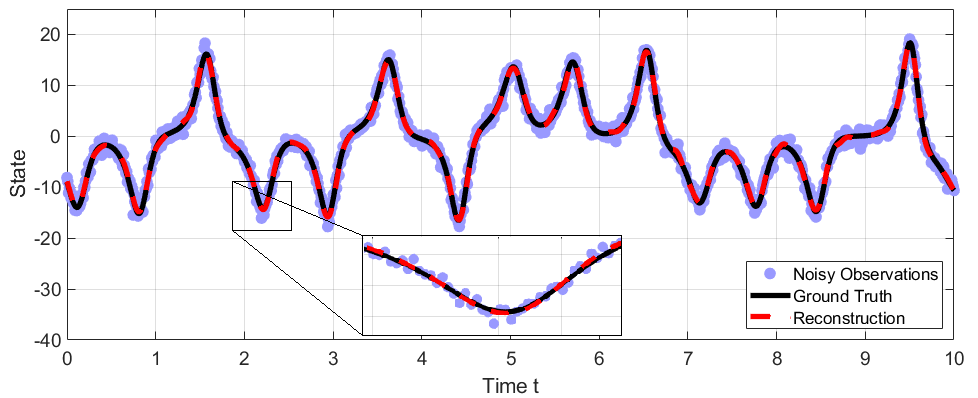}
    \end{minipage}
    \caption{An illustration that appropriately chosen test functions can filter noisy data.
    A noisy signal (blue dots) is reconstructed (dashed red) via \eqref{eq: weak reconstruction formula} using the test function \eqref{eq: poly test function} and compared to the ground truth (black).}
    \label{fig: typical reconstruction}
\end{figure}

In this work, we select polynomial parameters that are observed to filter given noisy data across a range of error metrics.
As a qualitative example, we consider here the Lorenz-63 system, discussed in more detail in \S\ref{sec: examples}, with 10\% noise corruption.
Figure \ref{fig: typical reconstruction} visualizes a typical filtering result for the $x$-component with $(p,L,h)=(5,\,0.5,\,0.1)$. 
We remark that filtering with appropriately chosen polynomials can provide competitive or  superior results to standard approaches, such as wavelet-based filtering.
We refer to Appendix \ref{appendix: test function parameters} for a quantitative comparison of polynomial and wavelet filtering across a range of error metrics, noise levels, and model systems.
The success of the filtering procedure justifies this choice of test function and explains the noise robustness of the weak formulation.  

\subsection{Bias-Variance Decomposition}\label{subsec: bias variance decomp}

To analyze the error in the general reconstruction \eqref{eq: weak reconstruction formula}, let $\bm{\mathcal{E}}=\hat{\mbU}-\mbU_{\text{clean}}$ denote the difference between the underlying clean data and the reconstruction of the noisy signal.
It follows from the decomposition
\begin{equation}
    \hat{\mbU} = \mbP \mbU = \mbP (\mbU_{\text{clean}} + \sigma \bm\xi), 
\end{equation}
where $\bm\xi = [\bm\xi_1,\dots,\bm\xi_N]^\top \in \mathbb{R}^{N\times n}$ is the collection of noise vectors, that we may write 
\begin{equation}\label{eq: bias variance decomposition}
    \mathbb{E}\left[\|\bm{\mathcal{E}}\|_W^2\right] = \sum_{\ell=1}^n B_\ell^2 + V_\ell^2, \quad \quad B_\ell^2 = \|(\mbP - \mbI)\mbU_{\text{clean}}^{(\ell)}\|_W^2, \quad V_\ell^2 = \sigma^2\eta_\ell^2\text{tr}(\mbW\mbP),
\end{equation}
which gives a bias-variance decomposition of the reconstruction error.
The terms $B_\ell^2$ correspond to the ``bias'' of the reconstruction, which describe the fidelity with which a clean signal can be reconstructed.
It measures how well the clean signal can be approximated in the span of the test functions.
The terms $V_\ell^2$ correspond to the ``variance'' of the signal, which describes additional error arising from noise corruption.

The variance term admits significant simplification in the case of uniform quadrature, $\mbW = c\mbI$ for $c\in \mathbb{R}$,
\begin{equation}\label{eq: trace}
    \text{tr}(\mbW\mbP) = c\cdot \text{tr}(\mbP) = c\cdot\text{rank}(\mbP) = c k^*,
\end{equation}
which holds because $\mbP$ is a projection matrix.
Equation \eqref{eq: trace} implies that filtering error due to noise is governed by the number of test functions, $k^*$, and is independent of the form of the test function.
Notice that the bias term is similarly scaled by $c$ in this case, so that a specific choice of this constant does not artificially inflate either the bias or variance error relative to the other.

In contrast, the bias term generically depends on the choice of test function and is generally difficult to analyze.
Performing computations in Fourier space has the potential to simplify the form of the bias.
One can show that in the ideal case of periodic data and uniform quadrature, $\mbW=\mbI$, the bias and variance have the form
\begin{equation}\label{eq: bias variance simplification}
    B_\ell^2 = \left\|(\mbH - \mbI)\mathcal{F}\left[\mbU_{\text{clean}}^{(\ell)}\right]\right\|_F^2, \quad V_\ell^2 = \sigma^2 \eta_\ell^2 k^*.
\end{equation}
where $k^* = N\Delta t/ h$ (due to the periodicity of the data) and $\mathcal{F}[\cdot]$ denotes the Fourier transform.
The matrix $\mbH\in \mathbb{R}^{N\times N}$ is a Fourier representation of $\mbP$,
\begin{equation}\label{eq: H}
    \mbH_{mn} = 
    \begin{cases}
        \frac{\hat{\varphi}_m \overline{\hat{\varphi}_n}}{\sum_{\ell=0}^{h-1} |\hat{\varphi}_{r+\ell k^*}|^2}, & m = n \;(\text{mod } k^*),
        \\
        0, & \text{otherwise},
    \end{cases} 
\end{equation}
where $r=m\text{ mod } k^*$ and $\hat{\varphi}_m$ is the $m$th component of the Fourier transform of the untranslated test function.
Note that $\mbH$ is sparse with nonzero banded diagonal entries.
The entries on the off-diagonal corresponding to aliasing artifacts.
We remark that \eqref{eq: H} holds for a general test function.
However, further simplification for a specific test function is often difficult or impossible.
We provide a brief derivation of \eqref{eq: H} in Appendix \ref{appendix: bias derivation}.
Finally, we remark that approximation theory may be used to bound the bias under certain conditions in terms of the sampling step of the data.
We refer to
\cite{aldroubi1994sampling,blu1999approximation, unser2002sampling, unser2002general} for further theoretical discussion of the bias reconstruction error. 

In the remainder of this section, we numerically study the bias and variance in \eqref{eq: bias variance simplification} with the specific choice of test function \eqref{eq: poly test function}.
We assume the ideal case of periodic data and uniform quadrature, $\mbW = \mbI$.
Similarly to \cite{messenger2021weak}, we observe that in certain parameter regimes the bias term may act as an ideal low-pass filter and is well approximated by the tail of the Fourier modes of the underlying clean signal,
\begin{equation}\label{eq: tail}
    B_\ell^2 \approx \sum_{m>k^*} \left|\mathcal{F}[\mbU_{\text{clean}}^{(\ell)}]\right|^2.
\end{equation}
To perform numerical experiments, we consider a noisy dataset, $\mbU = [\mbu_1,\dots,\mbu_N]$, and its clean counterpart, $\mbU_{\text{clean}}$, that have been generated from the Lorenz-63 system \eqref{eq: Lorenz 63 equations} (see \S\ref{sec: examples}).
We take $N=1000$, uniform timestep $\Delta t = 0.01$, and $\sigma = 0.1$.
To proceed, we numerically observe the total error,
\begin{equation}
    \mathbb{E}\left[\|\bm{\mathcal{E}}\|_F^2\right] = \mathbb{E}\left[\|\mbP \mbU - \mbU_{\text{clean}}\|_F^2\right],
\end{equation}
and the bias error
\begin{equation}\label{eq: oberved bias}
    \mathbb{E}\left[\|\bm{\mathcal{E}}_{\text{bias}}\|_F^2\right] = \mathbb{E}\left[\|\mbP \mbU_{\text{clean}} - \mbU_{\text{clean}}\|_F^2\right].
\end{equation}
The observed variance error is then computed according to 
\begin{equation}\label{eq: observed variance}
    \mathbb{E}\left[\|\bm{\mathcal{E}}_{\text{var}}\|_F^2\right] = \mathbb{E}\left[\|\bm{\mathcal{E}}\|_F^2 - \|\bm{\mathcal{E}}_{\text{bias}}\|_F^2\right].
\end{equation}
Above, the expectation is empirically approximated over 100 identical trials corrupted with independent noise.

\begin{figure*}[htpb]
    \centering
    \begin{minipage}{0.3\textwidth}
        \centering
        \includegraphics[width=\textwidth]{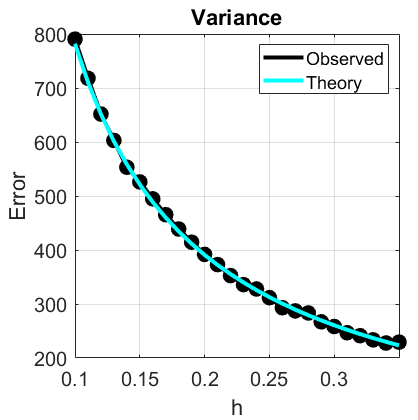}
        \includegraphics[width=\textwidth]{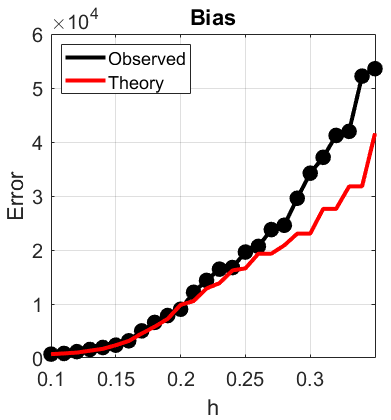}
        \small \textbf{(a) $p=12$ and $L=150$}
    \end{minipage}
    \hfill
    \begin{minipage}{0.3\textwidth}
        \centering
        \includegraphics[width=\textwidth]{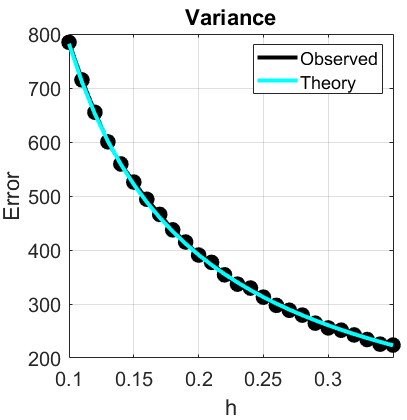}
        \includegraphics[width=\textwidth]{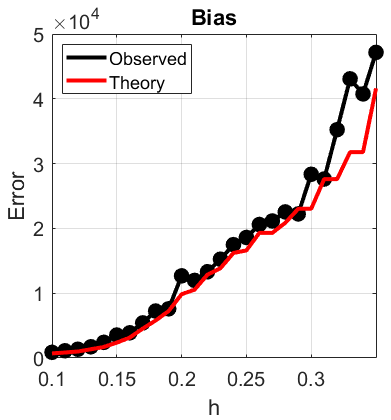}
        \small \textbf{(b) $p=4$ and $L=120$}
    \end{minipage}
    \hfill
    \begin{minipage}{0.3\textwidth}
        \centering
        \includegraphics[width=\textwidth]{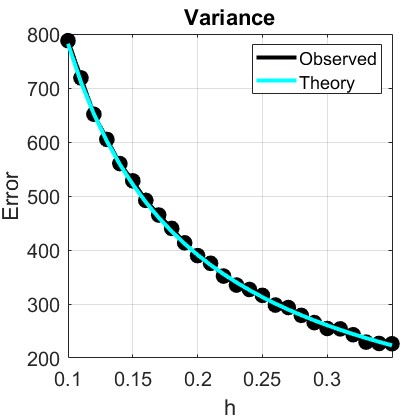}
        \includegraphics[width=\textwidth]{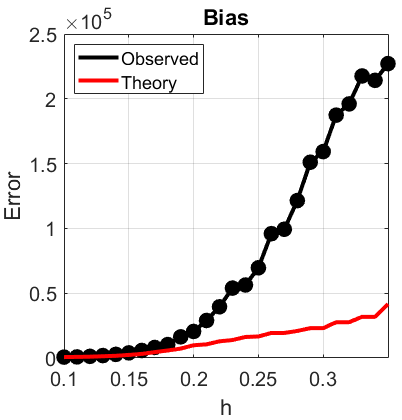}
        \small \textbf{(c) $p=10$ and $L=70$}
    \end{minipage}

    \caption{\textbf{Top:} Plots of the observed variance \eqref{eq: observed variance} as a function of $h$ with the theoretical expression for $V_\ell^2 = k^*$ from \eqref{eq: bias variance decomposition}.
    In all cases, there is excellent agreement. 
    \textbf{Bottom:} Plots of the observed bias \eqref{eq: oberved bias} as a function of $h$ with the approximation \eqref{eq: tail}. There is good agreement for large $L$ and small $h$.
    As $h$ increases, the approximation worsens.}
    \label{fig: bias_var_examples}
\end{figure*}

Results are reported in Figure \ref{fig: bias_var_examples} for different test-function parameters.
We demonstrate that the form of $V_\ell^2$ in \eqref{eq: bias variance simplification} captures the scaling of the observed variance, and observe that the bias may be well-approximated by \eqref{eq: tail} for large $L$ and small $h$.

\section{Proposed Approach: Weak-form Kernel Ridge Regression}\label{sec: proposed approach}

In this section, we propose Weak-form Kernel Ridge Regression (WKRR) as a noise robust extension of the strong KRR approach.
We will refer to this framework as the ``weak'' approach throughout the manuscript.

\subsection{Weak-form Kernel Ridge Regression}\label{subsec: proposed approach}

To understand WKRR, we follow the formulation of the strong case as in \S\ref{subsec: KRR} up to the formulation of the least squares problem \eqref{eq: least squares problem}, which enforces pointwise consistency across the given data.
In the context of a weak formulation, we will interpret the least squares problem as a residual, $\mbE = \mbY - \mbK(\epsilon)\bm{\alpha}$ with $\mbY$ and $\mbK(\epsilon)$ now depending on noisy data, and  select $\bm{\alpha}$ to enforce orthogonality of the residual $\mbE$ against a family of test functions.

As in \S\ref{sec: explaining the weak form}, let $\varphi_j(t)$ denote a family of smooth test functions, and let $\bm\Psi\mbW \in \mathbb{R}^{k^*\times (N-1)}$ be the matrix whose rows are the test functions scaled by quadrature weights.
Notice here that $\bm\Psi\mbW$ has $N-1$ rows to be consistent with the snapshot data $S$.
The core idea behind WKRR is to enforce the constraint $\bm\Psi\mbW\mbE = \bm0$.
In a literal sense, this procedure approximates the integral of the residual over test functions by a quadrature approximation.
In the sense described in \S\ref{sec: explaining the weak form}, this procedure amounts to projecting the noisy data onto the span of the test functions, which we interpret as filtering the noisy data.
Accordingly, we consider the modified least squares problem
\begin{equation}\label{eq: least squares problem weak}
    \mbY_\mbW = \mbK_\mbW(\epsilon)\bm\alpha,
\end{equation}
where $\mbY_\mbW = \bm\Psi\mbW \mbY \in \mathbb{R}^{k^*\times n}$ and $\mbK_\mbW(\epsilon) = \bm\Psi\mbW\mbK(\epsilon)\in \mathbb{R}^{k^*\times (N-1)}$.

Naively, classical KRR regularization would involve the term $\lambda \bm\alpha^\top \mbK_\mbW(\epsilon) \bm\alpha$; however, this expression is not defined because the weak-form kernel matrix is no longer square.
Instead, we regularize in the weak coefficient space in terms of the matrix $\widetilde{\mbK}_\mbW(\epsilon) = \bm\Psi\mbW\mbK(\epsilon)(\bm\Psi\mbW)^\top \in \mathbb{R}^{k^*\times k^*}$, leading to the problem
\begin{equation}\label{eq: first weak KRR equation}
    \hat{\bm\alpha}_\mbW = \arg\min_{\bm\alpha_\mbW} \left(\| \widetilde{\mbK}_\mbW(\epsilon)\bm\alpha_\mbW  - \mbY_\mbW\|_F^2 + \lambda \bm\alpha_\mbW^\top \widetilde{\mbK}_\mbW(\epsilon)\bm\alpha_\mbW \right).
\end{equation}
Note that $\bm\alpha_\mbW \in \mathbb{R}^{k^*\times n}$ is a weak-form representation of the coefficients $\bm\alpha \in \mathbb{R}^{(N-1)\times n}$ in the original coordinates.
To obtain a solution in the original coordinates, we study \eqref{eq: first weak KRR equation} with the substitution $\bm\alpha = (\bm\Psi\mbW)^\top \bm\alpha_\mbW$, which leads to the equivalent problem
\begin{equation}\label{eq: weak H KRR equation}
    \hat{\bm\alpha} = \arg\min_{\bm\alpha} \left(\| \mbK_\mbW(\epsilon)\bm\alpha - \mbY_\mbW\|_F^2 + \lambda \bm\alpha^\top \mbK(\epsilon)\bm\alpha \right),
\end{equation}
with unique solution given by
\begin{equation}\label{eq: weak H alpha hat}
    \hat{\bm\alpha} = (\bm\Psi\mbW )^\top (\widetilde{\mbK}_\mbW(\epsilon) + \lambda \mbI)^{-1}\mbY_\mbW.
\end{equation}
Once $\hat{\bm\alpha}$ is available, forecasting is performed via \eqref{eq: forecasting direct} or \eqref{eq: forecasting skip}.
We emphasize that forecasting is performed in the original coordinates, not in the weak-coefficient space.

Note that the weak least squares formulation arises from the strong formulation left-multiplied by the matrix $\bm\Psi\mbW$.
Although this procedure appears very simple, it grants a surprising number of benefits.
In addition to filtering, we will find that performing KRR in the weak-coefficient space reduces computational complexity and may increase model robustness compared to a strong implementation.

We remark that classically, a weak formulation involves integration-by-parts when integrating both sides of a differential equation over a family of test functions.
Here, when learning discrete-time solution operators, we do not integrate by parts.
Nevertheless, we continue to use the term ``weak'' due to the fact that these approaches enforce the differential equations in the integral or weak sense.
We note that the approach described here can easily be extended to learn the vector field of a dynamical system, and emphasize that the interpretation of the weak form as a filter remains unaltered in a continuous-time context.

The success of the WKRR approach hinges on the choice of test function used to construct the matrix $\bm\Psi\mbW$, and the hyperparameters $\epsilon$ and $\lambda$.
In this work, we consider only the test function \eqref{eq: poly test function} described in \S\ref{subsec: choice of test function}. 
Developing a systematic or optimal framework for choosing test function parameters $(p,L,h)$ is beyond the scope of the present work.
Instead, our objective is to demonstrate the robustness of the WKRR approach across a representative range of parameter values.
In the numerical experiments (see \S\ref{sec: examples}), we select fixed representative values of $(p,L,h)$ and demonstrate their success in the WKRR framework.
Additional diagnostics are provided in Appendix \ref{appendix: test function parameters}.
The following section describes a validation procedure to choose the parameters $\epsilon$ and $\lambda$, which extends the methodology implemented in \cite{song2025learning}.

\subsection{Validation}\label{subsec: validation}

The success of WKRR depends on appropriate selection of the bandwidth and regularization parameters $(\epsilon,\lambda)$.
In this section, we propose a validation procedure that we observe to be successful across a range of chaotic, high-dimensional, and experimental datasets; see \S\ref{sec: examples}.

Typically, one reserves a portion of given observation data for use in validation.
This validation data is not seen during training, and is used as a form of ground truth to select appropriate model hyperparameters.
In this section we let $\mbu_{\text{train},i}$ for $i=1,\dots,N$ and $\mbu_{\text{val},i}$ for  $i=1,\dots,N_V$ denote noisy training and validation data, respectively.
While training and validation can be performed on the raw data, in practice we observe that filtering prior to training and validation yields marginally better performance.
We refer to \S\ref{subsec: preprocess} for a more detailed discussion of our pre-processing procedure.
We denote $\hat\mbu_{\text{train},i}$ and $\hat\mbu_{\text{val},i}$ as filtered training and validation data, respectively.
In the present discussion, we assume training and validation are performed on the filtered data.
In our numerical experiments in \S\ref{sec: examples}, we consider training on both noisy and filtered data.

We assume that the matrix $\bm\Psi\mbW$ is available.
For a fixed $(\epsilon,\lambda)$ pair, let $\mbK(\epsilon)$ be the kernel Gram matrix formed over the filtered training data $\hat\mbu_{\text{train},i}$, and let $\hat{\bm\alpha}$ be computed according to \eqref{eq: weak H alpha hat}.
Denote $\tilde{\mbx}_i$ as the corresponding kernel forecast over the filtered validation data $\hat\mbu_{\text{val},i}$ via either \eqref{eq: forecasting direct} or \eqref{eq: forecasting skip}.
We assume that $\tilde{\mbx}_1 = \hat\mbu_{\text{val},1}$, i.e., that the forecast is initialized with the first step of known validation data.
We now review two error metrics that we will use for validation in the numerical examples considered in this manuscript.

\begin{enumerate}
    \item \textit{Valid Prediction Time (VPT):} We consider the VPT metric for chaotic dynamical systems
    \begin{equation}
        \text{VPT} = \Lambda\cdot t_\mathcal{I}, \quad \mathcal{I} =  \max \{\mathfrak{I}\mid E_i \leq \gamma \text{ for all } 1\leq  i \leq \mathfrak{I}\}, 
    \end{equation}
    where $\Lambda$ is the maximal Lyapunov exponent of the system, $\gamma$ is a user-defined threshold, and $E_i$ is a normalized error metric
    \begin{equation}\label{eq: VPT}
        E_i = \sqrt{\frac{1}{n}\sum_{\ell=1}^n \left(\frac{\Tilde{x}_i^{(\ell)}-\hat u_{\text{val},i}^{(\ell)}}{\zeta_\ell}\right)^2  },
    \end{equation}
    where $\zeta_\ell$ is the standard deviation of the $\ell$th component of the filtered validation data.

    In words, VPT measures how long the normalized error between the kernel predication and the validation data stays below a given threshold $\gamma$.
    The VPT metric is ideal for chaotic systems because the forecast is expected to diverge after a short time.
    Normalization by the Lyapunov exponent $\Lambda$ ensures the metric is unitless and enables fair comparison with other chaotic systems with faster or slower dynamics.

    \item \textit{Normalized Mean Square Error (NMSE):} For all other dynamical systems considered in this work, we use the NMSE metric
    \begin{equation}\label{eq: NMSE}
        E_i = \frac{\sum_{\ell=1}^n |\Tilde{x}_i^{(\ell)}-\hat u_{\text{val},i}^{(\ell)}|^2 }{\sum_{\ell=1}^n |\hat u_{\text{val},i}^{(\ell)}|^2}.
    \end{equation}
\end{enumerate}

Our goal now is to select appropriate hyperparameters $(\epsilon,\lambda)$ for use in the weak KRR problem \eqref{eq: weak H KRR equation}.
While systematic grid searches are possible, they are often computationally expensive.
Instead, to save computation time, we follow a data-driven heuristic method, outlined in Appendix \ref{appendix: validation heuristic}, to compute a reference bandwidth $\epsilon^*$ and a reference regularization parameter $\lambda^*$.
These reference values are used to confine the parameter search to a grid $H=[\epsilon^*\Delta \epsilon, \epsilon^*/\Delta \epsilon] \times [\lambda^*,\lambda^*/\Delta \lambda]$, where $\Delta \epsilon = 10^{-4}$ and $\Delta \lambda = 10^{-5}$.

Apart from the grid, the length of data over which forecasts are performed also plays a crucial role.
In the case of clean data, it is often sufficient to validate over a few long trajectories that match or exceed the expected forecast horizon, as in \cite{song2025learning}.
However, such a procedure may yield unstable results when the available data is noisy; predictions may diverge much more rapidly due to fluctuations, causing the model to fit noise instead of underlying dynamics.
To address this concern, here we propose to validate over many short segments, and select parameters that optimize average error over all segments.
To reduce computational complexity, we first select parameters from a coarse grid $H_c$.
We then zoom in on the optimal parameter choice from the coarse grid, and form a second refined grid $H_f$.
Validation over all $(\epsilon,\lambda)\in H_f$ yields an optimal parameter pair, which is subsequently used to form the final model.

We now provide implementation details for the proposed validation procedure.

\textit{Coarse validation:}
\begin{enumerate}
    \item \textit{Form the coarse grid:} Select reference parameters $(\epsilon^*,\lambda^*)$, form the coarse grid $H_c=[\epsilon^*\Delta \epsilon, \epsilon^*/\Delta \epsilon]\times [\lambda^*,\lambda^*/\Delta \lambda]$, and divide $H_c$ into an $N_{\text{mesh}}\times N_{\text{mesh}}$ mesh with uniformly spaced gridpoints in log-space.
    \item \textit{Divide validation data:} Divide the given validation data into $N_c$ (possibly overlapping) segments of length $\nu_c$, with initial conditions chosen uniformly and randomly from the full validation dataset. 
    \item \textit{Learn and forecast:} For each $(\epsilon,\lambda)$ pair in $H_c$, learn the kernel solution operator via equation \eqref{eq: weak H alpha hat}, forecast over the $N_c$ validation segments, and compute the average error over each segment.
    \item \textit{Select optimal coarse parameters:} Select the hyperparameter pair $(\epsilon_c,\lambda_c)$ that results in the best error over the $N_{\text{mesh}}^2$ measurements.
    \end{enumerate}

    \textit{Fine validation:}
    \begin{enumerate}
    \item \textit{Form the fine grid:} Define a zoom-in factor $F$ and a fine grid $H_f = [\epsilon_c/F,\epsilon_c\cdot F]\times [\lambda_c/F,\lambda_c\cdot F]$. 
    Break $H_f$ into an $N_{\text{mesh}}\times N_{\text{mesh}}$ mesh with uniformly spaced gridpoints in log-space. 
    \item \textit{Divide validation data:} To prevent overfitting, divide the given validation data into $N_f$ (possibly overlapping) segments of length $\nu_f$, with initial conditions chosen uniformly and randomly from the full validation dataset.
    This partition is independent from the previous partition in step 2.
    \item Repeat steps 3-4 above using the fine mesh $H_f$ to extract final hyperparameters $(\epsilon_f,\lambda_f)$.
\end{enumerate}

Unless otherwise stated, we take $N_{\text{mesh}}=20$, $N_c=20$, $N_f=30$, $F=10^{3/4}$, and $\nu\equiv\nu_c=\nu_f$.
The lengths $\nu_c$ and $\nu_f$ depend on the available data and should be chosen on a problem specific basis.

Notice that in the above procedure, the strong formulation requires inverting an $N\times N$ matrix for each parameter pair, which is computationally expensive when $N$ is large.
In contrast, WKRR involves inverting a $k^*\times k^*$ matrix, which greatly reduces computational cost in the typical case that $k^* \ll N$.

\subsection{Signal Pre-Processing}\label{subsec: preprocess}

Application of a weak formulation applied to KRR can be interpreted as filtering prior to regression.
As we will show numerically in \S\ref{sec: examples}, some form of filtering is essential prior to model training in the presence of noisy data, whether the filtering is accomplished via explicit pre-processing or implicitly via a weak formulation.
In our numerical experiments, we compare three approaches: (i) a strong formulation trained on pre-processed (filtered) data, (ii) a weak formulation trained on noisy data (but implicitly filtered via projection), and (iii) a weak formulation trained on pre-processed data (and further filtered implicitly via projection).
In all cases, we validate on filtered data for the sake of consistency with the training data.

Because the weak formulation acts to filter data, it is natural to compare the proposed WKRR approach to a strong approach applied to filtered data.
We now explain that WKRR grants several benefits beyond just filtering.
\begin{itemize}
    \item \textit{Accuracy:} We will show numerically in \S \ref{sec: examples} across several examples that WKRR exhibits similar or superior forecasting performance over strong methods when trained on identical filtered data.
    \item \textit{Speed:} WKRR leads to a compressed representation of observational data in the typical case of fewer test functions than samples, $k^*<N$.
    Consequently, training and validation involve inverting matrices of size $k^*\times k^*$, instead of $N\times N$ required for the strong formulation.
    \item \textit{Validation robustness:} In our numerical experiments, several strong models performed poorly during testing due to non-optimal validation.
    In contrast, WKRR models trained on identical data did not produce such outliers.
    This observation suggests that the weak-coefficient representation of the data may robustify the validation process, making specific hyperparameter choices less critical for success.
\end{itemize}

These observations lead us to incorporate data pre-processing in the WKRR pipeline.
We remark that increased performance due to filtering followed by a weak formulation has been reported in the literature for other learning methodologies and is not unique to WKRR \cite{messenger2025asymptotic}.
However, we emphasize that while pre-processing often enhances WKRR forecasting performance, WKRR without pre-processing still performs robustly across a range of examples.

\subsection{Summary}\label{subsec: summary}

We summarize the implementation of the proposed WKRR approach in Algorithm \ref{alg:summary}.
We assume that noisy observational data $\mbu_i$ of the form \eqref{eq: data} is given, which may be divided into training data $\mbu_{\text{train}}$ and validation data $\mbu_{\text{val}}$.

\begin{algorithm}
\caption{WKRR implementation}\label{alg:summary}

\begin{algorithmic}[1]
\Require \parbox[t]{\dimexpr.9\linewidth-\algorithmicindent}{Training data $\mbu_{\text{train}}$, and validation data $\mbu_{\text{val}}$.}

\State Form the test function matrix $\bm\Psi\mbW$.
See \S\ref{subsec: choice of test function} for discussion concerning test function selection.

\State Pre-process to obtain filtered data $\hat{\mbu}_{\text{train}}$ and $\hat{\mbu}_{\text{val}}$ according to \eqref{eq: weak reconstruction formula} or any other suitable method.

\State Use $\hat{\mbu}_{\text{train}}$ to form $\mbX$ according to \eqref{eq: X matrix clean} and $\mbY$ according to \eqref{eq: direct-connection matrix} or \eqref{eq: skip-connection matrix}.

\State Using \eqref{eq: Gaussian kernel} or any appropriate kernel function, such as the DM kernel in \eqref{eq: our discrete DM kernel}, and $\mbX$, form the gram matrix $\mbK(\epsilon)$.

\State Validate over $\hat{\mbu}_{\text{val}}$, using $\mbY_\mbW=\bm\Psi\mbW\mbY$ and $\widetilde{\mbK}_\mbW = \bm\Psi\mbW\mbK(\epsilon)(\bm\Psi\mbW)^\top$, according to the procedure outlined in \S\ref{subsec: validation} to compute the optimal parameter pair $(\epsilon_f,\lambda_f)$.

\State Compute $\hat{\bm{\alpha}}$ via \eqref{eq: weak H alpha hat}  using $(\epsilon_f,\lambda_f)$.

\State Perform out-of-sample forecasting via \eqref{eq: forecasting direct} or \eqref{eq: forecasting skip}.

\Ensure\parbox[t]{\dimexpr.9\linewidth-\algorithmicindent}{Optimal parameters $(\epsilon_f,\lambda_f)$ and coefficients $\hat{\bm\alpha}$ for forecasting via \eqref{eq: forecasting direct} or \eqref{eq: forecasting skip}.}
\end{algorithmic}
\end{algorithm}


\section{Numerical Study}\label{sec: examples}

In this section, we apply the proposed WKRR method to two baseline systems: the 3-dimensional Lorenz-63 (L63) system, and a 64-dimensional discretization of the Kuramoto-Sivashinsky (KS) system.
We compare the performance of the standard Gaussian kernel \eqref{eq: Gaussian kernel} with the DM kernel \eqref{eq: our discrete DM kernel}.
We also consider 15,000-dimensional experimental fluid data that corresponds to turbulent cavity flow (see Challenge 2.1 in \cite{schmidt2026data}).
When feasible, we compare WKRR with competitive baseline methods, such as strong KRR \cite{song2025learning}, RAFDA \cite{gottwald2021combining}, and LSTM \cite{hochreiter1997long}.

For the two baseline systems, we specify the signal-to-noise ratio of the given data \eqref{eq: data} by selecting intensity 
\begin{equation}\label{eq: noise intensity}
    \sigma \in \{0.01,\, 0.05,\, 0.1,\, 0.2\},
\end{equation}
which corresponds to 1\%, 5\%, 10\%, or 20\% noise corruption.

\subsection{Lorenz-63}

We first consider the Lorenz-63 (L63) dynamics, which are governed by the following equations \cite{lorenz2017deterministic}
\begin{equation}\label{eq: Lorenz 63 equations}
    \begin{split}
        x' &= \rho_1(y-x),
        \\
        y' &= x(\rho_2-z) - y,
        \\
        z' &= xy - \rho_3z.
    \end{split}
\end{equation}
Here, $\rho_1 = 10$, $\rho_2 = 28$, and $\rho_3=8/3$, and \eqref{eq: Lorenz 63 equations} admits chaotic dynamics with Lyapunov exponent $\Lambda \approx 0.91$ \cite{brugnago2020predicting, li2025weak, song2025learning}.

For data generation, we randomly generate an initial condition from the standard normal distribution.
We integrate using MATLAB's \texttt{ode113} syntax with relative error tolerance $10^{-10}$, absolute error tolerance $10^{-12}$, and timestep $\Delta t = 0.01$ to generate a trajectory consisting of $N=68,000$ steps.
Then, we corrupt the signal with a fixed intensity $\sigma$ from \eqref{eq: noise intensity}.
The first $10,000$ steps of the trajectory are pruned to eliminate transients and the resulting segment is divided into 2500 steps of noisy training data, 5500 steps of noisy validation data, and 50,000 of clean testing data.
We save only the clean testing data as ground truth to gauge predictive performance.
We repeat this procedure 100 times independently, which results in 100 segments consisting of training, validation, and testing data.

We then pre-process the training and validation data separately by either: (i) leaving it unfiltered, (ii) filtering with wavelets, or (iii) filtering with the polynomial test functions  $\{\varphi_j(t):=\varphi(t -jh)\}$ as defined in \eqref{eq: poly test function}.
To filter with wavelets, we use MATLAB's \texttt{wdenoise} syntax with a \texttt{sym12} wavelet.
To filter with polynomials, we reconstruct the noisy signal via \eqref{eq: weak reconstruction formula} using test functions generated from \eqref{eq: poly test function}.
The polynomial parameters $(p,L,h)$ are fixed for each noise intensity.
Appendix \ref{appendix: test function parameters} contains diagnostics which empirically show that the selected parameters act as effective filters, justifying their use in this step.
After filtering, we prune the first and last 250 steps to avoid boundary artifacts arising from the polynomial reconstruction.
For each noise intensity and each filtering procedure, this process results in 100 segments consisting of 2000 steps of filtering training data, 5000 steps of filtered validation data, and 50,000 steps of clean testing data.

We train, validate, and test a distinct WKRR model over each of the 100 trajectories using a \textit{skip-connection} scheme (see \S\ref{sec: preliminaries}).
We show results obtained from both the Gaussian kernel \eqref{eq: Gaussian kernel} and the DM kernel \eqref{eq: our discrete DM kernel}.
For each model, validation is performed according to the process described in \S\ref{subsec: validation} using the VPT error metric with threshold $\gamma=0.3$, and hyperparameters $N_c=20$, $N_f=30$, and $\nu = 200$.
The reference parameters $(\epsilon^*,\lambda^*)$ are computed according to the heuristic described in Appendix \ref{appendix: validation heuristic}.
Typical validation landscapes are shown in Appendix \ref{appendix: validation landscape}.

\begin{figure}[htpb]
    \centering
    \begin{minipage}[c]{0.64\textwidth}
        \centering
        \includegraphics[width=\textwidth]{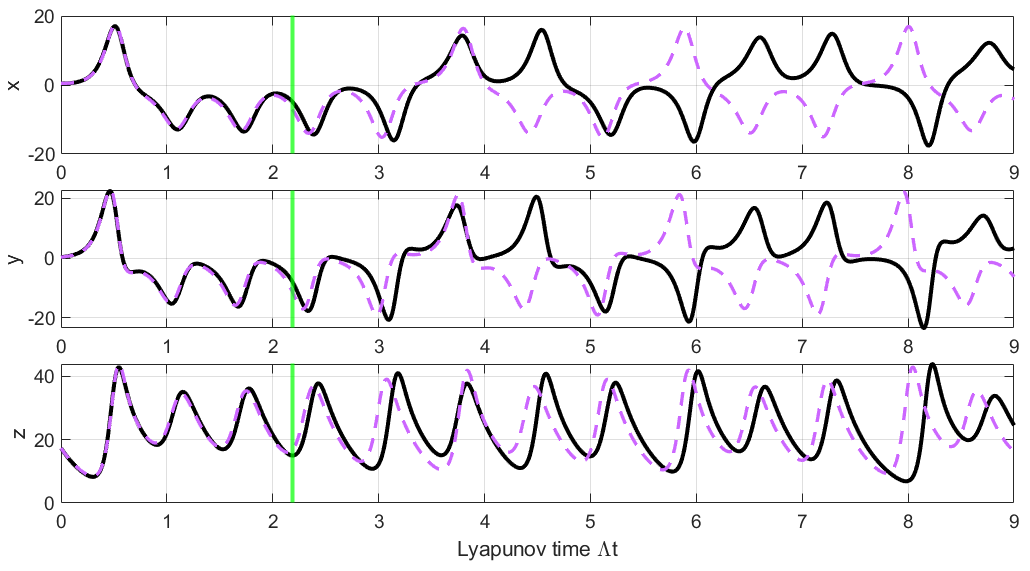}
    \end{minipage}
    \hfill
    \begin{minipage}[c]{0.33\textwidth}
        \centering
        \includegraphics[width=\textwidth]{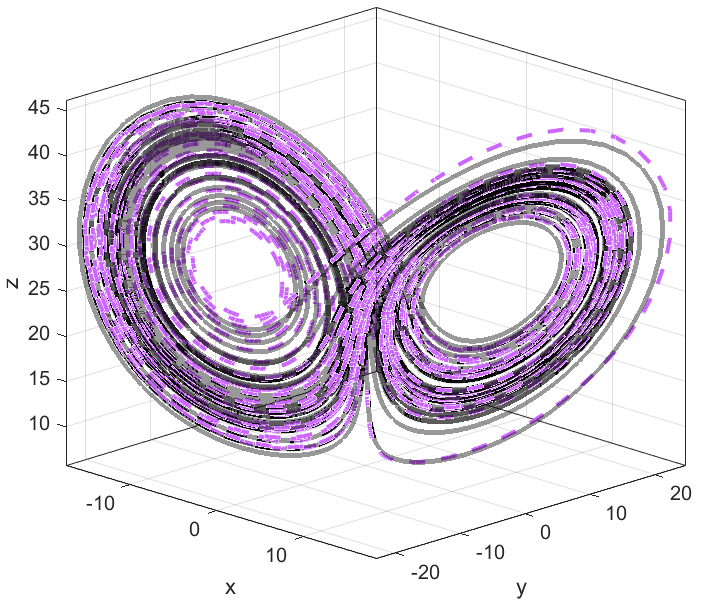}
    \end{minipage}
    \caption{\textbf{Left:} A comparison between ground truth (black) and a typical WKRR forecast (dashed purple) for the L63 system under 5\% noise corruption.
    The VPT, marked with a green line, is approximately $2.18$.
    \textbf{Right:} The butterfly attractor formed from the ground truth (black) and the WKRR reconstruction (dashed purple) over 5000 timesteps.}
    \label{fig: L63 typical forecast}
\end{figure}

For testing, we randomly select 500 integers in the interval $[1, \,47,000]$ to be used across all models.
We use these indices as starting points to extract 500 segments of length 3000 from each of the 100 testing trajectories.
For each model, we forecast over each of these 500 segments.
This process results in a mean VPT for each of the 100 models which are computed using 50,000 total VPTs.
We show a typical forecasting result obtained with the Gaussian kernel at 5\% noise in Figure \ref{fig: L63 typical forecast}.

We now compare the following frameworks across noise intensities: (i) strong KRR trained and validated on unfiltered data, (ii) RAFDA trained on unfiltered data, (iii) strong KRR trained and validated on wavelet filtered data, (iv) strong KRR trained and validated on polynomial filtered data, (v) proposed WKRR trained on unfiltered data and validated on polynomial filtered data, (vi) proposed WKRR trained and validated on polynomial filtered data, (vii) proposed WKRR with a DM kernel trained on unfiltered data and validated on polynomial filtered data, and (viii) proposed WKRR with a DM kernel trained and validated on polynomial filtered data.
Note that RAFDA does not receive filtered data due to its predictive mechanism.
Note that knowledge of the noise level $\sigma$ is required for RAFDA, but is not needed for either weak or strong KRR.
The case of strong KRR applied to unfiltered training and validation data is included as a baseline.
Each framework is tested over 100 models with identical data.

\begin{figure*}[htbp]
    \centering

    \begin{minipage}{0.47\textwidth}
        \centering
        \includegraphics[width=\textwidth]{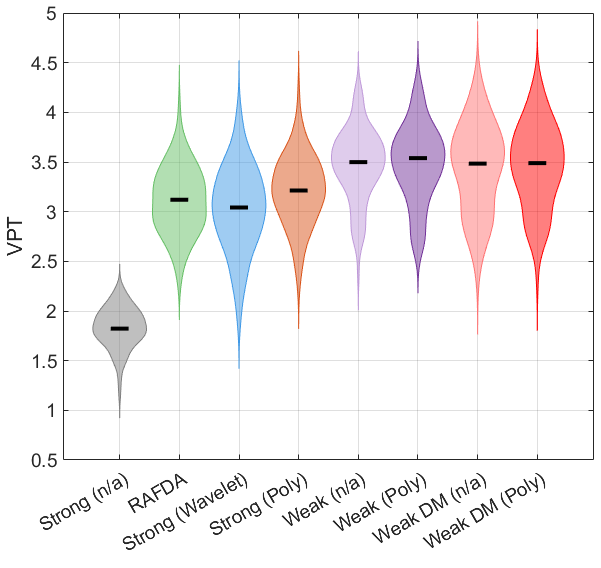}
        \small \textbf{(a) 1\% Noise}
    \end{minipage}
    \hfill
    \begin{minipage}{0.47\textwidth}
        \centering
        \includegraphics[width=\textwidth]{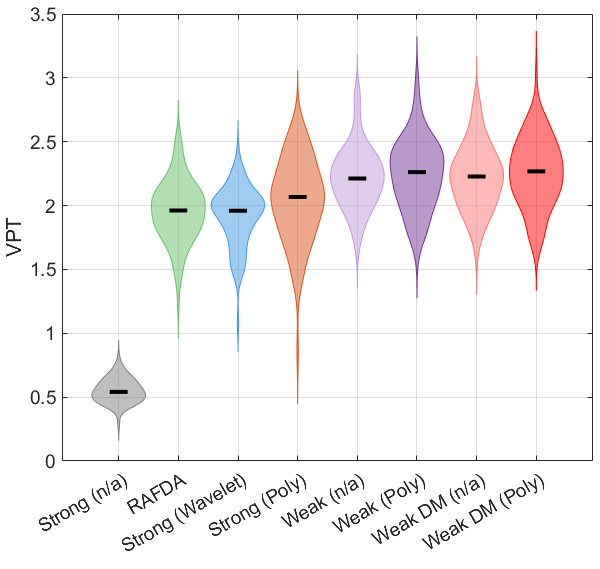}
        \small \textbf{(b) 5\% Noise}
    \end{minipage}
    
    \vspace{0.1cm}

    \begin{minipage}{0.47\textwidth}
        \centering
        \includegraphics[width=\textwidth]{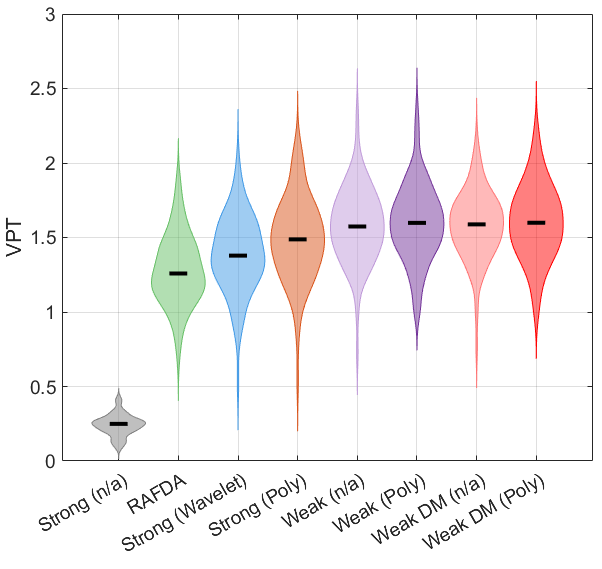}
        \small \textbf{(c) 10\% Noise}
    \end{minipage}
    \hfill
    \begin{minipage}{0.47\textwidth}
        \centering
        \includegraphics[width=\textwidth]{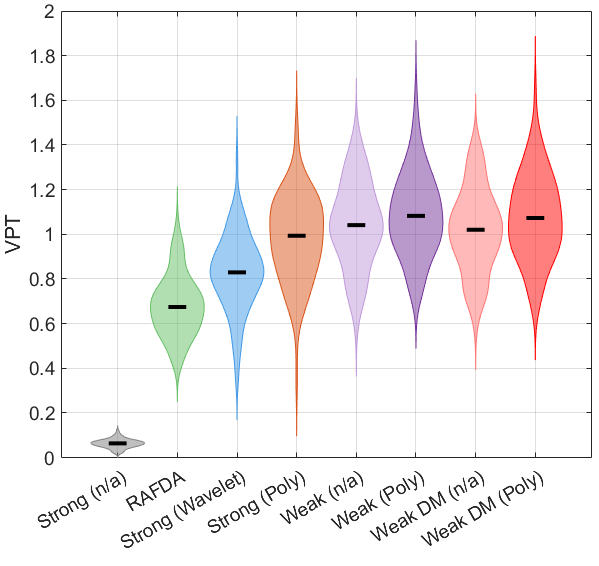}
        \small \textbf{(d) 20\% Noise}
    \end{minipage}

    \vspace{0.5cm} 
    
    \centering
    \small
    \setlength{\tabcolsep}{3.9pt}
    \begin{tabular}{l cccc cccc}
        \toprule
        & \multicolumn{4}{c}{\textbf{Mean VPT}} & \multicolumn{4}{c}{\textbf{Test Function Parameters} $(p, L, h)$} 
        \\
        \cmidrule(lr){2-5} \cmidrule(lr){6-9}
        \textbf{Method} & 1\% & 5\% & 10\% & 20\% & 1\% & 5\% & 10\% & 20\% 
        \\
        \midrule
        Strong (n/a)         
        &  $1.83 \pm 0.20$
        &  $0.55 \pm 0.09$
        &  $0.24 \pm 0.07$
        &  $0.06 \pm 0.02$
        & --- & --- & --- & --- 
        \\
        RAFDA                
        & $3.11 \pm 0.33$
        & $1.96 \pm 0.24$
        & $1.28 \pm 0.23$
        & $0.68 \pm 0.14$
        & --- & --- & --- & --- 
        \\
        Strong (Wavelet)     
        & $3.03 \pm 0.42$
        & $1.92 \pm 0.23$
        & $1.39 \pm 0.25$ 
        & $0.83 \pm 0.17$
        & --- & --- & --- & --- 
        \\
        Strong (Poly)        
        & $3.21 \pm 0.35$
        & $2.04 \pm 0.32$ 
        & $1.49 \pm 0.28$
        & $0.98 \pm 0.19$
        &($12, 0.6, 0.06$) 
        & ($5, 0.4, 0.08$) 
        & ($5, 0.5, 0.1$) 
        & ($5, 0.5, 0.1$)
        \\
        Weak (n/a)  
        &  $3.47 \pm 0.37$
        &  $2.22 \pm 0.25$
        &  $1.58 \pm 0.25$
        &  $1.04 \pm 0.19$
        & ($12, 0.6, 0.06$) 
        & ($5, 0.4, 0.08$) 
        & ($5, 0.5, 0.1$) 
        & ($5, 0.5, 0.1$)
        \\
        \textbf{Weak (Poly)} 
        & $\bm{3.48 \pm 0.39}$
        & $\bm{2.24 \pm 0.27}$
        & $\bm{1.60 \pm 0.26}$ 
        & $\bm{1.09 \pm 0.19}$ 
        & ($12, 0.6, 0.06$) 
        & ($5, 0.4, 0.08$) 
        & ($5, 0.5, 0.1$) 
        & ($5, 0.5, 0.1$) 
        \\
        Weak DM (n/a)  
        &  $3.43 \pm 0.44$
        &  $2.22 \pm 0.25$
        &  $1.58 \pm 0.23$
        &  $1.02 \pm 0.19$
        & ($12, 0.6, 0.06$) 
        & ($5, 0.4, 0.08$) 
        & ($5, 0.5, 0.1$) 
        & ($5, 0.5, 0.1$)
        \\
        \textbf{Weak DM (Poly)} 
        & $\bm{3.46 \pm 0.42}$
        & $\bm{2.26 \pm 0.27}$
        & $\bm{1.60 \pm 0.25}$
        & $\bm{1.08 \pm 0.19}$
        & ($12, 0.6, 0.06$) 
        & ($5, 0.4, 0.08$) 
        & ($5, 0.5, 0.1$) 
        & ($5, 0.5, 0.1$) 
        \\
        \bottomrule
    \end{tabular}

    \vspace{0.3cm}
    \caption{Empirical VPT densities for the L63 system \eqref{eq: Lorenz 63 equations} under various noise intensities. 
    Median VPT is marked as a horizontal black line. 
    ``Strong'' and ``Weak'' denote classical KRR and proposed WKRR frameworks, and parentheses indicate filtering applied to the training data, where (n/a) denotes unfiltered data. 
    Validation data is filtered as the training data for strong formulations, while polynomials are used for the weak formulations.
    The bottom table provides statistics across the 100 mean VPTs and selected test function parameters.}
    \label{fig: Lorenz violin}
\end{figure*}

Figure \ref{fig: Lorenz violin} reports the empirical mean VPT density for each of the above frameworks in the top four panels, and reports the corresponding statistics and test function parameters in the table below.
Across all noise levels, WKRR trained and validated on filtered data (Gaussian kernel - dark purple, DM kernel - dark red) consistently yields the best performance, while WKRR trained on unfiltered data (light purple and light red, respectively) is competitive or better than the strong approaches.
Notice that several VPT density tails for strong KRR applied to filtered data (e.g., Figure \ref{fig: Lorenz violin}, panel b) are quite long, indicating several models which produced VPTs well below the mean.
In contrast, WKRR did not produce such outliers in any of our numerical experiments.
This observation indicates that model validation may be more stable using a weak formulation.
RAFDA exhibits strong performance for small noise, but its predictive performance worsens as noise increases.

We observe that the Gaussian kernel and DM kernel give nearly identical results for the L63 system, which has an ambient dimension of three and an intrinsic dimension of roughly 2.06 on the buttefly attractor \cite{kuznetsov2020lorenz}.
This result suggests that the RBF kernel may be appropriate for low-dimensional systems, or for systems with ambient and intrinsic dimension roughly equal.

\begin{figure*}[htpb]
    \centering
    \begin{minipage}[b]{0.45\textwidth}
        \centering
        \includegraphics[width=\textwidth]{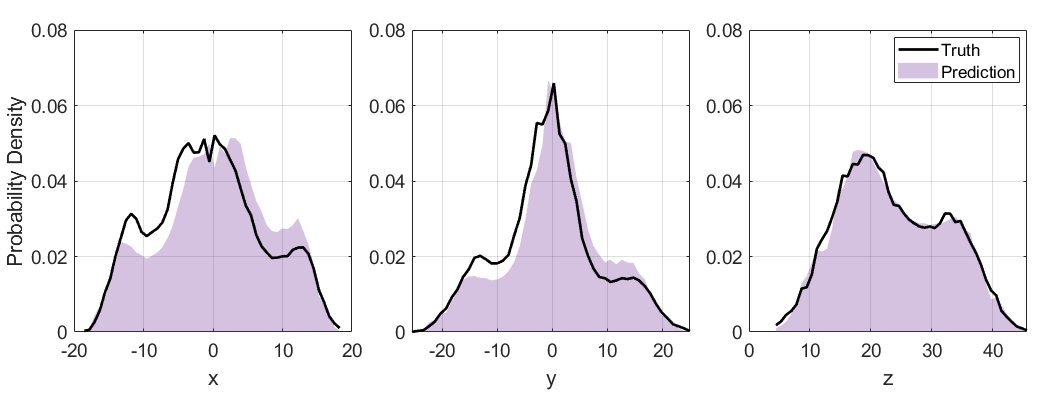} 
        \\
        \vspace{0.02cm}
        \includegraphics[width=\textwidth]{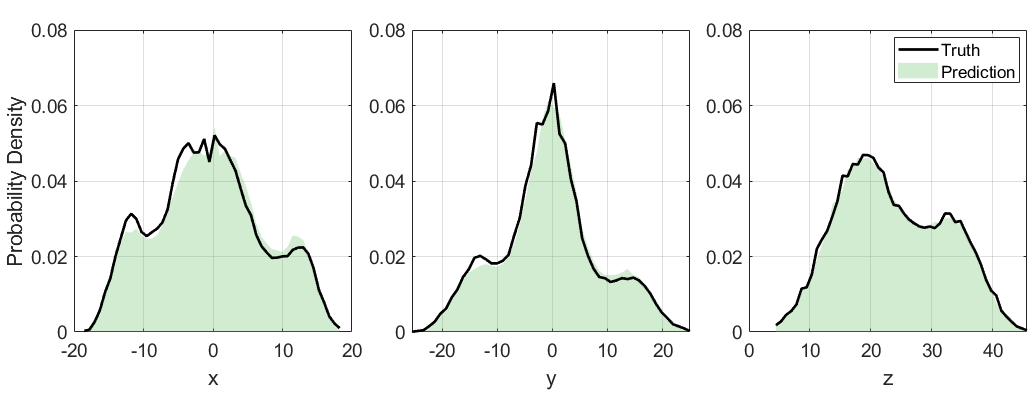} 
        \\
        \vspace{0.2cm}
        \small \textbf{(a) 1\% Noise}
    \end{minipage}
    \hfill
    \begin{minipage}[b]{0.45\textwidth}
        \centering
        \includegraphics[width=\textwidth]{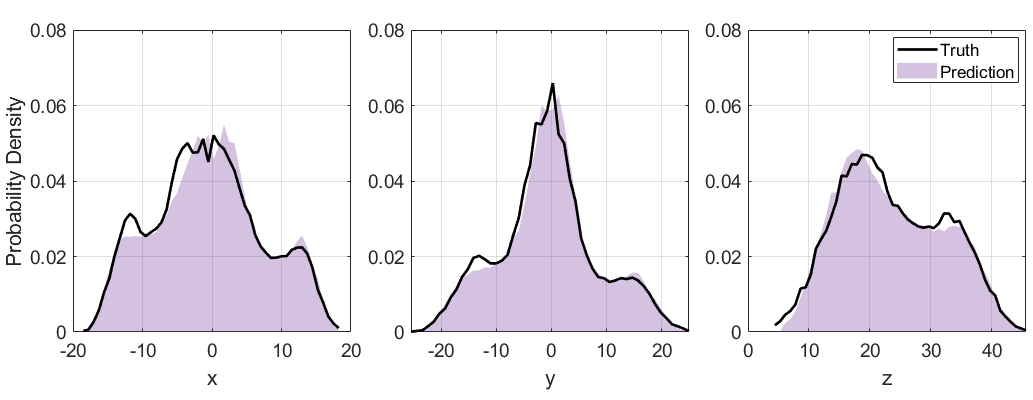} 
        \\
        \vspace{0.02cm}
        \includegraphics[width=\textwidth]{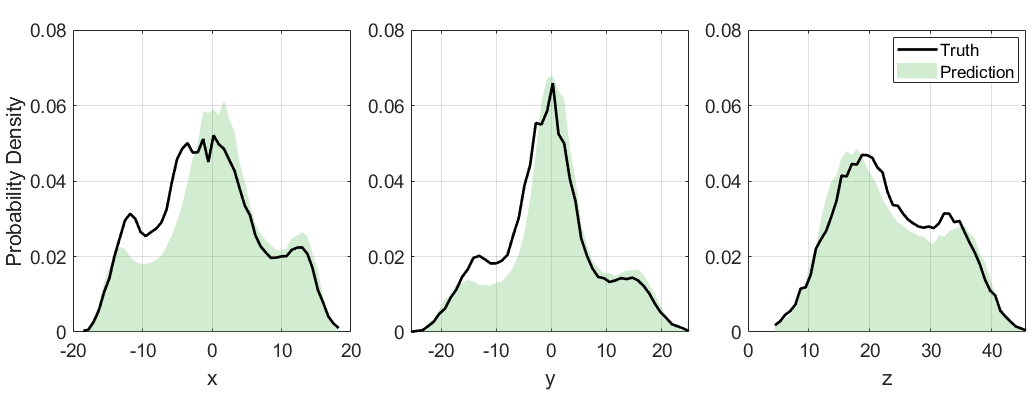} 
        \\
        \vspace{0.2cm}
        \small \textbf{(b) 5\% Noise}
    \end{minipage}
    
    \vspace{0.2cm}

    \begin{minipage}[b]{0.45\textwidth}
        \centering
        \includegraphics[width=\textwidth]{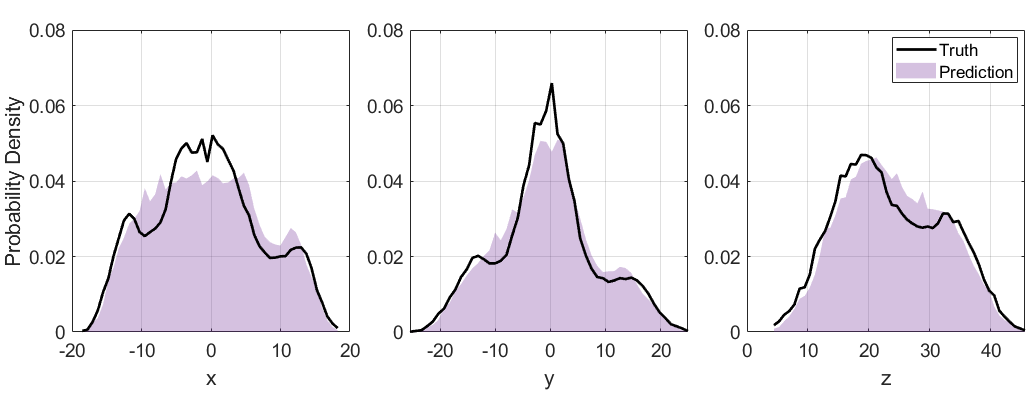} 
        \\
        \vspace{0.02cm}
        \includegraphics[width=\textwidth]{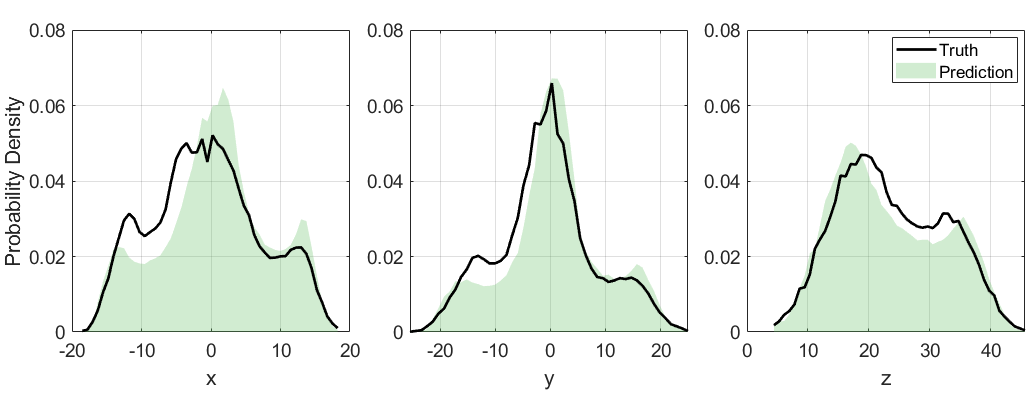} 
        \\
        \vspace{0.2cm}
        \small \textbf{(c) 10\% Noise}
    \end{minipage}
    \hfill
    \begin{minipage}[b]{0.45\textwidth}
        \centering
        \includegraphics[width=\textwidth]{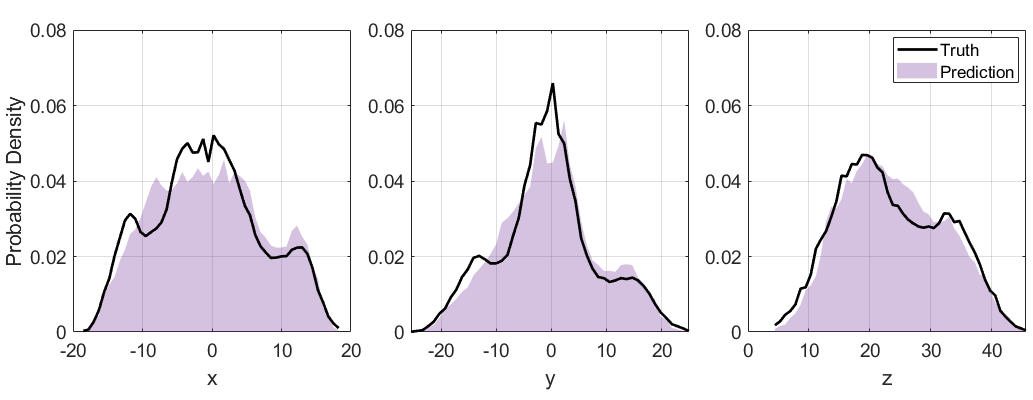} 
        \\
        \vspace{0.02cm}
        \includegraphics[width=\textwidth]{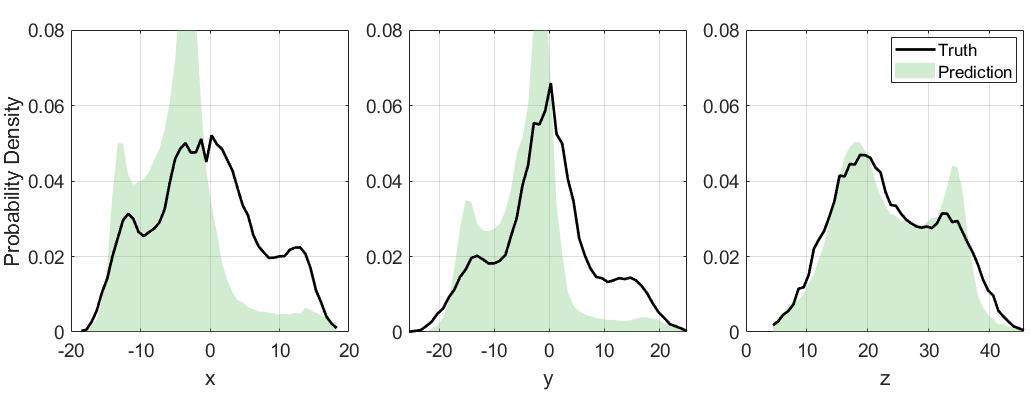} 
        \\
        \vspace{0.2cm}
        \small \textbf{(d) 20\% Noise}
    \end{minipage}

    \vspace{0.1cm}

    \caption{Invariant measure comparison for the learned Lorenz-63 system \eqref{eq: Lorenz 63 equations} under varying noise intensities comparing ground truth (black) to the proposed WKRR approach (purple) and RAFDA (green).
    }
    \label{fig: Lorenz tracking histos}
\end{figure*}

We remark that VPT measures short-term predictive capability, but does not capture long-term behavior.
To study long-term forecasting fidelity, we train WKRR and RAFDA on identical data and forecast over a segment of length 40,000.
We bin the forecasts and the ground truth in a histogram, which are shown in Figure \ref{fig: Lorenz tracking histos} for each noise intensity.
Both WKRR and RAFDA closely track the long-time marginal densities at small noise.
However, while WKRR recovers most qualitative features of marginal densities at high noise, the performance of RAFDA deteriorates.

\subsection{Kuramoto-Sivashinsky}\label{subsec: KS section}

We now apply WKRR to the Kuramoto-Sivashinsky (KS) system
\begin{equation}\label{eq: KS system}
    u_t + u u_x + u_{xx} + u_{xxxx} = 0, \quad x\in [0,J], \quad t\geq 0,
\end{equation}
with periodic boundary conditions $u(t,0)=u(t,J)$.
Following \cite{song2025learning}, we take $J=22$ and discretize \eqref{eq: KS system} in space over a uniform grid consisting of 64 points.
Accordingly, we view the KS system as a 64-dimensional ordinary differential equation.
In this setup, the KS system admits chaotic dynamics with Lyapunov exponent $\Lambda \approx 0.043$ and has Kaplan-Yorke dimension approximately $5.2$ \cite{edson2019lyapunov}, much smaller than the ambient 64-dimensional space.

To generate data we select $u(0,x) = \sin(16\pi x/J)$ as an initial condition, and integrate in time using the Exponential Time Differencing with RK4 (ETDRK4) method \cite{kassam2005fourth, song2025learning} with stepsize $\Delta t = 0.01$ for $N=3.05\cdot 10^7$ steps.
The data is then downsampled by a factor of 10, and the first 50,000 points are pruned to avoid transient behavior.
We then divide the remainder into 50 consecutive segments of 60,000 points.
Each of these 50 segments consists of 4500 steps of training data, 5500 steps of validation data, and 50,000 steps of testing data.
This data is corrupted with independent noise as in the L63 case.

The pre-processing step is performed identically to the L63 case, and  results in 50 segments consisting of 4000 steps of filtered training data, 5000 steps of filtered validation data, and 50,000 steps of clean testing data.

\begin{figure*}[htpb]
    \centering

    \begin{minipage}[b]{0.32\textwidth}
        \centering
        \includegraphics[width=\textwidth]{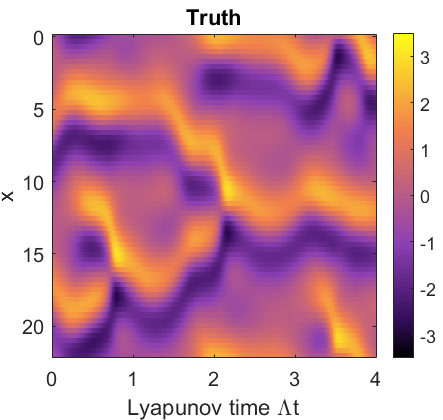}
    \end{minipage}
    \hfill
    \begin{minipage}[b]{0.32\textwidth}
        \centering
        \includegraphics[width=\textwidth]{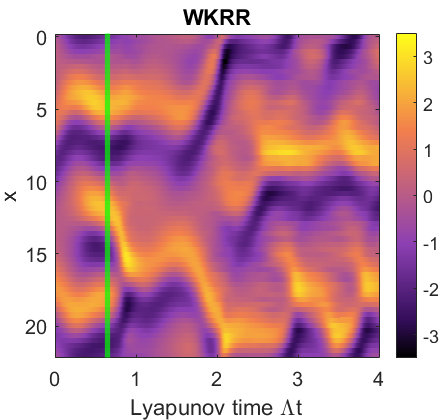}
    \end{minipage}
    \hfill
    \begin{minipage}[b]{0.32\textwidth}
        \centering
        \includegraphics[width=\textwidth]{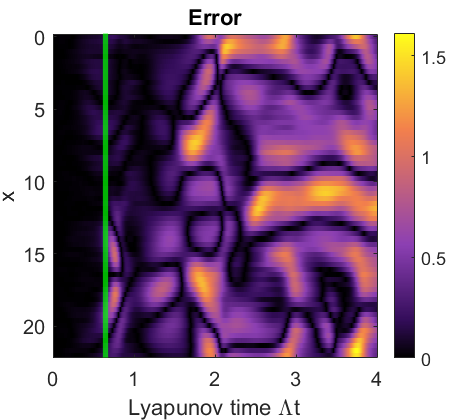}
    \end{minipage}
    \caption{Ground truth data (right) and a typical WKRR forecast (middle) with 5\% noise corruption.
    The error (right) is defined to be $E_i = |\mbu_i - \tilde\mbu_i|/\max|\mbu_i|$, where $\mbu_i$ is the ground truth and $\tilde\mbu_i$ is the WKRR prediction.
    The VPT is approximately $0.65$ and is marked with a green line.}
    \label{fig: KS typical forecast}
\end{figure*}

To validate and test, we employ a \textit{direct-connection} scheme (see \S\ref{sec: preliminaries}).
Otherwise, the validation process is similar to the L63 case.
We validate according to the procedure described in \S\ref{subsec: validation} using the VPT error metric \eqref{eq: VPT} with $\gamma=0.5$, and hyperparameters $N_c=20$, $N_f=30$, and $\nu=200$.
The reference parameters $(\epsilon^*,\lambda^*)$ were computed according to the method in Appendix \ref{appendix: validation heuristic}.
We scale $\lambda^*$ by a factor of 1000 for 5\%, 10\%, and 20\% noise to promote more stable validation.
Typical validation landscapes are shown in Appendix \ref{appendix: validation landscape}.
Testing was performed as in the L63 case, resulting in 50 mean VPTs across 50 WKRR models.
Figure \ref{fig: KS typical forecast} illustrates a typical forecast under 5\% noise with Gaussian kernel.

In our numerical experiments, we consider the same frameworks as for the L63 case, with the exception of RAFDA which was omitted due to prohibitively large computational time.

\begin{figure*}[htpb]
    \centering

    \begin{minipage}[b]{0.47\textwidth}
        \centering
        \includegraphics[width=\textwidth]{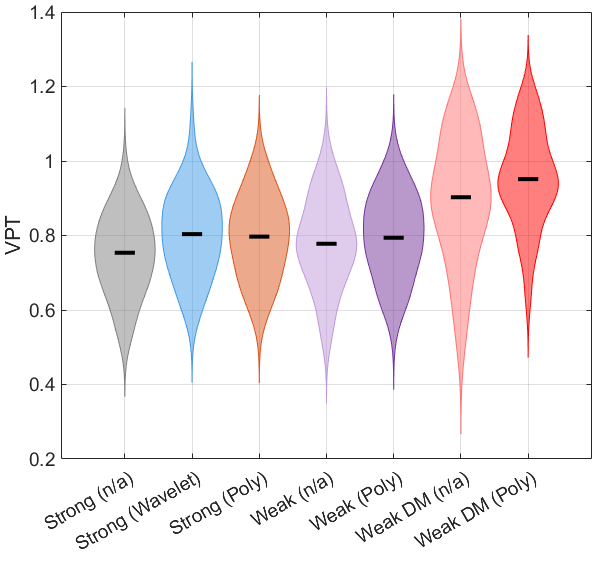}
        \small \textbf{(a) 1\% Noise}
    \end{minipage}
    \hfill
    \begin{minipage}[b]{0.47\textwidth}
        \centering
        \includegraphics[width=\textwidth]{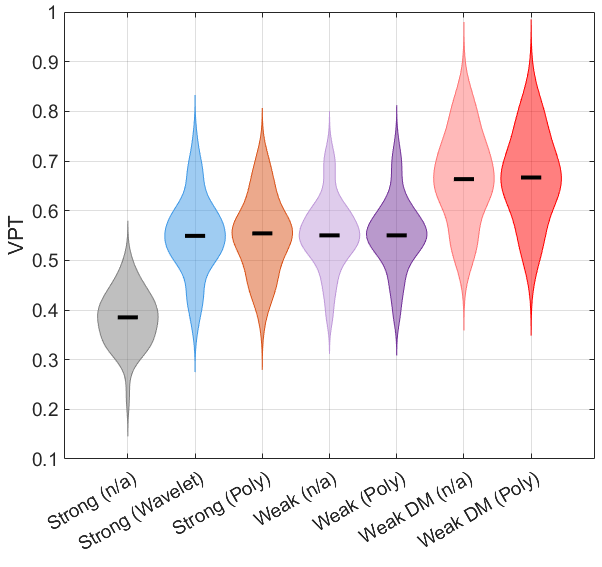}
        \small \textbf{(b) 5\% Noise}
    \end{minipage}
    
    \vspace{0.1cm}

    \begin{minipage}[b]{0.47\textwidth}
        \centering
        \includegraphics[width=\textwidth]{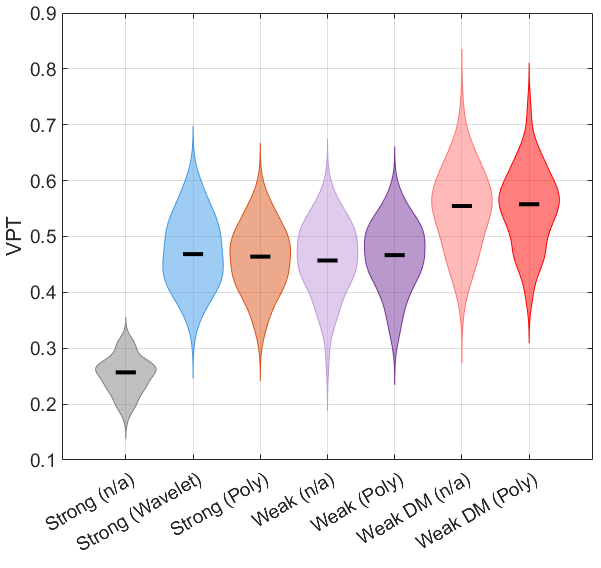}
        \small \textbf{(c) 10\% Noise}
    \end{minipage}
    \hfill
    \begin{minipage}[b]{0.47\textwidth}
        \centering
        \includegraphics[width=\textwidth]{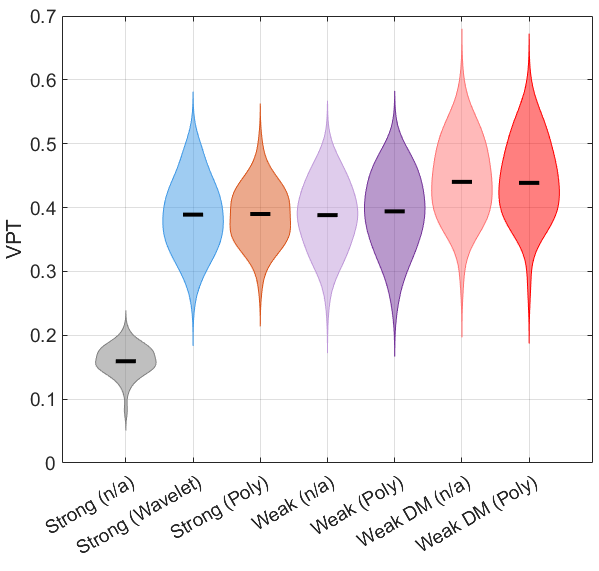}
        \small \textbf{(d) 20\% Noise}
    \end{minipage}

    \vspace{0.5cm}
    
    \centering
    \small
    \setlength{\tabcolsep}{5pt}
    \begin{tabular}{l cccc cccc}
        \toprule
        & \multicolumn{4}{c}{\textbf{Mean VPT}} & \multicolumn{4}{c}{\textbf{Test Function Parameters} $(p, L, h)$} 
        \\
        \cmidrule(lr){2-5} \cmidrule(lr){6-9}
        \textbf{Method} & 1\% & 5\% & 10\% & 20\% & 1\% & 5\% & 10\% & 20\% 
        \\
        \midrule
        Strong (n/a)       
        & $0.75 \pm 0.11$
        & $0.38 \pm 0.06$
        & $0.25 \pm 0.03$
        & $0.16 \pm 0.02$
        & --- & --- & --- & --- 
        \\
        Strong (Wavelet)     
        & $0.80 \pm 0.11$
        & $0.55 \pm 0.08$
        & $0.47 \pm 0.06$
        & $0.39 \pm 0.05$
        & --- & --- & --- & --- 
        \\
        Strong (Poly)        
        & $0.79 \pm 0.11$
        & $0.55 \pm 0.08$
        & $0.46 \pm 0.06$
        & $0.39 \pm 0.04$
        &($7, 5.5, 1.1$) 
        & ($5, 8, 1.6$) 
        & ($5, 9.5, 1.9$) 
        & ($6, 11.5, 2.3$)
        \\
        Weak (n/a)           
        & $0.78 \pm 0.12$
        & $0.55 \pm 0.07$
        & $0.46 \pm 0.06$
        & $0.39 \pm 0.05$
        & ($7, 5.5, 1.1$) 
        & ($5, 8, 1.6$) 
        & ($5, 9.5, 1.9$) 
        & ($6, 11.5, 2.3$)
        \\
        \textbf{Weak (Poly)} 
        & $\bm{0.80 \pm 0.11}$
        & $\bm{0.55 \pm 0.07}$
        & $\bm{0.47 \pm 0.06}$
        & $\bm{0.39 \pm 0.06}$
        & ($7, 5.5, 1.1$) 
        & ($5, 8, 1.6$) 
        & ($5, 9.5, 1.9$) 
        & ($6, 11.5, 2.3$) 
        \\
        Weak DM (n/a)           
        & $0.89 \pm 0.17$
        & $0.67 \pm 0.09$
        & $0.55 \pm 0.07$
        & $0.44 \pm 0.06$
        & ($7, 5.5, 1.1$) 
        & ($5, 8, 1.6$) 
        & ($5, 9.5, 1.9$) 
        & ($6, 11.5, 2.3$)
        \\
        \textbf{Weak DM (Poly)} 
        & $\bm{0.95 \pm 0.13}$
        & $\bm{0.67 \pm 0.09}$
        & $\bm{0.55 \pm 0.07}$
        & $\bm{0.44 \pm 0.06}$
        & ($7, 5.5, 1.1$) 
        & ($5, 8, 1.6$) 
        & ($5, 9.5, 1.9$) 
        & ($6, 11.5, 2.3$) 
        \\
        \bottomrule
    \end{tabular}

    \vspace{0.3cm}
    \caption{Empirical VPT densities for the KS system \eqref{eq: KS system} under various noise intensities. Median VPT is marked as a horizontal black line. ``Strong'' and ``Weak'' denote classical KRR and proposed WKRR frameworks, and parentheses indicate filtering applied to the training data, where (n/a) denotes unfiltered data. 
    Validation data is filtered as the training data for strong formulations, while polynomials are used for the weak formulations.
    The bottom table provides statistics across the 50 mean VPTs and selected test function parameters.}
    \label{fig: KS violin}
\end{figure*}

Figure \ref{fig: KS violin} depicts violin plots of the mean VPT densities across each of the frameworks and noise intensities.
The table below provides the corresponding statistics and test function parameters.
In all cases, WKRR with a Gaussian kernel applied to filtered data (dark purple) achieved comparable performance to the strong approaches.
However, WKRR with a DM kernel applied to filtered data (dark red), achieved superior performance across noise levels.
This result suggests that the DM kernel may be more appropriate for this problem. 
We suspect that the DM kernel exhibits superior performance because the flow map of the KS system is rather smooth, relating to a recent finding \cite{harlim2026diffusion} that suggests the DM kernel is superior to the Gaussian kernel in identifying smooth labels, especially those spanned by lower-order eigenfunctions of the Laplace-Beltrami operator on a submanifold of $\BR^n$. 
For completeness, we provide results which show the superior performance of WKRR over KRR using the DM kernel in Appendix \ref{appendix: KS DM data}.

\subsection{Validation Length Study}

The validation procedure forms a substantial part of the WKRR framework.
In this section, we perform a sensitivity analysis with respect to the validation segment length $\nu$.
Our goal is to demonstrate that the proposed method is robust across several validation strategies and does not require specific parameters to achieve good performance.

Our previous experiments averaged $N_c=20$ trajectories consisting of $\nu=200$ points over a coarse parameter grid, and subsequently averaged $N_f=30$ trajectories consisting of $\nu=200$ points over a fine parameter grid.
This modeling choice provided consistent forecasting performance across noise levels for both the L63 and KS systems.

Here, we compare this baseline validation strategy to four others: (i) $N_c=80$, $N_f=120$, and $\nu=50$; (ii) $N_c = 40$, $N_f=60$, and $\nu=100$; (iii) $N_c=10$, $N_f=15$, and $\nu=400$; and (iv) $N_c=8$, $N_F=12$, and $\nu=500$.
Otherwise, the experimental procedure is identical to the baseline cases.
These frameworks were considered to provide validation segments whose lengths are both larger and smaller than the expected model forecast.
To enact a direct comparison, we define $\nu_{\text{Lyap}} = \nu \cdot \Delta t \cdot \Lambda$, which has VPT units and may be directly compared to mean VPT. 
Here, we only show results with the Gaussian kernel since the same conclusion is also valid for the DM kernel.

\begin{figure}[htbp]
    \centering

    \begin{minipage}{0.49\textwidth}
        \centering
        \includegraphics[width=0.48\textwidth]{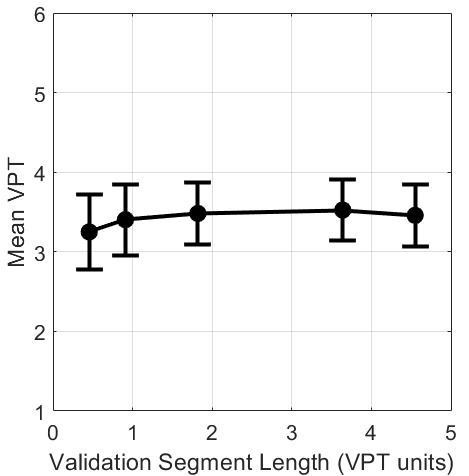}\hfill
        \includegraphics[width=0.48\textwidth]{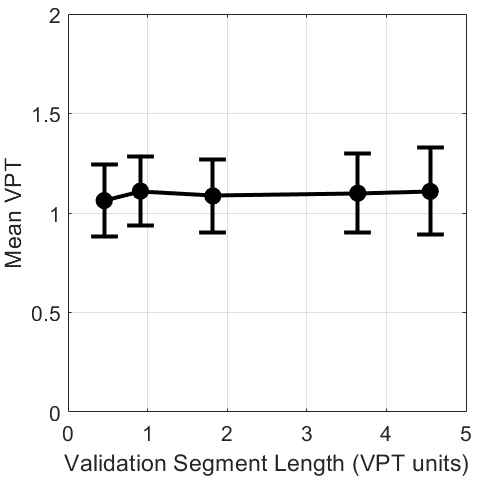}
        \vskip 1ex
        {\small \textbf{(a) Lorenz-63: left: 1\% noise, right: 20\% noise}}
    \end{minipage}\hfill
    \begin{minipage}{0.49\textwidth}
        \centering
        \includegraphics[width=0.48\textwidth]{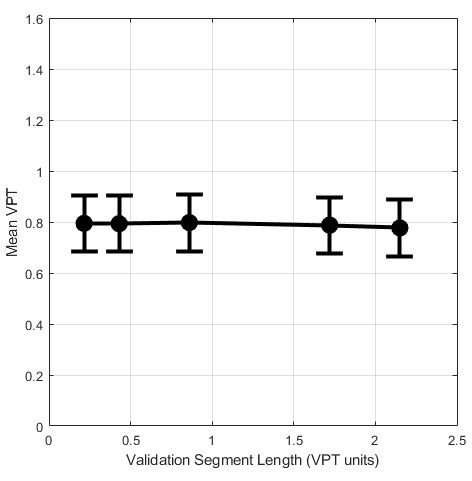}\hfill
        \includegraphics[width=0.48\textwidth]{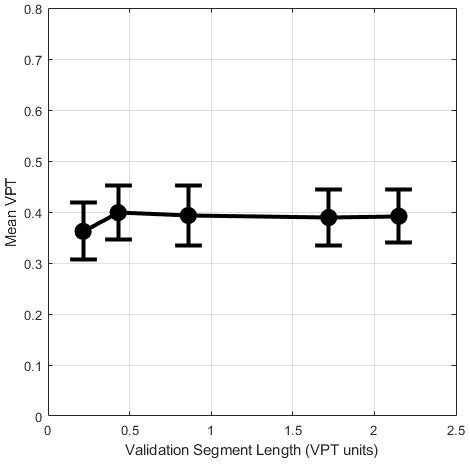}
        \vskip 1ex
        {\small \textbf{(b) Kuramoto-Sivashinsky: left: 1\% noise, right: 20\% noise}}
    \end{minipage}

    \vskip 3ex 

    \begin{minipage}[t]{0.49\textwidth}
        \centering
        \begin{tabular}{ccc}
            \toprule
            \textbf{Length $\nu_{\text{Lyap}}$} & \textbf{Mean VPT (1\%)} & \textbf{Mean VPT (20\%)} 
            \\
            \midrule
            0.46 & 
            3.25 $\pm$ 0.47 & 
            1.06 $\pm$ 0.18 
            \\
            0.91 & 
            3.40 $\pm$ 0.45 & 
            1.11 $\pm$ 0.17 
            \\
            1.82 & 
            3.48 $\pm$ 0.39 & 
            1.09 $\pm$ 0.19 
            \\
            3.64 & 
            3.52 $\pm$ 0.39 & 
            1.10 $\pm$ 0.20
            \\
            4.55 & 
            3.46 $\pm$ 0.39 & 
            1.11 $\pm$ 0.22 
            \\
            \bottomrule
        \end{tabular}
    \end{minipage}
    \hfill
    \begin{minipage}[t]{0.49\textwidth}
        \centering
        \begin{tabular}{ccc}
            \toprule
            \textbf{Length $\nu_{\text{Lyap}}$} & \textbf{Mean VPT (1\%)} & \textbf{Mean VPT (20\%)} 
            \\
            \midrule
            0.22 & 
            0.79 $\pm$ 0.11 & 
            0.36 $\pm$ 0.06 
            \\
            0.43 & 
            0.79 $\pm$ 0.11 & 
            0.40 $\pm$ 0.05 
            \\
            0.86 &  
            0.80 $\pm$ 0.11 & 
            0.39 $\pm$ 0.06 
            \\
            1.72 & 
            0.79 $\pm$ 0.11 & 
            0.39 $\pm$ 0.05 
            \\
            2.15 &  
            0.78 $\pm$ 0.11 & 
            0.39 $\pm$ 0.05 
            \\
            \bottomrule
        \end{tabular}
    \end{minipage}

    \vskip 1ex
    \caption{Mean VPT statistics as a function of validation length $\nu$ for \textbf{(a)} the L63 system and \textbf{(b)} the KS system at two noise levels.
    The tables present the quantitative values.
    The first column lists the validation length in Lyapunov time, $\nu_{\text{Lyap}}=\nu\cdot \Delta t \cdot \Lambda$.}
    \label{fig: validation study}
\end{figure}

Figure \ref{fig: validation study} (a) reports the mean and standard deviation of the mean VPTs at 1\% noise (left) and 20\% noise (right) over 100 models for the L63 system, while panel (b) reports these statistics over 50 models for the KS system.
We remark that the mean VPT remains stable across the validation strategies, highlighting the robustness of the proposed validation procedure.
We emphasize that comparable VPTs were achieved from validation segments which are both larger and smaller than the expected testing VPT.
While validating on segments whose length exceeds the expected forecast horizon is a typical approach to support appropriate generalization, our results suggest that WKRR still achieves comparable generalization even if it uses shorter validation segments.

\subsection{Clean Data Performance}

In this section, we use WKRR and classical strong KRR to study the L63 system \eqref{eq: Lorenz 63 equations} and the KS system \eqref{eq: KS system} as above, but with clean $(\sigma = 0)$ training and validation data.
The training and validation data are not pre-processed, but are otherwise constructed identically to the above cases.

The validation scheme used here differs from that described in \S\ref{subsec: validation} only in that we now utilize $N_c=N_f=3$ (possibly overlapping) validation segments consisting of $\nu=1500$ points, following the setup in \cite{song2025learning}.
Validating over longer segments improves performance because the expected forecast horizon under clean data is significantly longer than the noisy cases.
For WKRR, we scale the reference regularization parameter $\lambda^*$ by a factor of $10^{-6}$ to promote stable validation.

\begin{figure*}[htpb]
    \centering 

    \begin{minipage}[t]{0.39\textwidth}
        \centering
        \hspace*{-0.4cm}\includegraphics[width=\linewidth]{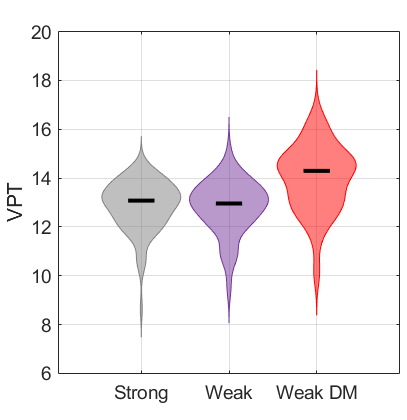}
        \vspace{0.4cm}
        {\small \textbf{(a) Lorenz-63 system}}
        \vspace{0cm}
        \small
        \begin{tabular}{l cc}
            \toprule
            \textbf{Method} & \textbf{Mean VPT} & \textbf{Parameters $(p, L, h)$} 
            \\
            \midrule
            Strong         
            & $12.87 \pm 1.02$ 
            & --- 
            \\
            Weak
            & $12.80 \pm 1.10$ 
            & ($14, 0.2, 0.02$) 
            \\
            Weak DM
            & $13.96 \pm 1.37$ 
            & ($14, 0.2, 0.02$) 
            \\
            \bottomrule
        \end{tabular}
    \end{minipage}
    \hspace{1.5cm}
    \begin{minipage}[t]{0.40\textwidth}
        \centering
        \hspace*{-0.4cm}\includegraphics[width=\linewidth]{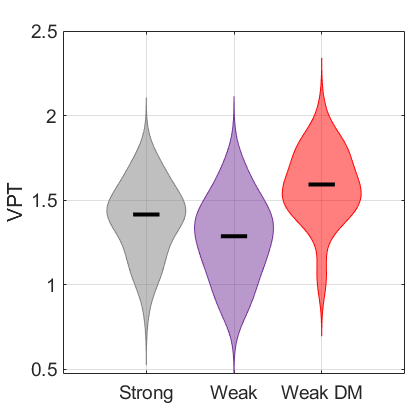}
        \vspace{0.4cm}
        {\small \textbf{(b) Kuramoto-Sivashinsky system}}
        \vspace{0cm}
        \small
        \begin{tabular}{l cc}
            \toprule
            \textbf{Method} & \textbf{Mean VPT} & \textbf{Parameters $(p, L, h)$} \\
            \midrule
            Strong         
            & $1.39 \pm 0.21$ 
            & --- 
            \\
            Weak
            & $1.28 \pm 0.24$ 
            & $(14, 1.5, 0.15)$ 
            \\
            Weak DM
            & $1.58 \pm 0.22$ 
            & $(14, 1.5, 0.15)$ 
            \\
            \bottomrule
        \end{tabular}
    \end{minipage}

    \vspace{0.2cm}
    \caption{Empirical VPT densities over clean data for \textbf{(a)} the L63 system \eqref{eq: Lorenz 63 equations} and \textbf{(b)} the KS system \eqref{eq: KS system}. 
    Median VPT is marked as a horizontal black line. 
    ``Strong'' and ``Weak'' denote classical KRR and proposed WKRR frameworks. 
    The bottom tables provide selected test function parameters and statistics across 100 mean VPTs for the L63 system and 50 mean VPTs for the KS system, respectively.}
    \label{fig: clean violin}
\end{figure*}

Figure \ref{fig: clean violin} reports the empirical VPT density over 100 models for the L63 system, and 50 models for the KS system, respectively.
The tables below provide corresponding quantitative statistics.
We remark that WKRR loses only a small amount of accuracy compared to the strong KRR approach, indicating its robustness across both clean and noisy data.
We remark that WKRR with a DM kernel exhibits superior performance for both systems.
Previous work showed that classic strong KRR with a DM kernel often exhibits superior performance compared to a Gaussian kernel over clean data \cite{song2025learning}.
The present results suggest the robustness of the DM kernel with a weak formulation.

One might argue that selecting appropriate test functions for WKRR make the method difficult to use in practice.
However, the selection is straightforward when clean data is available.
One procedure is as follows.
For a given family of test functions, reconstruct a portion of the given training data via \eqref{eq: weak reconstruction formula}.
Then, compare the available clean data to its reconstruction using a suitable error metric, and select the best-performing test function.

\subsection{Experimental Data: Fluid Dynamics}\label{subsec: example experimental data}

We now apply WKRR with both the Gaussian and DM kernel to experimental fluid data, made available by \cite{schmidt2026data} as a Community Challenge.
The data consists of streamwise velocity $\mbu$ and wall-normal velocity $\mbv$ measured on cavity flow. The dataset includes measurement noise as well as fluctuations due to the inherent turbulence in the flow.  Furthermore, the flow is strongly convective and heavily depends on the time-varying inlet boundary condition on the left, that is time-varying and not precisely known due to the experimental setup. 
Due to the unknown inlet boundary condition during prediction, it is challenging to achieve long forecasting horizon in this setting.
We refer to Problem~2.1 in \cite{schmidt2026data} for details of the experimental procedure and data collection process.

We set
\begin{equation}
    \mbU = \begin{bmatrix}
        \mbu & \mbv
    \end{bmatrix} \in \mathbb{R}^{N \times n},
\end{equation}
where $n=2\cdot 7854=15708$ is the spatial dimension and $N$ denotes the number of available data timeslices.
We are given $N_{\text{train}}=12,800$ samples as training data, and 32 segments consisting of $N_{\text{seg}}=70$ of validation data.
The original challenge as described in \cite{schmidt2026data} is to forecasting 30 timesteps past the validation data for each of the segments.
However, because this data is not available to us, we instead break apart the 32 segments into $N_{\text{val}}=40$ for model validation, and $N_{\text{test}}=30$ for model testing.
This procedure allows us to quantitatively analyze the performance of WKRR and compare it to other baseline methods.
To fix notation, let $\mbU_{\text{train}}=[\mbu_{\text{train}} \; \mbv_{\text{train}}]\in \mathbb{R}^{12800 \times 15708}$ denote the given training data, let $\mbU^{(q)}_{\text{val}}=[\mbu_{\text{val}} \; \mbv_{\text{val}}]\in \mathbb{R}^{40 \times 15708}$ denote the validation data, and let $\mbU^{(q)}_{\text{test}}=[\mbu_{\text{test}} \; \mbv_{\text{test}}]\in \mathbb{R}^{30 \times 15708}$ denote the testing data for $q=1,\dots,32$.

We do not directly apply WKRR to the given data, because its size makes computations prohibitively expensive.
Instead, we adopt the following pre-processing procedure.
Set $\mu_u, \mu_v \in \mathbb{R}^{1\times 7854}$ to be the time-average of $\mbu_{\text{train}}$ and $\mbv_{\text{train}}$, respectively.
Similarly set $\sigma_u,\sigma_v\in\mathbb{R}^{1\times 7854}$ to be the standard deviation of the training data.
Then for both training and validation data $\mbu,\mbv$, we normalize 
\begin{equation}
    \hat\mbU = \begin{bmatrix}
        \frac{\mbu-\mu_u}{\sigma_u} & \frac{\mbv-\mu_v}{\sigma_v}
    \end{bmatrix},
\end{equation}
where division is interpreted component-wise.
We then implement a POD procedure by taking a truncated SVD
\begin{equation}
    \hat\mbU = \mbU_r \mbS_r \mbV_r^\top, 
\end{equation}
where $r$ modes are retained.
The normalized data is then projected onto the POD modes
\begin{equation}
    \tilde\mbU = \hat\mbU \mbV_r = \begin{bmatrix}
        \tilde \mbu & \tilde \mbv
    \end{bmatrix}.
\end{equation}
Finally, we normalize per mode by writing
\begin{equation}
    \mbU^\dagger = \begin{bmatrix}
        \frac{\tilde\mbu-\mathfrak{u}_u}{\varsigma_u} & \frac{\tilde\mbv-\mathfrak{u}_v}{\varsigma_v}
    \end{bmatrix},
\end{equation}
where $\mathfrak{u}_u, \mathfrak{u}_v \in \mathbb{R}^{1\times r}$ and $\varsigma_u, \varsigma_v \in \mathbb{R}^{1\times r}$ are the mean and standard deviation of the POD modes that arise from the training data.
We emphasize that the mean and standard deviations are computed only from the training data, although these values are used to normalize both the training and validation data.

Polynomial parameters are chosen according to the following procedure.
We select a segment of training data consisting of 1000 timesteps.
The data is normalized according to the above procedure.
We then filter and reconstruct the normalized signals over a grid of parameter triples $(p,L,h)$.
The reconstruction is the unnormalized per-mode, projected onto original coordinates by the action of $V_r^\top$, and multiplied by the standard deviation.
The result is a mean-subtracted signal.
We then compare this signal to the ground truth (also mean-subtracted) using the NSME metric \eqref{eq: NMSE}, and select the parameter pair which minimizes the error.

We then apply WKRR to $\mbU^\dagger_{\text{train}}$.
Both the training and validation data is normalized according to the above procedure.
Validation is performed according to the procedure described in \S\ref{subsec: validation}, but with the following modifications.
The coarse grid is initialized to $H_c = [10^2,10^4]\times [10^{-16},10^{-13}]$, and NMSE (computed over the mode-normalized coefficients) is averaged over $N_c=3$ and $N_f=5$ segments of length $\nu = 15$ for the coarse and fine grids, respectively.
To alleviate computational time, we validate on only one of the 32 validation segments.\footnote{Several validation segments were randomly chosen and forecasting results did not change significantly, indicating that the particular choice of validation segment does not greatly alter the outcome.}
The resulting model is tested on all 32 testing segments.
This testing data is unnormalized to mean-subtracted physical coordinates and compared to the mean-subtracted ground truth using the NMSE metric.
This metric is consistent with that considered in the Community Challenge paper \cite{schmidt2026data}.

We compare the performance of WKRR with LSTM, used as a baseline in the Community Challenge paper \cite{schmidt2026data}.
The LSTM implementation, provided by \cite{schmidt2026data}, was modified to use the same normalized data as WKRR.
The baseline results in \cite{schmidt2026data} using LSTM were computed with $r=25$, noting that increasing $r$ may destabilize the training of the LSTM networks.
In the following, we will compare WKRR and LSTM with both $r=25$ and $r=100$.

\begin{figure}[htbp]
    \centering
    \includegraphics[width=0.99\linewidth]{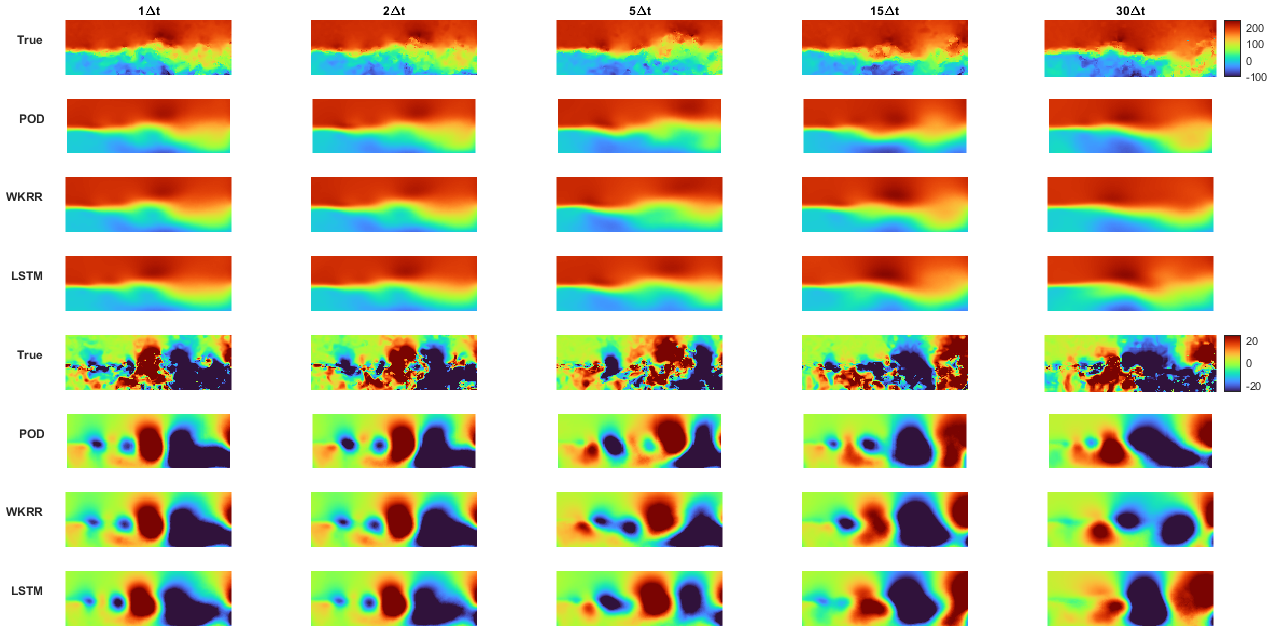}
    \caption{A visualization of the data and typical forecasts.
    The first four rows correspond to $\mbu$ data, and the bottom four rows correspond to $\mbv$ data. 
    From top to bottom, the rows in each block correspond to: (i) ground truth, (ii) POD representation of the data with $r=25$, (iii)  WKRR prediction, and (iv) LSTM prediction.}
    \label{fig: CC r 25}
\end{figure}

\begin{figure}[htbp]
    \centering
    \includegraphics[width=0.32\textwidth]{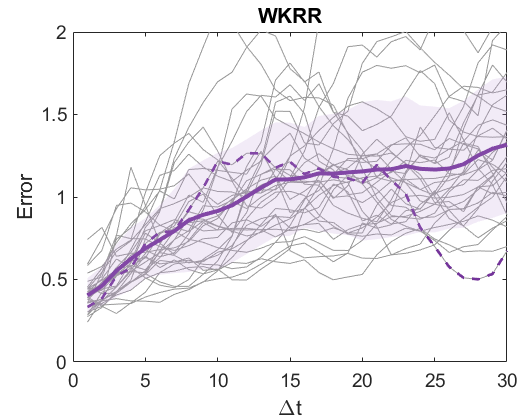} \hfill
    \includegraphics[width=0.32\textwidth]{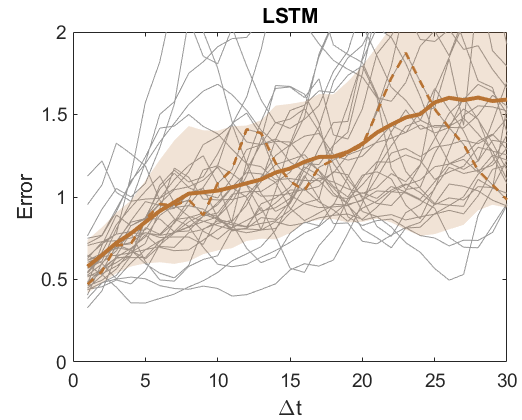} \hfill
    \includegraphics[width=0.32\textwidth]{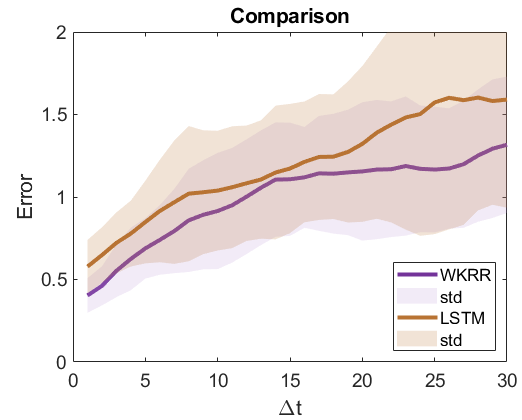}
    \caption{NMSE plots for WKRR (left), LSTM (middle), and a comparison of the two (right).
    Light gray lines correspond to each of the 32 trials, dashed colored lines correspond to the trajectory chosen for visualize in the top panel, solid colored lines correspond to mean NMSE over all 32 trials, and the colored shaded region corresponds to the standard deviation over all 32 trials.}
    \label{fig: cc r 25 error}
\end{figure}

Figure \ref{fig: CC r 25} shows a typical forecast for each method with $r=25$ over 30 timesteps.
We observe that WKRR with the Gaussian and DM kernel gives visually indistinguishable results, and so we report only results using the Gaussian kernel for this case.
The top block of four rows correspond to $\mbu$ data, while the last block of four rows corresponds to $\mbv$ data.
In each of the two blocks, we compare the ground truth (first row) with the $r=25$ POD representation (second row), the WKRR forecast (third row), and the LSTM forecast (fourth row) over one of the 32 testing segments.
The error plots in Figure \ref{fig: cc r 25 error} report the NMSE for each of the 32 trials (light gray), the average error over all 32 trials (solid colored line), and the corresponding standard deviation (light shaded band).
The dashed colored lines correspond to the trajectory represented in the two blocks.

Both methods produce predictions that reasonably reproduce the POD representation of the data.
Our results indicate that both methods are competitive in this regime.

\begin{figure}[htbp]
    \centering
    \includegraphics[width=0.97\linewidth]{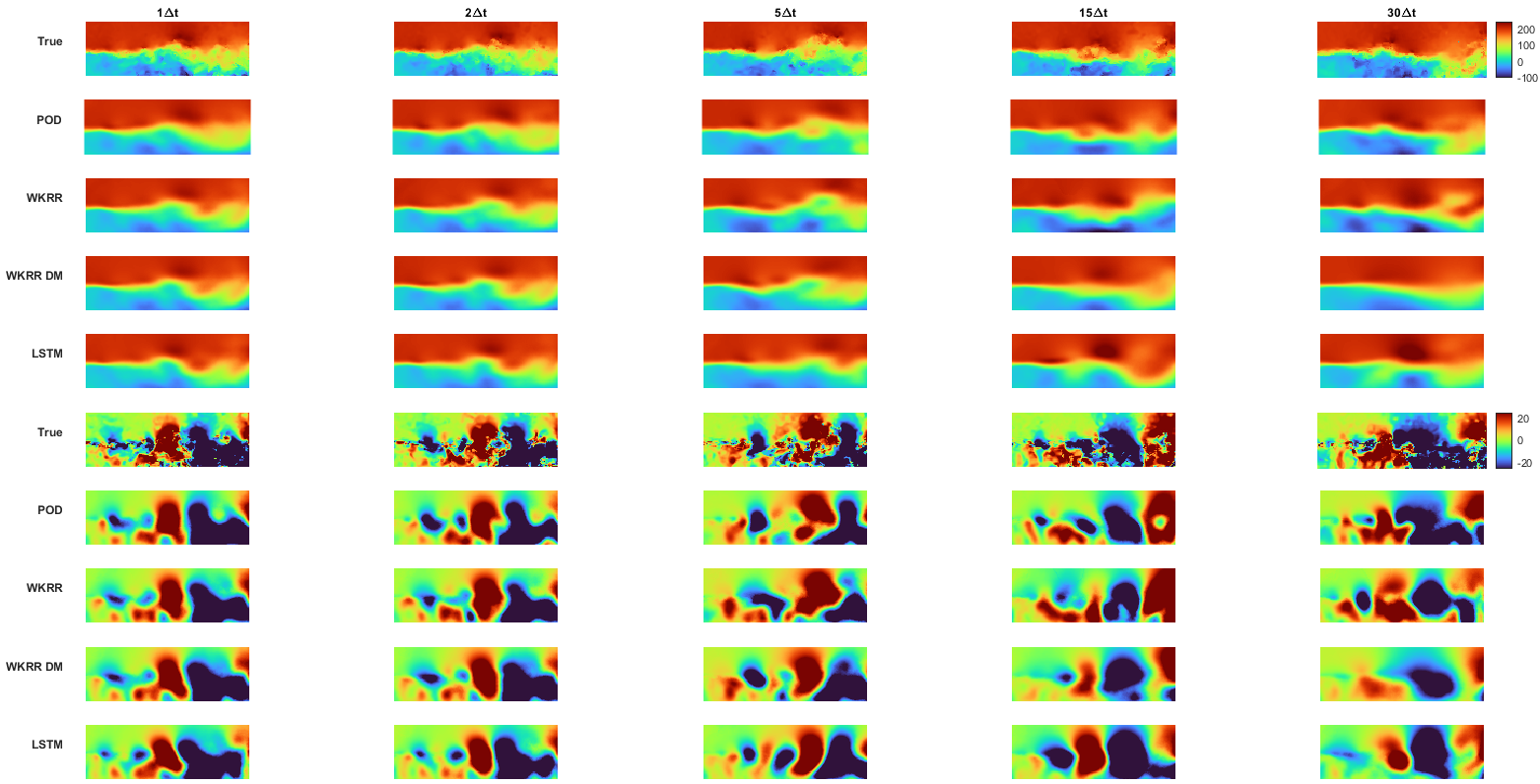}
    \caption{Typical data and forecast.
    The first five rows correspond to $\mbu$ data, and the bottom five rows correspond to $\mbv$ data. 
    From top to bottom, the rows in each block correspond to: (i) ground truth, (ii) POD representation of the data with $r=100$, (iii)  WKRR with a Gaussian kernel, (iv) WKRR with a DM kernel, and (v) LSTM prediction.}
    \label{fig: cc r 100}
\end{figure}

\begin{figure}[htbp]
    \centering
    \includegraphics[width=0.32\textwidth]{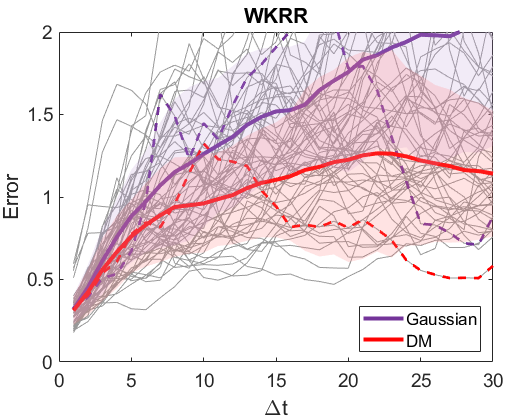} \hfill
    \includegraphics[width=0.32\textwidth]{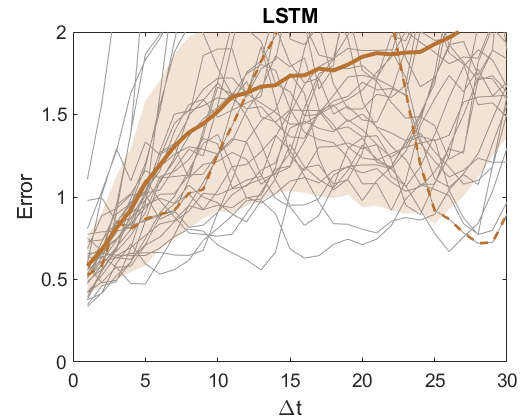} \hfill
    \includegraphics[width=0.32\textwidth]{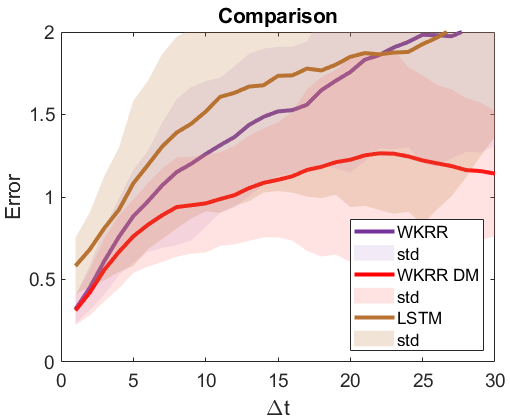}
    \caption{NMSE plots for WKRR (left), LSTM (middle), and a comparison of the two (right).
    Light gray lines correspond to each of the 32 trials, dashed colored lines correspond to the trajectory chosen for visualize in the top panel, solid colored lines correspond to mean NMSE over all 32 trials, and colored regions correspond to standard deviation over all 32 trials.
    We report results for WKRR with a Gaussian kernel (purple) and with a DM kernel (red).}
    \label{fig: cc r 100 error}
\end{figure}

We now repeat the forecasting procedure taking $r=100$ POD modes.
In this case, we report both the Gaussian and DM kernel results.
Figures \ref{fig: cc r 100} and \ref{fig: cc r 100 error} report the results in the same style as above.
The Gaussian kernel results are shown in purple, while the DM kernel results are shown in red.
We observe that WKRR with both kernels exhibits noticeably more consistent qualitative agreement and smaller mean error over short-term horizons.
This improved performance for higher-dimensional data highlights a key advantage of WKRR.

The performance of WKRR with the Gaussian and DM kernel differs in this regime.
Over longer horizons, WKRR with the DM kernel has noticeably smaller quantitative error.
However, this metric may be misleading.
Qualitatively, forecasts with the DM kernel tend to produce overly smooth predictions which may fail to capture the long-term statistical behavior of the dynamics.
In contrast, while predictions with the Gaussian kernel produce larger quantitative error, the forecasts appear to be qualitatively more consistent with the underlying dynamics over long time periods.
Over short time periods, the two kernels perform similarly, with a slight edge to the DM kernel.

\section{Discussion}\label{sec: discussion}

In this paper, we propose Weak-form Kernel Ridge Regression (WKRR) as a data-driven, noise-robust learning framework.
The proposed approach is computationally cheaper than classical strong-form KRR, and demonstrates competitive performance in the presence of noise over a range of chaotic, high-dimensional, and experimental data-sets.
While selecting appropriate model hyperparameters via validation is often a difficult task that may be sensitive to modeling assumptions, we perform sensitivity studies to demonstrate that WKRR can achieve robust performance with multiple strategies across several baseline systems and noise levels.
Furthermore, we show that with appropriately chosen test functions, WKRR applied to clean data greatly reduces computational complexity with only marginal loss in predictive performance.
This observation positions WKRR as a flexible framework across a range of datasets when the underlying noise level is unknown.
Finally, the success of kernel-based learning frameworks depends strongly on the choice of kernel function.
We demonstrate that WKRR achieves competitive performance with the standard Gaussian kernel over a range of baseline and experimental data, highlighting the method's simple implementation.
We also consider the Diffusion Maps (DM) kernel \cite{cl:2006,song2025learning} as an alternative choice, and show that the DM kernel can lead to increased predictive performance for systems whose invariant set dimension is much lower than the ambient dimension, even with noisy observational data.
The success of WKRR with multiple kernels broadens the applicability of the method and demonstrates that its noise robustness is not tied to a specific choice of kernel function.
Moreover, these findings suggest that the practitioner has a modeling choice: a Gaussian kernel reduces runtime and appears to be effective for low-dimensional systems, while a DM kernel may lead to increased forecast horizon for systems with complicated geometry. 
We emphasize that the proposed WKRR approach has advantages over existing learning methods which have incorporated a weak formulation \cite{bortz2024weak,li2025weak,messenger2021weak, messenger2025asymptotic, messenger2024weak} because it (i) does not require a choice of dictionary functions that span the target vector field, and (ii) does not require precise parameter tuning common to many machine learning architectures.

Despite the success of WKRR, several open questions remain.
We observed that model validation using WKRR often produced fewer outliers with poor forecast horizons than strong KRR over pre-processed data.
We hypothesize that validating over weak-form coefficients may lead to improved validation landscapes than using data in the original coordinates. 
It would be fruitful to pursue this observation, which may lead to more efficient validation strategies.

Selecting appropriate test functions is a challenge common to all weak-form learning approaches.
The bias-variance decomposition in \S\ref{subsec: bias variance decomp} suggests that using a smaller family of test functions reduces error due to noise.
In contrast, one generally expects that using a larger family of test functions reduces error due to signal reconstruction, although this analysis is complicated and typically depends on the specific choice of test function.
Developing a systematic approach to balance reconstruction and noise error constitutes a challenging yet worthwhile goal.

We demonstrated numerically that training over noisy observational data is significantly more effective with WKRR than with classical KRR.
While we observed that pre-processing training data often leads to marginally improved forecast horizons, the role of pre-processing in training and validation is not yet well-understood.
Incorporating multiple bandwidth parameters, as opposed to a single scalar considered in this work, may also influence training and validation.
Establishing theoretical convergence results for WKRR in these contexts would strengthen the proposed framework.

The present manuscript assumes that the given data captures all states of an underlying dynamical system.
However, in practice, given data may be only partially observed.
Additionally, many physical systems may depend on parameters or external non-autonomous forcings.
Extending WKRR to handle partially observed data, time-dependent data, or data which depends on parameters would broaden the applicability of WKRR.

\section*{Acknowledgment}

This work is partially supported under the NSF grants DMS-2505605 and CMMI-2340266, and the ICDS Penn State seed grant.

\section*{Data Availability}

The code to produce the figures in this paper is publicly available at \url{https://github.com/MaxKreider/WKRR}.
\appendix

\section{Test Function Parameters}\label{appendix: test function parameters}

In this section, we provide additional diagnostics for polynomial test functions as filters.
We also compare with wavelet pre-processing as a standard baseline filter.

In our numerical experiments in \S\ref{sec: examples}, we apply WKRR to both the L63 system \eqref{eq: Lorenz 63 equations} and the KS system \eqref{eq: KS system}.
In both cases, the training and validation data were corrupted with various noise intensities.
In Figures \ref{fig: Lorenz violin} and \ref{fig: KS violin}, we demonstrated that both strong and weak forms applied to polynomial filtered data resulted in higher mean VPT than the unfiltered case for a specific set of polynomial parameters $(p,L,h)$.
Here, we numerically demonstrate that the weak formulation with these parameters effectively filters the data, explaining the mechanism underlying the success of WKRR in these scenarios.

For each of the 100 L63 models or 50 KS models, we explicitly reconstruct the noise-corrupted training data with polynomial test functions via \eqref{eq: weak reconstruction formula}.
We also filter this training data with wavelets using MATLAB's \texttt{wdenoise} syntax with a \texttt{sym12} wavelet.
In both cases, we prune the first and last 250 data points to avoid boundary artifacts.
In this section, we denote the pruned ground truth data as $\mbu_i = [u_i^{(1)},\dots,u_i^{(n)}]^\top$ and the pruned filtered data as $\hat{\mbu}_i = [\hat{u}_i^{(1)},\dots,\hat{u}_i^{(n)}]^\top$ for $i=1,\dots,\mathfrak{N}$.
We measure the filter performance using three error metrics:
\begin{equation}\label{eq: three error metrics}
\begin{split}
    \mathcal{E}_{RMSE} &= \frac{1}{n}\sum_{\ell = 1}^n \sqrt{\frac{1}{\mathfrak{N}}\sum_{i=1}^{\mathfrak{N}} \left|u_i^{(\ell)} - \hat{u}_i^{(\ell)}\right|^2},
    \\ 
    \mathcal{E}_\theta &= \frac{1}{\mathfrak{N}}\sum_{i = 1}^\mathfrak{N} \arccos\left(\frac{\mbu_i \cdot \hat{\mbu}_i}{\|\mbu_i\|_2\|\hat{\mbu}_i\|_2}\right),
    \\
    \mathcal{E}_{LSD} &= \frac{1}{n}\sum_{\ell=1}^n \sqrt{\frac{1}{M} \sum_{m=1}^M \left|10\log_{10}\mathcal{P}[\mbu^{(\ell)}]_m - 10\log_{10}\mathcal{P}[\hat{\mbu}^{(\ell)}]_m\right|^2},
\end{split}
\end{equation}
where $\mathcal{P}[\cdot]_m$ is the $m$th component of the power spectral density of the signal, assumed to be of length $M$.

The metric $\mathcal{E}_{RMSE}$ computes RMSE over time, and averages these errors over spatial dimensions.
It provides a standard measure of pointwise error.
The metric $\mathcal{E}_{\theta}$ measures the angle between the truth and reconstruction at a fixed sample time, and averages the result over all available sample times.
Loosely speaking, it provides a measure of correctness of direction.
The metric $\mathcal{E}_{LSD}$ computes the log-spectra distance between the components of the true and reconstructed signals, and averages these errors over spatial dimension.
It provides a standard measure of spectral difference between two signals. 

\begin{figure}[htbp]
    \centering

    \begin{minipage}[b]{0.2\textwidth}
    \centering
        \includegraphics[width=\textwidth]{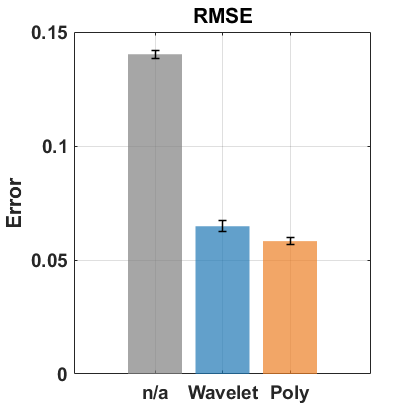}
    \end{minipage}
    \hfill
    \begin{minipage}[b]{0.2\textwidth}
    \centering
        \includegraphics[width=\textwidth]{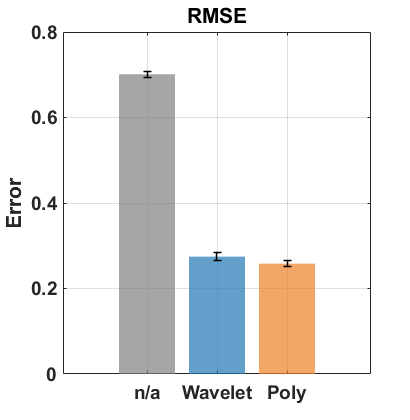}
    \end{minipage}
    \hfill
    \begin{minipage}[b]{0.2\textwidth}
    \centering
        \includegraphics[width=\textwidth]{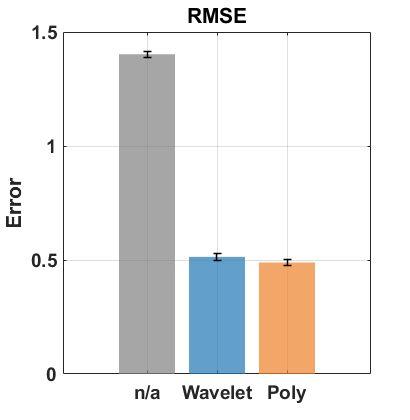}
    \end{minipage}
    \hfill
    \begin{minipage}[b]{0.2\textwidth}
    \centering
        \includegraphics[width=\textwidth]{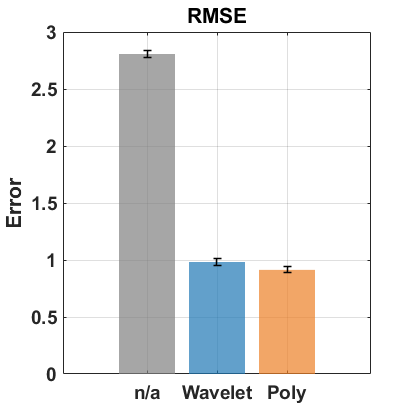}
    \end{minipage}

    \begin{minipage}[b]{0.2\textwidth}
    \centering
        \includegraphics[width=\textwidth]{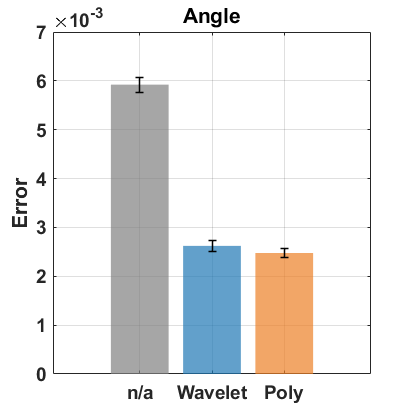}
    \end{minipage}
    \hfill
    \begin{minipage}[b]{0.2\textwidth}
    \centering
        \includegraphics[width=\textwidth]{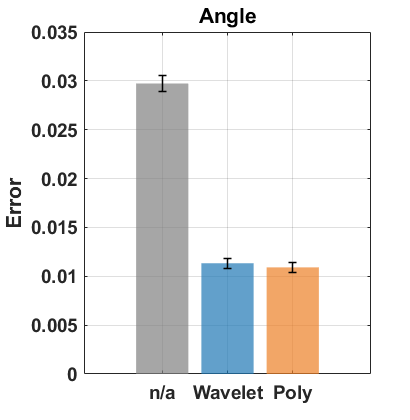}
    \end{minipage}
    \hfill
    \begin{minipage}[b]{0.2\textwidth}
    \centering
        \includegraphics[width=\textwidth]{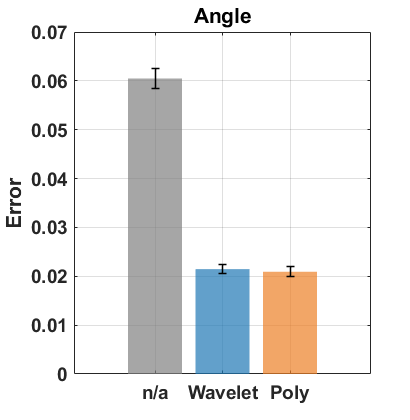}
    \end{minipage}
    \hfill
    \begin{minipage}[b]{0.2\textwidth}
    \centering
        \includegraphics[width=\textwidth]{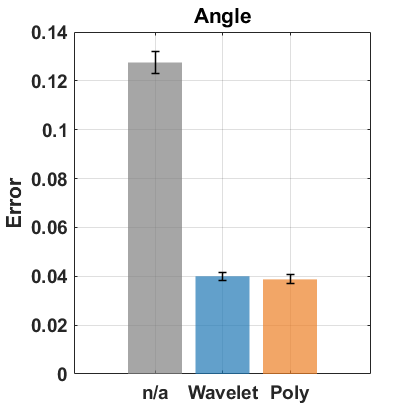}
    \end{minipage}

    \begin{minipage}[b]{0.2\textwidth}
    \centering
        \includegraphics[width=\textwidth]{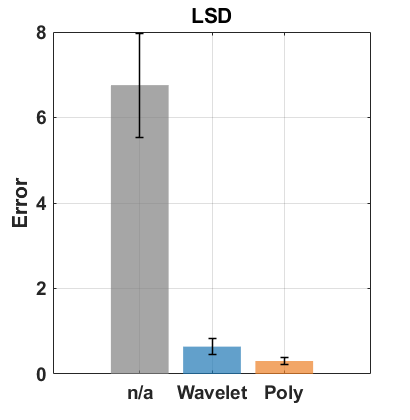}
        {\small \textbf{(a) 1\% noise}}
    \end{minipage}
    \hfill
    \begin{minipage}[b]{0.2\textwidth}
    \centering
        \includegraphics[width=\textwidth]{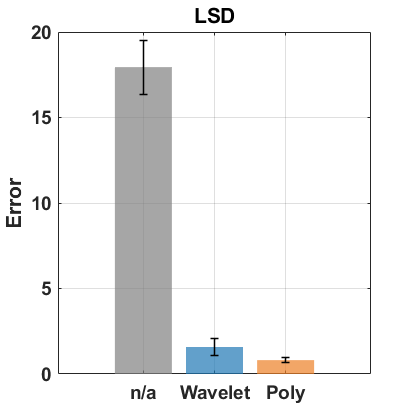}
        {\small \textbf{(b) 5\% noise}}
    \end{minipage}
    \hfill
    \begin{minipage}[b]{0.2\textwidth}
    \centering
        \includegraphics[width=\textwidth]{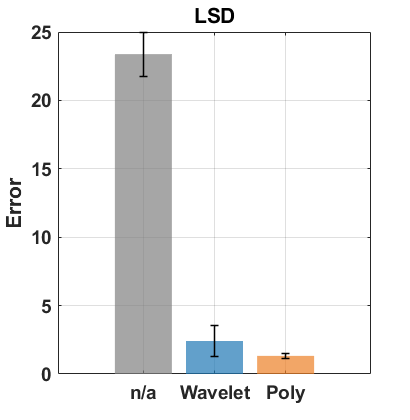}
        {\small \textbf{(c) 10\% noise}}
    \end{minipage}
    \hfill
    \begin{minipage}[b]{0.2\textwidth}
    \centering
        \includegraphics[width=\textwidth]{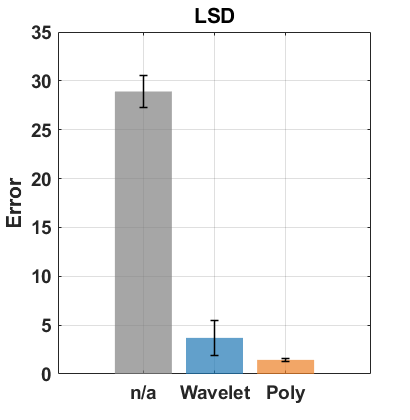}\
        {\small \textbf{(d) 20\% noise}}
    \end{minipage}
    
    \caption{A comparison of the filtering properties of wavelets (blue) and polynomial test functions (orange) with an unfiltered baseline (gray) for the L63 system over three error metrics: \textbf{Top row:} RMSE $\mathcal{E}_{RMSE}$, \textbf{Middle row:} Angle $\mathcal{E}_{\theta}$, and \textbf{Bottom row:} Log-spectral distance $\mathcal{E}_{LSD}$ \eqref{eq: three error metrics}.}
    \label{fig: L63 bar}
\end{figure}

\begin{figure}[htbp]
    \centering

    \begin{minipage}[b]{0.2\textwidth}
    \centering
        \includegraphics[width=\textwidth]{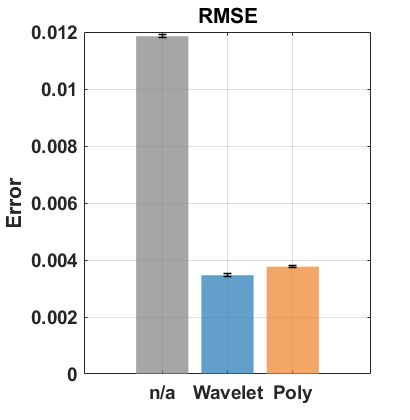}
    \end{minipage}
    \hfill
    \begin{minipage}[b]{0.2\textwidth}
    \centering
        \includegraphics[width=\textwidth]{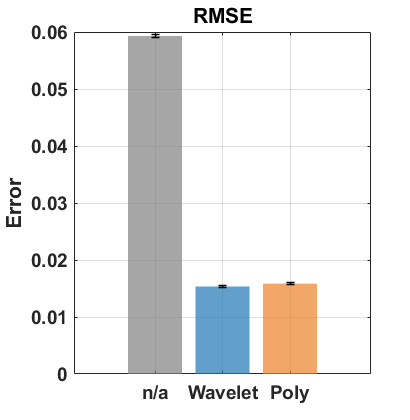}
    \end{minipage}
    \hfill
    \begin{minipage}[b]{0.2\textwidth}
    \centering
        \includegraphics[width=\textwidth]{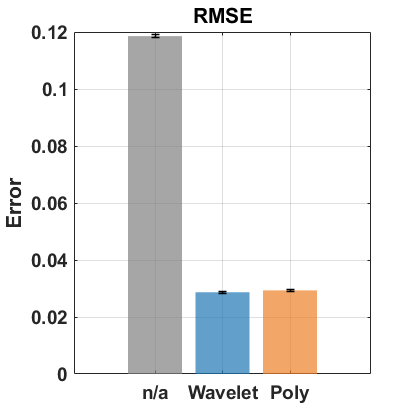}
    \end{minipage}
    \hfill
    \begin{minipage}[b]{0.2\textwidth}
    \centering
        \includegraphics[width=\textwidth]{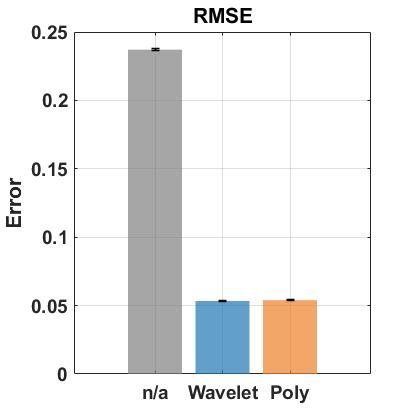}
    \end{minipage}

    \begin{minipage}[b]{0.2\textwidth}
    \centering
        \includegraphics[width=\textwidth]{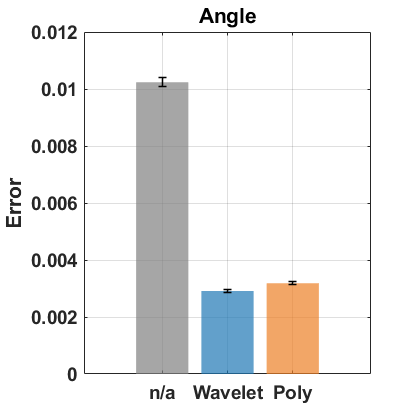}
    \end{minipage}
    \hfill
    \begin{minipage}[b]{0.2\textwidth}
    \centering
        \includegraphics[width=\textwidth]{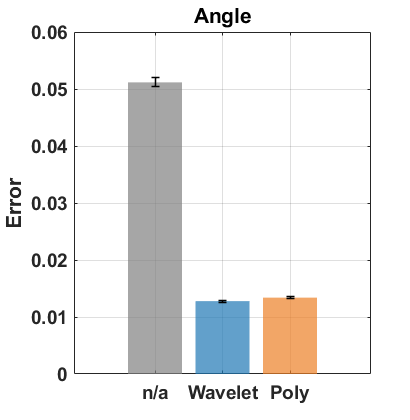}
    \end{minipage}
    \hfill
    \begin{minipage}[b]{0.2\textwidth}
    \centering
        \includegraphics[width=\textwidth]{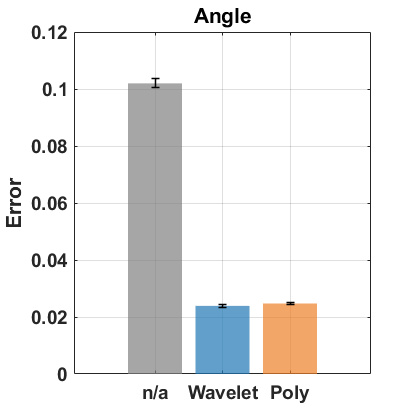}
    \end{minipage}
    \hfill
    \begin{minipage}[b]{0.2\textwidth}
    \centering
        \includegraphics[width=\textwidth]{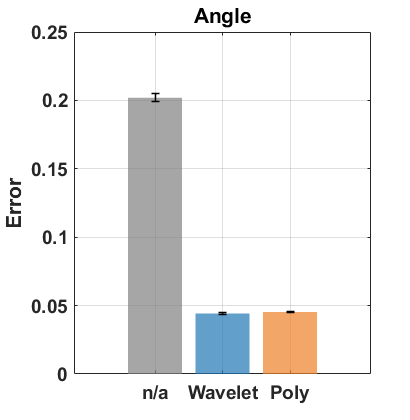}
    \end{minipage}

    \begin{minipage}[b]{0.2\textwidth}
    \centering
        \includegraphics[width=\textwidth]{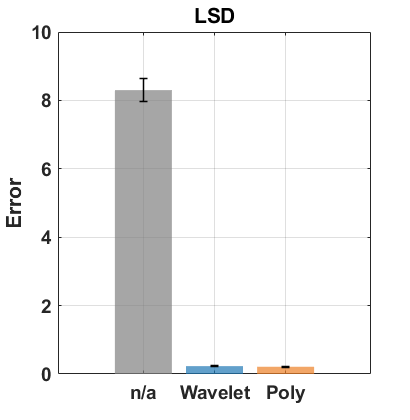}
        {\small \textbf{(a) 1\% noise}}
    \end{minipage}
    \hfill
    \begin{minipage}[b]{0.2\textwidth}
    \centering
        \includegraphics[width=\textwidth]{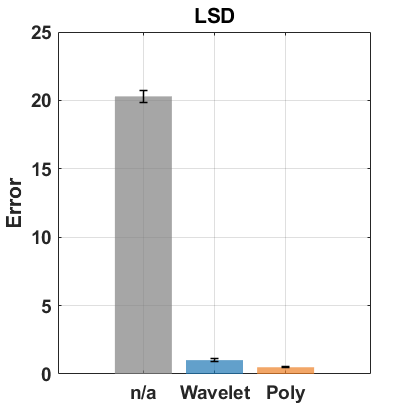}
        {\small \textbf{(b) 5\% noise}}
    \end{minipage}
    \hfill
    \begin{minipage}[b]{0.2\textwidth}
    \centering
        \includegraphics[width=\textwidth]{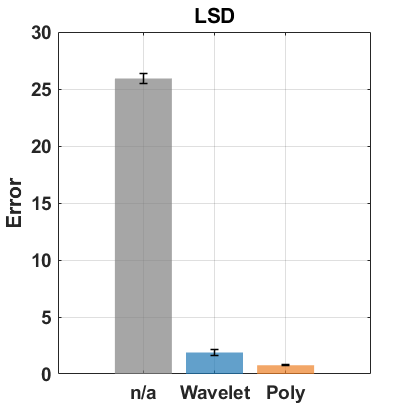}
        {\small \textbf{(c) 10\% noise}}
    \end{minipage}
    \hfill
    \begin{minipage}[b]{0.2\textwidth}
    \centering
        \includegraphics[width=\textwidth]{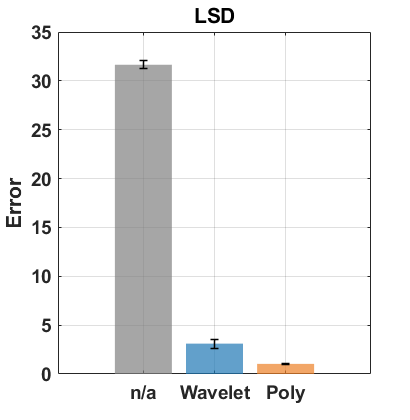}\
        {\small \textbf{(d) 20\% noise}}
    \end{minipage}
    
    \caption{A comparison of the filtering properties of wavelets (blue) and polynomial test functions (orange) with an unfiltered baseline (gray) for the KS system over three error metrics: \textbf{Top row:} RMSE $\mathcal{E}_{RMSE}$, \textbf{Middle row:} Angle $\mathcal{E}_{\theta}$, and \textbf{Bottom row:} Log-spectral distance $\mathcal{E}_{LSD}$ \eqref{eq: three error metrics}.}
    \label{fig: KS bar}
\end{figure}

We report the mean and standard deviation of these error metrics averaged over 100 trials for the L63 system in Figure \ref{fig: L63 bar}  and averaged over 50 trials for the KS system in Figure \ref{fig: KS bar}.
In all cases, the data $\hat{\mbu}$ is filtered with wavelets (blue), polynomial test functions (orange), or left unfiltered (gray).
Across noise levels, both wavelets and polynomials give comparable results in terms of RMSE and angle differences. The polynomial filter produces filtered data with smaller LSD errors.
The resulting error metrics are significantly improved relative to the unfiltered baseline, indicating that both wavelets and polynomials effectively filter the data.
In particular, the success of the polynomials in filtering the data justifies the $(p,L,h)$ parameter selection used in our numerical experiments.

\section{H Matrix}\label{appendix: bias derivation}

Suppose that data $\mbU = [\mbu_1,\dots,\mbu_N]^\top\in \mathbb{R}^{N\times n}$, with corresponding clean data $\mbU_{\text{clean}}$, is given.
We assume that the underlying clean data is periodic, which implies that $k^* = N/(h/\Delta t)$.
We also assume that $\mbW = \mbI$.

Our main goal in this section is to show that the general expression for the bias given in \eqref{eq: bias variance decomposition}, under the above ideal assumptions, can be expressed in the form
\begin{equation}\label{eq: bias goal}
    B_\ell^2 = \|(\mbP - \mbI)\mbU_{\text{clean}}^{(\ell)}\|_F^2 = \left\|(\mbH - \mbI)\mathcal{F}\left[\mbU_{\text{clean}}^{(\ell)}\right]\right\|_F^2,
\end{equation}
in terms of the Fourier transform of the $\ell$th component of the true underlying signal, and the matrix $\mbH\in \mathbb{R}^{N\times N}$ as defined in \eqref{eq: H}.
This approach is inspired by the approach taken in \cite{unser2002sampling}, which performs computations in continuous time.

We first observe that when $\mbW=\mbI$, we have that $\mbP = \bm\Psi^\top ( \bm\Psi\bm\Psi^\top)^{-1} \bm\Psi$.
Notice that the entries of the term in parenthesis are quadrature approximations of the integral
\begin{equation}
    \begin{split}
        [( \bm\Psi\bm\Psi^\top)]_{ij} &= \int_{\mathbb{R}} \td t\; \varphi_i(t) \varphi_j(t) =  \int_{\mathbb{R}} \td \tau\; \varphi(\tau) \varphi(\tau - (j-i)h),
    \end{split}
\end{equation}
which shows this matrix is Toeplitz, and leveraging periodicity, circulant.

This observation is important because the discrete Fourier Transform (DFT) diagonalizes circulant matrices.
Let $\mbF\in \mathbb{R}^{N\times N}$ with entries $\mbF_{mn} = 1/\sqrt{N}\exp(-2\pi m ni/N)$.
Notice that $\mbF^*\mbF = \mbI_{N\times N}$.
Consider the frequency domain representation of the test functions $\hat{\bm\Psi} = \mbF \bm\Psi^\top\in \mathbb{C}^{N\times k^*}$.
Leveraging periodicity and straightforward manipulations, one can show that the $(m,j)$ entry of this object has the form
\begin{equation}
    \begin{split}
        \hat{\bm\Psi}_{mj} &= \sum_{p=0}^{N-1} \frac{1}{\sqrt{N}} \exp(-2\pi m p i /N) \varphi[p-jh]
        = \exp(-2\pi m j i /k^*) \hat{\varphi}_m,
    \end{split}
\end{equation}
where $\hat{\varphi}_m$ is the $m$th component of the DFT applied to the mother polynomial function.
Let $\mbE\in \mathbb{C}^{N\times k^*}$ be the matrix whose $(m,j)$ entries are $\exp(-2\pi m j i /k^*)$, and let $\mbD\in \mathbb{C}^{N\times N}$ be the diagonal matrix whose $m$th entries is  $\hat{\varphi}_m$.
Then, we have 
\begin{equation}
    \hat{\bm\Psi} = \mbF \bm \Psi^\top =  \mbD \mbE.
\end{equation}
Moreover, note that the columns of $\mbE$ are $k^*$-periodic, so the matrix $\mbE$ is composed of $h$ blocks of size $k^*\times k^*$ stacked on top of each other.
Each of these blocks is the $k^*\times k^*$ DFT matrix $\sqrt{k^*}\mbF_{k^*\times k^*}$.
Letting $\mbJ$ be the block identity matrix of size $N\times k^*$, we can write 
\begin{equation}
    \hat{\bm\Psi} = \mbF \bm\Psi^\top = \sqrt{k^*}\mbD \mbJ \mbF_{k^*\times k^*}, 
\end{equation}

Now that this setup is complete, we can perform several substitutions to recover the $\mbH$ matrix above.
Consider 
\begin{equation}
    \begin{split}
        (\mbP-\mbI) &= (\bm\Psi^\top (\bm\Psi \bm\Psi^\top)^{-1} \bm\Psi-\mbI)
        \\
        &= (\bm\Psi^\top (\bm\Psi \mbF^*\mbF \bm\Psi^\top)^{-1} \bm\Psi-\mbF^*\mbF)
        \\
        &= (\bm\Psi^\top (k^*\mbF_{k^*\times k^*}^* [\mbJ^* |\mbD|^2 \mbJ] \mbF_{k^*\times k^*})^{-1} \bm\Psi-\mbF^*\mbF).
    \end{split}
\end{equation}
Let $\mbS = \mbJ^* |\mbD|^2 \mbJ \in \mathbb{R}^{k^*\times k^*}$.
The $m$th entry of the diagonal matrix $|\mbD|^2$ is $|\hat{\phi}_m|^2$.
The action of $\mbJ$ on either side is to sum up the ``blocks'', so $\mbS$ is a diagonal matrix whose $m$th entry is $\sum_{\ell=0}^{h-1} |\hat{\phi}_{m+\ell k^*}|^2$. 
Continuing, we have
\begin{equation}
    \begin{split}
        (\mbP-\mbI) &= \frac{1}{k^*}(\bm\Psi^\top\mbF^*_{k^*\times k^*} \mbS^{-1}\mbF_{k^*\times k^*}\bm\Psi-\mbF^*\mbF)
        \\
        &= \frac{1}{k^*}(\mbF^* \hat{\bm\Psi} \mbF^*_{k^*\times k^*}\mbS^{-1}\mbF_{k^*\times k^*}\hat{\bm\Psi}^*\mbF-\mbF^*\mbF)
        \\
        &= \frac{1}{k^*}\mbF^*\left([\sqrt{k^*}\mbD \mbJ \mbF_{k^*\times k^*}\mbF^*_{k^*\times k^*}] \mbS^{-1}[\sqrt{k^*}\mbD \mbJ \mbF_{k^*\times k^*}\mbF_{k^*\times k^*}^*]^*-\mbI \right)\mbF 
        \\
        &=\mbF^*\left((\mbD \mbJ)  \mbS^{-1}(\mbJ^*\mbD^*)-\mbI \right)\mbF. 
    \end{split}
\end{equation}
Define $\mbH = (\mbD \mbJ)  \mbS^{-1}(\mbJ^*\mbD^*)\in \mathbb{R}^{N\times N}$.
One can show, using the block structure of $\mbJ$, that the entries of $\mbH$ are of the form
\begin{equation}
    \mbH_{mn} = 
    \begin{cases}
        \frac{\hat{\varphi}_m \overline{\hat{\varphi}_n}}{\sum_{\ell=0}^{h-1} |\hat{\varphi}_{r+\ell k^*}|^2}, & m = n \;(\text{mod } k^*),
        \\
        0, & \text{otherwise}.
    \end{cases} 
\end{equation}
where $r = m \text{ mod } k^*$.
This computation recovers the form \eqref{eq: bias goal}.

\section{Validation Heuristic}\label{appendix: validation heuristic}

In this section, we review a validation heuristic to compute reference bandwidth and regularization parameters $(\epsilon^*,\lambda^*)$ which was utilized in \cite{song2025learning} for strong KRR applied to clean data.
Here, we modify it slightly to improve robustness in the case of noisy data.
We emphasize that the procedure described below is a heuristic which has been observed to provide reasonable results over a range of examples \cite{huang2025learning, song2025learning}.
While it often provides a good starting point, it should not be interpreted as an optimal approach that is appropriate for all problems. 

Suppose that training data $\mbu_i$ for $i=1,\dots,N$ of the form \eqref{eq: data} is given.
This data could be filtered, but is not required to be.
Let $L_*$  be the maximum pairwise $L_2$ distance between these data points.
Let
\begin{equation}
    \rho(\mbx,\mby;\eta) = \exp(-\|\mbx-\mby\|_2^2/(L_*^2 \eta)
\end{equation}
be a standard Gaussian RBF kernel with bandwidth $\eta$.
Further define
\begin{equation}
    S(\eta) = \frac{1}{N^2}\sum_{i,j=1}^N \rho(\mbu_i,\mbu_j;\eta).
\end{equation}
Following \cite{coifman2008graph,song2025learning}, this fact leads one to consider
\begin{equation}
    V(\eta) = 2\frac{\eta}{S(\eta)}\frac{\td S(\eta)}{\td \eta},
\end{equation}
and to consider the problem
\begin{equation}
    \eta^* = \arg\max_\eta V(\eta).
\end{equation}
For clean data, $V(\eta)$ is often uni-modal, and an appropriate choice of $\eta^*$ is unambiguous.
For noisy data, we numerically observe that $V(\eta)$ may admit multiple peaks with similar values.
In practice, if $\eta^*_r$ denotes the arguments at which the local maxima occur, we recommend selecting $\eta^* = \max_r \eta^*_r$.
The goal of this procedure is to avoid selecting an extremely small bandwidth, which could lead to poor generalization.

Once $\eta^*$ has been computed, we use it to define a reference bandwidth $\epsilon^*$ via the following formula
\begin{equation}
    \epsilon^* = 250 L_*^2\eta^*.
\end{equation}

To select the reference regularization parameter $\lambda^*$, let $k(\mbx,\mby;\epsilon) = \exp(-\|\mbx-\mby\|_2^2/\epsilon)$ be the Gaussian RBF kernel, and  let $\mbK_{\text{RBF}}(\epsilon)$ be the corresponding Gram matrix whose $(i,j)$ entry is $k(\mbu_i,\mbu_j;\epsilon)$.
We choose $\lambda^*$ to be the minimum eigenvalue of $\mbK_{\text{RBF}}(\epsilon)$.

\section{Validation Landscapes}\label{appendix: validation landscape}

\begin{figure}[htbp]
    \centering
    \begin{minipage}{0.49\textwidth}
        \centering
        \includegraphics[width=0.48\textwidth]{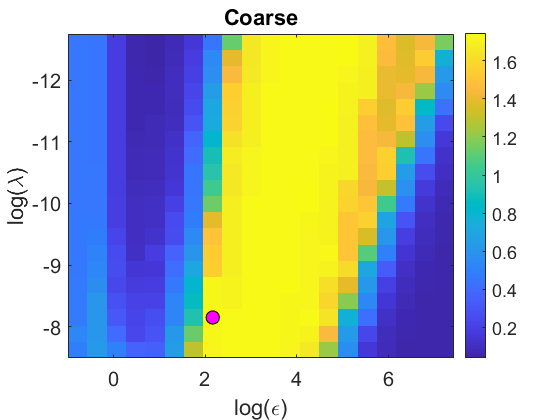}\hfill
        \includegraphics[width=0.48\textwidth]{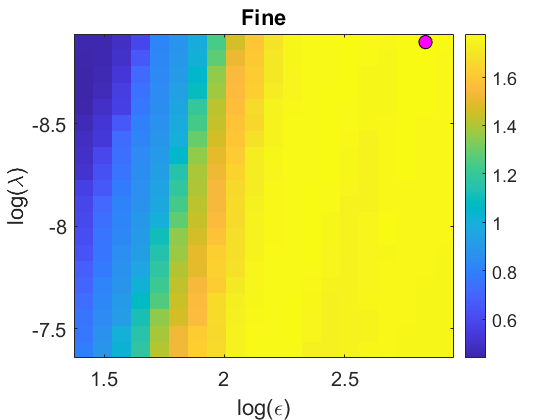}
        \vskip 0.5ex
        \textbf{\small (a) 1\% Noise}
    \end{minipage}\hfill
    \begin{minipage}{0.49\textwidth}
        \centering
        \includegraphics[width=0.48\textwidth]{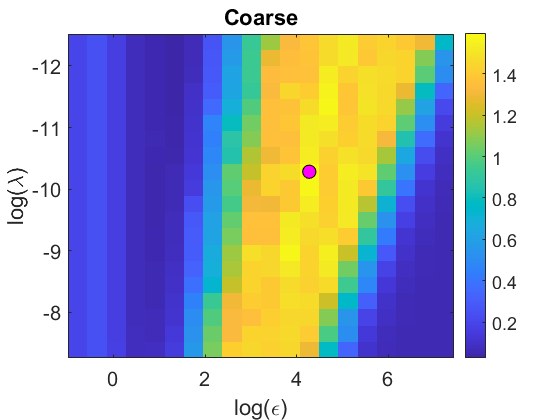}\hfill
        \includegraphics[width=0.48\textwidth]{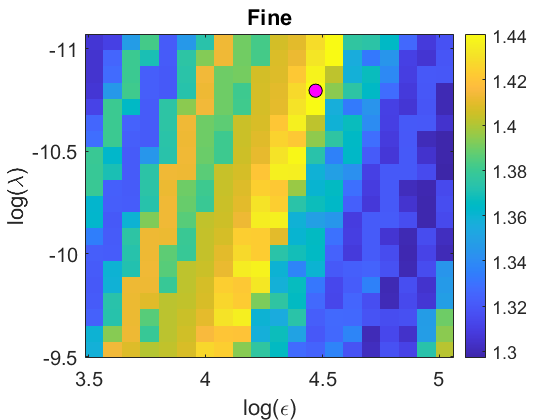}
        \vskip 0.5ex
        \textbf{\small (b) 5\% Noise}
    \end{minipage}
    
    \vskip 1ex 
    
    \begin{minipage}{0.49\textwidth}
        \centering
        \includegraphics[width=0.48\textwidth]{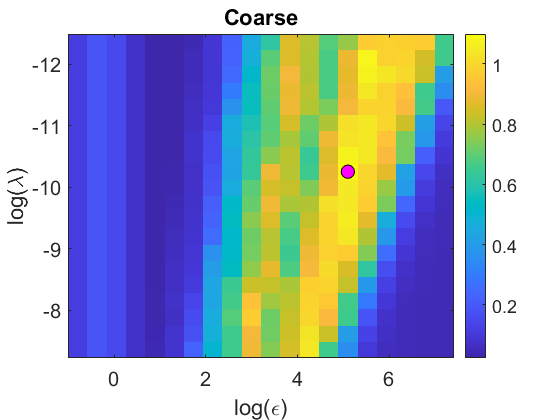}\hfill
        \includegraphics[width=0.48\textwidth]{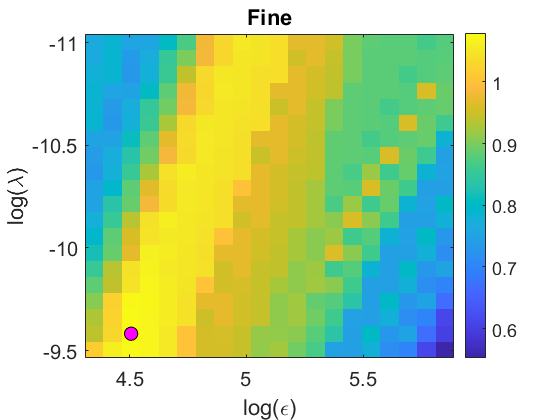}
        \vskip 0.5ex
        \textbf{\small (c) 10\% Noise}
    \end{minipage}\hfill
    \begin{minipage}{0.49\textwidth}
        \centering
        \includegraphics[width=0.48\textwidth]{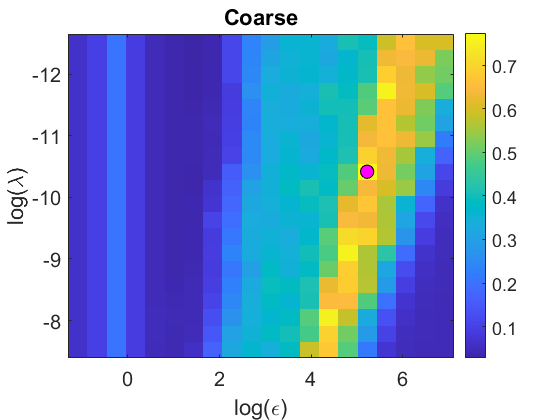}\hfill
        \includegraphics[width=0.48\textwidth]{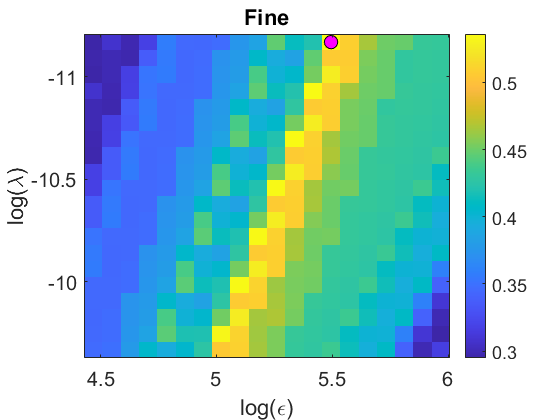}
        \vskip 0.5ex
        \textbf{\small (d) 20\% Noise}
    \end{minipage}

    \vskip 1ex
    \caption{Typical coarse (left sub-columns) and fine (right sub-columns) validation landscapes for the L63 system \eqref{eq: Lorenz 63 equations} at various noise intensities. Color denotes VPT. The pink dots denote the chosen parameter pair.}
    \label{fig: L63 val}
\end{figure}

\begin{figure}[htbp]
    \centering
    \begin{minipage}{0.49\textwidth}
        \centering
        \includegraphics[width=0.48\textwidth]{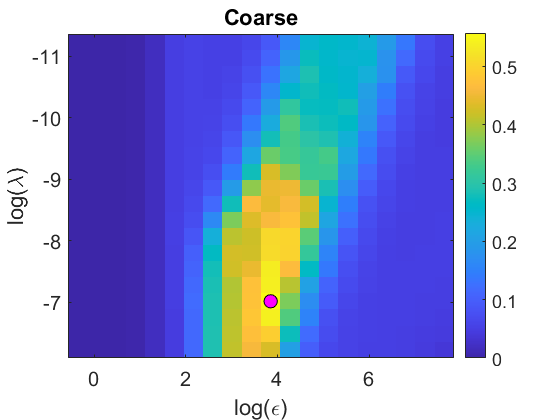}\hfill
        \includegraphics[width=0.48\textwidth]{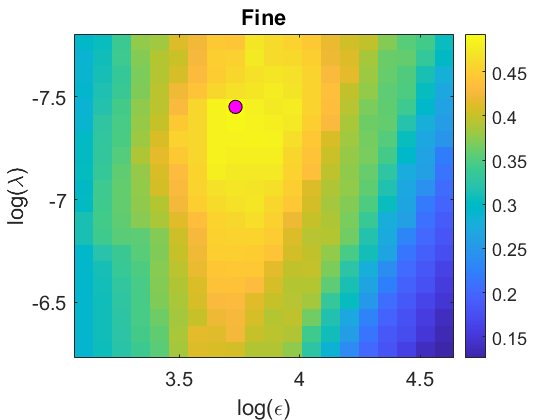}
        \vskip 0.5ex
        \textbf{\small (a) 1\% Noise}
    \end{minipage}\hfill
    \begin{minipage}{0.49\textwidth}
        \centering
        \includegraphics[width=0.48\textwidth]{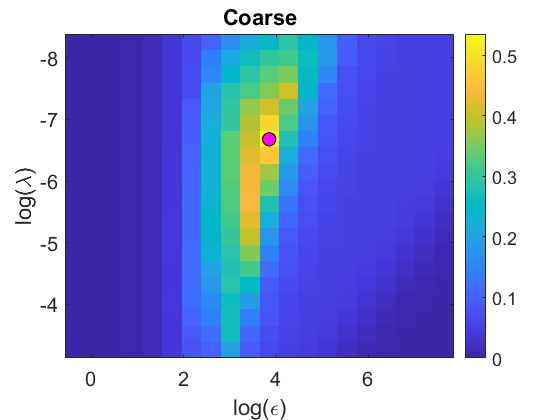}\hfill
        \includegraphics[width=0.48\textwidth]{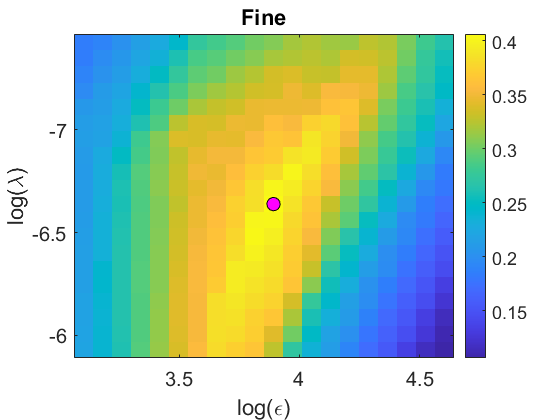}
        \vskip 0.5ex
        \textbf{\small (b) 5\% Noise}
    \end{minipage}
    
    \vskip 1ex 
    
    \begin{minipage}{0.49\textwidth}
        \centering
        \includegraphics[width=0.48\textwidth]{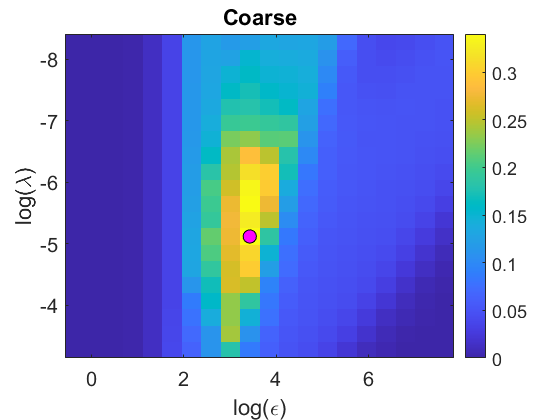}\hfill
        \includegraphics[width=0.48\textwidth]{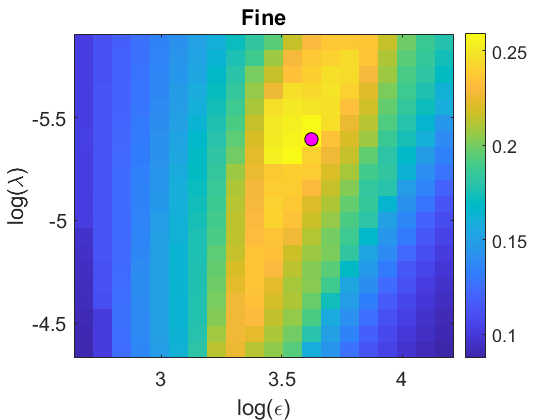}
        \vskip 0.5ex
        \textbf{\small (c) 10\% Noise}
    \end{minipage}\hfill
    \begin{minipage}{0.49\textwidth}
        \centering
        \includegraphics[width=0.48\textwidth]{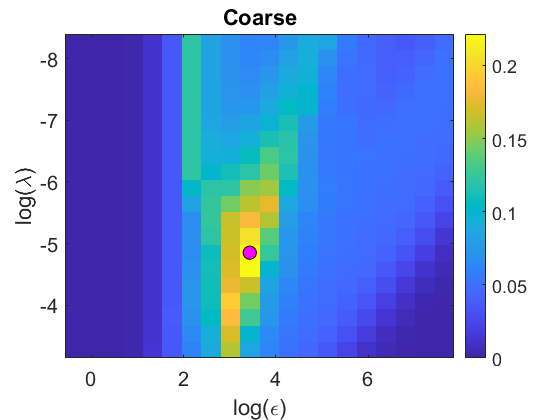}\hfill
        \includegraphics[width=0.48\textwidth]{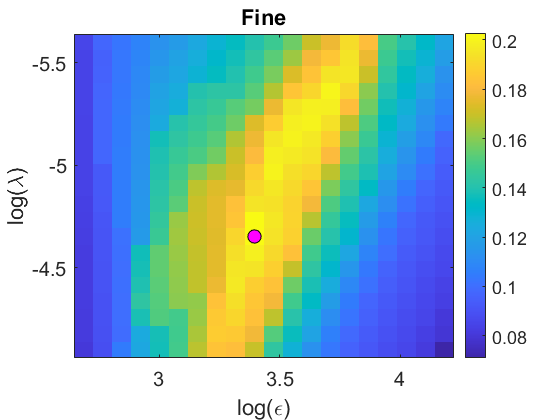}
        \vskip 0.5ex
        \textbf{\small (d) 20\% Noise}
    \end{minipage}

    \vskip 1ex
    \caption{Typical coarse (left sub-columns) and fine (right sub-columns) validation landscapes for the KS system \eqref{eq: KS system} at various noise intensities. Color denotes VPT. The pink dots denote the chosen parameter pair.}
    \label{fig: ks val}
\end{figure}

\begin{figure}[htbp]
    \centering
    \includegraphics[width=0.35\textwidth]{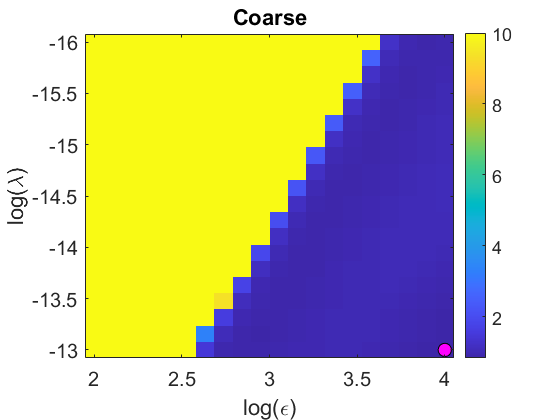}
    \includegraphics[width=0.35\textwidth]{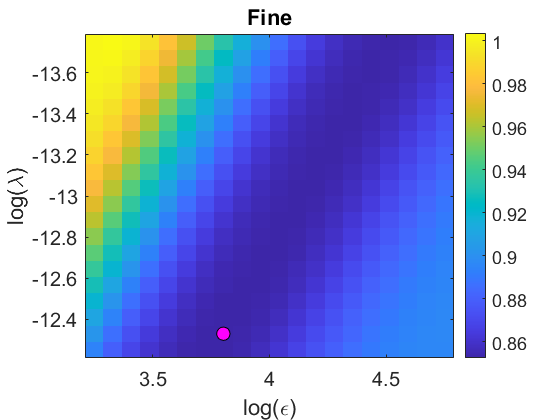}
    \caption{Typical coarse (left) and fine (right) validation landscapes for the Community Challenge fluid data with $r=25$ POD modes retained. Error values are thresholded at 10 to better visualize the landscape.}
    \label{fig: cc validation}
\end{figure}

In this section, we depict typical WKRR validation landscapes across noise levels for the L63 system \eqref{eq: Lorenz 63 equations} in Figure \ref{fig: L63 val}, for the KS system \eqref{eq: KS system} in Figure \ref{fig: ks val}, and for the experimental fluid data in Figure \ref{fig: cc validation}.

Note that in the case of the experimental fluid data, over-regularizing may improve quantitative performance in terms of the error metric at the cost of qualitative fidelity, i.e., over-regularizing often causes WKRR to rapidly converge to a mean flow that does not capture important qualitative features of the data.

\newpage
\section{KS Diffusion Maps Kernel Data}\label{appendix: KS DM data}

Here, we repeat the experimental procedure for the KS system described in \S\ref{subsec: KS section} with the DM kernel \eqref{eq: our discrete DM kernel} instead of the Gaussian kernel \eqref{eq: Gaussian kernel}.
The results are shown in Table \ref{table: appendix KS}, and demonstrate the superior performance of WKRR over the strong formulation across all noise levels.

\begin{table}[htpb]
    \centering
    \small
    \setlength{\tabcolsep}{5pt}
    \begin{tabular}{l cccc cccc}
        \toprule
        & \multicolumn{4}{c}{\textbf{Mean VPT}} & \multicolumn{4}{c}{\textbf{Test Function Parameters} $(p, L, h)$} 
        \\
        \cmidrule(lr){2-5} \cmidrule(lr){6-9}
        \textbf{Method} & 1\% & 5\% & 10\% & 20\% & 1\% & 5\% & 10\% & 20\% 
        \\
        \midrule
        Strong DM (n/a)       
        & $0.74 \pm 0.11$
        & $0.37 \pm 0.05$
        & $0.25 \pm 0.03$
        & $0.15 \pm 0.03$
        & --- & --- & --- & --- 
        \\
        Strong DM (Wavelet)     
        & $0.87 \pm 0.13$
        & $0.62 \pm 0.08$
        & $0.51 \pm 0.07$
        & $0.40 \pm 0.05$
        & --- & --- & --- & --- 
        \\
        Strong DM (Poly)        
        & $0.85 \pm 0.12$
        & $0.63 \pm 0.08$
        & $0.51 \pm 0.06$
        & $0.39 \pm 0.05$
        & ($7, 5.5, 1.1$) 
        & ($5, 8, 1.6$) 
        & ($5, 9.5, 1.9$) 
        & ($6, 11.5, 2.3$)
        \\
        Weak DM (n/a)           
        & $0.89 \pm 0.17$
        & $0.67 \pm 0.09$
        & $0.55 \pm 0.07$
        & $0.44 \pm 0.06$
        & ($7, 5.5, 1.1$) 
        & ($5, 8, 1.6$) 
        & ($5, 9.5, 1.9$) 
        & ($6, 11.5, 2.3$)
        \\
        \textbf{Weak DM (Poly)} 
        & $\bm{0.95 \pm 0.13}$
        & $\bm{0.67 \pm 0.09}$
        & $\bm{0.55 \pm 0.07}$
        & $\bm{0.44 \pm 0.06}$
        & ($7, 5.5, 1.1$) 
        & ($5, 8, 1.6$) 
        & ($5, 9.5, 1.9$) 
        & ($6, 11.5, 2.3$) 
        \\
        \bottomrule
    \end{tabular}
    \vspace{.3cm}
    \caption{VPT statistics for the KS system \eqref{eq: KS system} under various noise intensities using the DM kernel.
    ``Strong'' and ``Weak'' denote classical KRR and proposed WKRR frameworks, and parentheses indicate filtering applied to the training data, where (n/a) denotes unfiltered data. 
    Validation data is filtered as the training data for strong formulations, while polynomials are used for the weak formulations.}
    \label{table: appendix KS}
\end{table}

\bibliographystyle{plain}  
\bibliography{arxiv}

@article{kassam2005fourth,
  title={Fourth-order time-stepping for stiff {PDEs}},
  author={Kassam, Aly-Khan and Trefethen, Lloyd N},
  journal={SIAM Journal on Scientific Computing},
  volume={26},
  number={4},
  pages={1214--1233},
  year={2005},
  publisher={SIAM}
}

@article{edson2019lyapunov,
  title={{Lyapunov} exponents of the {Kuramoto}--{Sivashinsky} {PDE}},
  author={Edson, Russell A and Bunder, Judith E and Mattner, Trent W and Roberts, Anthony J},
  journal={The ANZIAM Journal},
  volume={61},
  number={3},
  pages={270--285},
  year={2019},
  publisher={Cambridge University Press}
}

@article{kuznetsov2020lorenz,
  title={The {Lorenz} system: hidden boundary of practical stability and the {Lyapunov} dimension},
  author={Kuznetsov, NV and Mokaev, TN and Kuznetsova, OA and Kudryashova, EV},
  journal={Nonlinear Dyn},
  volume={102},
  pages={713--732},
  year={2020}
}

@article{brugnago2020predicting,
  title={Predicting regime changes and durations in {Lorenz}'s atmospheric convection model},
  author={Brugnago, Eduardo L and Gallas, Jason AC and Beims, Marcus W},
  journal={Chaos: An Interdisciplinary Journal of Nonlinear Science},
  volume={30},
  number={10},
  year={2020},
  publisher={AIP Publishing}
}

@incollection{lorenz2017deterministic,
  title={Deterministic nonperiodic flow 1},
  author={Lorenz, Edward N},
  booktitle={Universality in Chaos, 2nd edition},
  pages={367--378},
  year={2017},
  publisher={Routledge}
}

@article{coifman2008graph,
  title={Graph {Laplacian} tomography from unknown random projections},
  author={Coifman, Ronald R and Shkolnisky, Yoel and Sigworth, Fred J and Singer, Amit},
  journal={IEEE Transactions on Image Processing},
  volume={17},
  number={10},
  pages={1891--1899},
  year={2008},
  publisher={IEEE}
}

@article{savitzky1964smoothing,
  title={Smoothing and differentiation of data by simplified least squares procedures.},
  author={Savitzky, Abraham and Golay, Marcel JE},
  journal={Analytical chemistry},
  volume={36},
  number={8},
  pages={1627--1639},
  year={1964},
  publisher={ACS Publications}
}

@book{maybeck1982stochastic,
  title={Stochastic models, estimation, and control},
  author={Maybeck, Peter S},
  volume={3},
  year={1982},
  publisher={Academic press}
}

@article{kalman1960new,
  title={A new approach to linear filtering and prediction problems},
  author={Kalman, Rudolph Emil},
  journal={Transactions of the ASME–Journal of Basic Engineering},
  year={1960}
}

@book{box2015time,
  title={Time series analysis: forecasting and control},
  author={Box, George EP and Jenkins, Gwilym M and Reinsel, Gregory C and Ljung, Greta M},
  year={2015},
  publisher={John Wiley \& Sons}
}

@article{ljung1978measure,
  title={On a measure of lack of fit in time series models},
  author={Ljung, Greta M and Box, George EP},
  journal={Biometrika},
  volume={65},
  number={2},
  pages={297--303},
  year={1978},
  publisher={Oxford University Press}
}

@article{blu1999approximation,
  title={Approximation error for quasi-interpolators and (multi-) wavelet expansions},
  author={Blu, Thierry and Unser, Michael},
  journal={Applied and Computational Harmonic Analysis},
  volume={6},
  number={2},
  pages={219--251},
  year={1999},
  publisher={Elsevier}
}

@article{aldroubi1994sampling,
  title={Sampling procedures in function spaces and asymptotic equivalence with {Shannon}'s sampling theory},
  author={Aldroubi, Akram and Unser, Michael and Aldroubi, Akram},
  journal={Numerical functional analysis and optimization},
  volume={15},
  number={1-2},
  pages={1--21},
  year={1994},
  publisher={Taylor \& Francis}
}

@article{unser2002general,
  title={A general sampling theory for nonideal acquisition devices},
  author={Unser, Michael and Aldroubi, Akram},
  journal={IEEE Transactions on Signal Processing},
  volume={42},
  number={11},
  pages={2915--2925},
  year={2002},
  publisher={IEEE}
}

@article{wyner1998introduction,
  title={Introduction to ‘Communication in the presence of noise’ by {CE Shannon}},
  author={Wyner, Aaron D and Shamai, Shlomo},
  journal={Proc. IEEE},
  volume={86},
  number={2},
  pages={442--446},
  year={1998}
}

@article{shannon1949communication,
  title={Communication in the presence of noise},
  author={Shannon, Claude E},
  journal={Proceedings of the IRE},
  volume={37},
  number={1},
  pages={10--21},
  year={1949},
  publisher={IEEE}
}

@article{unser2002polynomial,
  title={Polynomial spline signal approximations: filter design and asymptotic equivalence with {Shannon}'s sampling theorem},
  author={Unser, Michael and Aldroubi, Akram and Eden, Murray},
  journal={IEEE Transactions on Information Theory},
  volume={38},
  number={1},
  pages={95--103},
  year={2002},
  publisher={IEEE}
}

@article{unser2002generalized,
  title={A generalized sampling theory without band-limiting constraints},
  author={Unser, Michael and Zerubia, Josiane},
  journal={IEEE transactions on circuits and systems II: analog and digital signal processing},
  volume={45},
  number={8},
  pages={959--969},
  year={2002},
  publisher={IEEE}
}

@article{unser2002sampling,
  title={Sampling-50 years after {Shannon}},
  author={Unser, Michael},
  journal={Proceedings of the IEEE},
  volume={88},
  number={4},
  pages={569--587},
  year={2002},
  publisher={IEEE}
}

@article{messenger2025asymptotic,
  title={Asymptotic consistency of the {WSINDy} algorithm in the limit of continuum data},
  author={Messenger, Daniel A and Bortz, David M},
  journal={IMA Journal of Numerical Analysis},
  volume={45},
  number={6},
  pages={3264--3312},
  year={2025},
  publisher={Oxford University Press}
}

@incollection{bortz2024weak,
  title={Weak form-based data-driven modeling: computationally efficient and noise robust equation learning and parameter inference},
  author={Bortz, David M and Messenger, Daniel A and Tran, April},
  booktitle={Handbook of Numerical Analysis},
  volume={25},
  pages={53--82},
  year={2024},
  publisher={Elsevier}
}

@article{messenger2024weak,
  title={The weak form is stronger than you think},
  author={Messenger, Daniel A and Tran, April and Dukic, Vanja and Bortz, David M},
  journal={arXiv preprint arXiv:2409.06751},
  year={2024}
}

@article{yang2021inference,
  title={Inference of dynamic systems from noisy and sparse data via manifold-constrained Gaussian processes},
  author={Yang, Shihao and Wong, Samuel WK and Kou, SC},
  journal={Proceedings of the National Academy of Sciences},
  volume={118},
  number={15},
  pages={e2020397118},
  year={2021},
  publisher={National Academy of Sciences}
}

@inproceedings{yan2009gaussian,
  title={Gaussian process for long-term time-series forecasting},
  author={Yan, Weizhong and Qiu, Hai and Xue, Ya},
  booktitle={2009 international joint conference on neural networks},
  pages={3420--3427},
  year={2009},
  organization={IEEE}
}

@article{girard2002gaussian,
  title={Gaussian process priors with uncertain inputs application to multiple-step ahead time series forecasting},
  author={Girard, Agathe and Rasmussen, Carl and Candela, Joaquin Q and Murray-Smith, Roderick},
  journal={Advances in neural information processing systems},
  volume={15},
  year={2002}
}

@article{cheng2023machine,
  title={Machine learning with data assimilation and uncertainty quantification for dynamical systems: a review},
  author={Cheng, Sibo and Quilodr{\'a}n-Casas, C{\'e}sar and Ouala, Said and Farchi, Alban and Liu, Che and Tandeo, Pierre and Fablet, Ronan and Lucor, Didier and Iooss, Bertrand and Brajard, Julien and others},
  journal={IEEE/CAA Journal of Automatica Sinica},
  volume={10},
  number={6},
  pages={1361--1387},
  year={2023},
  publisher={IEEE}
}

@article{gottwald2021supervised,
  title={Supervised learning from noisy observations: Combining machine-learning techniques with data assimilation},
  author={Gottwald, Georg A and Reich, Sebastian},
  journal={Physica D: Nonlinear Phenomena},
  volume={423},
  pages={132911},
  year={2021},
  publisher={Elsevier}
}

@incollection{vovk2013kernel,
  title={Kernel ridge regression},
  author={Vovk, Vladimir},
  booktitle={Empirical inference: Festschrift in honor of vladimir n. vapnik},
  pages={105--116},
  year={2013},
  publisher={Springer}
}

@article{ahmed2022kernel,
  title={Kernel ridge regression hybrid method for wheat yield prediction with satellite-derived predictors},
  author={Ahmed, AA Masrur and Sharma, Ekta and Jui, S Janifer Jabin and Deo, Ravinesh C and Nguyen-Huy, Thong and Ali, Mumtaz},
  journal={Remote Sensing},
  volume={14},
  number={5},
  pages={1136},
  year={2022},
  publisher={MDPI}
}

@article{ali2020complete,
  title={Complete ensemble empirical mode decomposition hybridized with random forest and kernel ridge regression model for monthly rainfall forecasts},
  author={Ali, Mumtaz and Prasad, Ramendra and Xiang, Yong and Yaseen, Zaher Mundher},
  journal={Journal of Hydrology},
  volume={584},
  pages={124647},
  year={2020},
  publisher={Elsevier}
}

@article{exterkate2016nonlinear,
  title={Nonlinear forecasting with many predictors using kernel ridge regression},
  author={Exterkate, Peter and Groenen, Patrick JF and Heij, Christiaan and van Dijk, Dick},
  journal={International Journal of Forecasting},
  volume={32},
  number={3},
  pages={736--753},
  year={2016},
  publisher={Elsevier}
}

@article{floryan2022data,
  title={Data-driven discovery of intrinsic dynamics},
  author={Floryan, Daniel and Graham, Michael D},
  journal={Nature Machine Intelligence},
  volume={4},
  number={12},
  pages={1113--1120},
  year={2022},
  publisher={Nature Publishing Group UK London}
}

@article{gauthier2021next,
  title={Next generation reservoir computing},
  author={Gauthier, Daniel J and Bollt, Erik and Griffith, Aaron and Barbosa, Wendson AS},
  journal={Nature communications},
  volume={12},
  number={1},
  pages={5564},
  year={2021},
  publisher={Nature Publishing Group UK London}
}

@article{tanaka2019recent,
  title={Recent advances in physical reservoir computing: A review},
  author={Tanaka, Gouhei and Yamane, Toshiyuki and H{\'e}roux, Jean Benoit and Nakane, Ryosho and Kanazawa, Naoki and Takeda, Seiji and Numata, Hidetoshi and Nakano, Daiju and Hirose, Akira},
  journal={Neural Networks},
  volume={115},
  pages={100--123},
  year={2019},
  publisher={Elsevier}
}

@book{nakajima2021reservoir,
  title={Reservoir computing},
  author={Nakajima, Kohei and Fischer, Ingo},
  year={2021},
  publisher={Springer}
}

@article{yan2024emerging,
  title={Emerging opportunities and challenges for the future of reservoir computing},
  author={Yan, Min and Huang, Can and Bienstman, Peter and Tino, Peter and Lin, Wei and Sun, Jie},
  journal={Nature Communications},
  volume={15},
  number={1},
  pages={2056},
  year={2024},
  publisher={Nature Publishing Group UK London}
}

@article{lindemann2021survey,
  title={A survey on long short-term memory networks for time series prediction},
  author={Lindemann, Benjamin and M{\"u}ller, Timo and Vietz, Hannes and Jazdi, Nasser and Weyrich, Michael},
  journal={Procedia Cirp},
  volume={99},
  pages={650--655},
  year={2021},
  publisher={Elsevier}
}

@article{yu2019review,
  title={A review of recurrent neural networks: {LSTM} cells and network architectures},
  author={Yu, Yong and Si, Xiaosheng and Hu, Changhua and Zhang, Jianxun},
  journal={Neural computation},
  volume={31},
  number={7},
  pages={1235--1270},
  year={2019},
  publisher={MIT Press One Rogers Street, Cambridge, MA 02142-1209, USA journals-info~…}
}

@inproceedings{zhao2025accelerating,
  title={Accelerating neural {ODEs}: a variational formulation-based approach},
  author={Zhao, Hongjue and Wang, Yuchen and Qi, Hairong and Huang, Zijie and Zhao, Han and Sha, Lui and Shao, Huajie},
  booktitle={The Thirteenth International Conference on Learning Representations},
  year={2025}
}

@article{goyal2022neural,
  title={Neural {ODEs} with irregular and noisy data},
  author={Goyal, Pawan and Benner, Peter},
  journal={arXiv preprint arXiv:2205.09479},
  year={2022}
}

@article{oh2025comprehensive,
  title={Comprehensive review of neural differential equations for time series analysis},
  author={Oh, YongKyung and Kam, Seungsu and Lee, Jonghun and Lim, Dong-Young and Kim, Sungil and Bui, Alex},
  journal={arXiv preprint arXiv:2502.09885},
  year={2025}
}

@article{chen2018neural,
  title={Neural ordinary differential equations},
  author={Chen, Ricky TQ and Rubanova, Yulia and Bettencourt, Jesse and Duvenaud, David K},
  journal={Advances in neural information processing systems},
  volume={31},
  year={2018}
}

@article{zhang2019convergence,
  title={On the convergence of the {SINDy} algorithm},
  author={Zhang, Linan and Schaeffer, Hayden},
  journal={Multiscale Modeling \& Simulation},
  volume={17},
  number={3},
  pages={948--972},
  year={2019},
  publisher={SIAM}
}

@article{kaheman2020sindy,
  title={{SINDy-PI}: a robust algorithm for parallel implicit sparse identification of nonlinear dynamics},
  author={Kaheman, Kadierdan and Kutz, J Nathan and Brunton, Steven L},
  journal={Proceedings. Mathematical, physical, and engineering sciences},
  volume={476},
  number={2242},
  pages={20200279},
  year={2020}
}

@article{brunton2016discovering,
  title={Discovering governing equations from data by sparse identification of nonlinear dynamical systems},
  author={Brunton, Steven L and Proctor, Joshua L and Kutz, J Nathan},
  journal={Proceedings of the national academy of sciences},
  volume={113},
  number={15},
  pages={3932--3937},
  year={2016},
  publisher={National Academy of Sciences}
}

@article{brunton2016sparse,
  title={Sparse identification of nonlinear dynamics with control ({SINDYc})},
  author={Brunton, Steven L and Proctor, Joshua L and Kutz, J Nathan},
  journal={IFAC-PapersOnLine},
  volume={49},
  number={18},
  pages={710--715},
  year={2016},
  publisher={Elsevier}
}

@article{mezic2022numerical,
  title={On numerical approximations of the {Koopman} operator},
  author={Mezi{\'c}, Igor},
  journal={Mathematics},
  volume={10},
  number={7},
  pages={1180},
  year={2022},
  publisher={MDPI}
}

@article{colbrook2023residual,
  title={Residual dynamic mode decomposition: robust and verified {Koopmanism}},
  author={Colbrook, Matthew J and Ayton, Lorna J and Sz{\H{o}}ke, M{\'a}t{\'e}},
  journal={Journal of Fluid Mechanics},
  volume={955},
  pages={A21},
  year={2023},
  publisher={Cambridge University Press}
}

@article{li2017extended,
  title={Extended dynamic mode decomposition with dictionary learning: A data-driven adaptive spectral decomposition of the {Koopman} operator},
  author={Li, Qianxiao and Dietrich, Felix and Bollt, Erik M and Kevrekidis, Ioannis G},
  journal={Chaos: An Interdisciplinary Journal of Nonlinear Science},
  volume={27},
  number={10},
  year={2017},
  publisher={AIP Publishing}
}

@article{colbrook2023mpedmd,
  title={The mpEDMD algorithm for data-driven computations of measure-preserving dynamical systems},
  author={Colbrook, Matthew J},
  journal={SIAM Journal on Numerical Analysis},
  volume={61},
  number={3},
  pages={1585--1608},
  year={2023},
  publisher={SIAM}
}

@book{kutz2016dynamic,
  title={Dynamic mode decomposition: data-driven modeling of complex systems},
  author={Kutz, J Nathan and Brunton, Steven L and Brunton, Bingni W and Proctor, Joshua L},
  year={2016},
  publisher={SIAM}
}

@article{north2023review,
  title={A review of data-driven discovery for dynamic systems},
  author={North, Joshua S and Wikle, Christopher K and Schliep, Erin M},
  journal={International Statistical Review},
  volume={91},
  number={3},
  pages={464--492},
  year={2023},
  publisher={Wiley Online Library}
}

@book{brunton2022data,
  title={Data-driven science and engineering: Machine learning, dynamical systems, and control},
  author={Brunton, Steven L and Kutz, J Nathan},
  year={2022},
  publisher={Cambridge University Press}
}

@article{wang2005gaussian,
  title={Gaussian process dynamical models},
  author={Wang, Jack and Hertzmann, Aaron and Fleet, David J},
  journal={Advances in neural information processing systems},
  volume={18},
  year={2005}
}

@article{williams2015data,
  title={A data--driven approximation of the {Koopman} operator: Extending dynamic mode decomposition},
  author={Williams, Matthew O and Kevrekidis, Ioannis G and Rowley, Clarence W},
  journal={Journal of Nonlinear Science},
  volume={25},
  number={6},
  pages={1307--1346},
  year={2015},
  publisher={Springer}
}

@article{ghadami2022data,
  title={Data-driven prediction in dynamical systems: recent developments},
  author={Ghadami, Amin and Epureanu, Bogdan I},
  journal={Philosophical transactions. Series A, Mathematical, physical, and engineering sciences},
  volume={380},
  number={2229},
  pages={20210213},
  year={2022}
}

@article{xu2016big,
  title={Big data driven mobile traffic understanding and forecasting: A time series approach},
  author={Xu, Fengli and Lin, Yuyun and Huang, Jiaxin and Wu, Di and Shi, Hongzhi and Song, Jeungeun and Li, Yong},
  journal={IEEE transactions on services computing},
  volume={9},
  number={5},
  pages={796--805},
  year={2016},
  publisher={IEEE}
}

@article{antoniou2013dynamic,
  title={Dynamic data-driven local traffic state estimation and prediction},
  author={Antoniou, Constantinos and Koutsopoulos, Haris N and Yannis, George},
  journal={Transportation Research Part C: Emerging Technologies},
  volume={34},
  pages={89--107},
  year={2013},
  publisher={Elsevier}
}

@article{avila2020data,
  title={Data-driven analysis and forecasting of highway traffic dynamics},
  author={Avila, Allan M and Mezi{\'c}, Igor},
  journal={Nature communications},
  volume={11},
  number={1},
  pages={2090},
  year={2020},
  publisher={Nature Publishing Group UK London}
}

@article{waheed2026data,
  title={Data-driven neural modeling and chaos control in fractional-order financial dynamical systems},
  author={Waheed, Shamaila and Qayyum, Mubashir and Khan, Omar and Chambashi, Gilbert},
  journal={AIP Advances},
  volume={16},
  number={1},
  year={2026},
  publisher={AIP Publishing}
}

@inproceedings{castillo1995intelligent,
  title={An intelligent system for financial time series prediction combining dynamical systems theory, fractal theory, and statistical methods},
  author={Castillo, Oscar and Melin, Patricia},
  booktitle={Proceedings of 1995 Conference on Computational Intelligence for Financial Engineering (CIFEr)},
  pages={151--155},
  year={1995},
  organization={IEEE}
}

@article{agostini2020exploration,
  title={Exploration and prediction of fluid dynamical systems using auto-encoder technology},
  author={Agostini, Lionel},
  journal={Physics of Fluids},
  volume={32},
  number={6},
  year={2020},
  publisher={AIP Publishing}
}

@article{erge2022combining,
  title={Combining physics-based and data-driven modeling in well construction: Hybrid fluid dynamics modeling},
  author={Erge, Oney and Van Oort, Eric},
  journal={Journal of Natural Gas Science and Engineering},
  volume={97},
  pages={104348},
  year={2022},
  publisher={Elsevier}
}

@inproceedings{long2018hybridnet,
  title={Hybridnet: integrating model-based and data-driven learning to predict evolution of dynamical systems},
  author={Long, Yun and She, Xueyuan and Mukhopadhyay, Saibal},
  booktitle={Conference on robot learning},
  pages={551--560},
  year={2018},
  organization={PMLR}
}

@article{gilpin2020learning,
  title={Learning dynamics from large biological data sets: machine learning meets systems biology},
  author={Gilpin, William and Huang, Yitong and Forger, Daniel B},
  journal={Current Opinion in Systems Biology},
  volume={22},
  pages={1--7},
  year={2020},
  publisher={Elsevier}
}

@article{xing2022reconstructing,
  title={Reconstructing data-driven governing equations for cell phenotypic transitions: integration of data science and systems biology},
  author={Xing, Jianhua},
  journal={Physical Biology},
  volume={19},
  number={6},
  pages={061001},
  year={2022},
  publisher={IOP Publishing}
}

@article{prokop2025data,
  title={Data-driven discovery of dynamical models in biology},
  author={Prokop, Bartosz and Gelens, Lendert},
  journal={arXiv preprint arXiv:2509.06735},
  year={2025}
}

@article{luo2011ecological,
  title={Ecological forecasting and data assimilation in a data-rich era},
  author={Luo, Yiqi and Ogle, Kiona and Tucker, Colin and Fei, Shenfeng and Gao, Chao and LaDeau, Shannon and Clark, James S and Schimel, David S},
  journal={Ecological Applications},
  volume={21},
  number={5},
  pages={1429--1442},
  year={2011},
  publisher={Wiley Online Library}
}

@article{ye2015equation,
  title={Equation-free mechanistic ecosystem forecasting using empirical dynamic modeling},
  author={Ye, Hao and Beamish, Richard J and Glaser, Sarah M and Grant, Sue CH and Hsieh, Chih-hao and Richards, Laura J and Schnute, Jon T and Sugihara, George},
  journal={Proceedings of the National Academy of Sciences},
  volume={112},
  number={13},
  pages={E1569--E1576},
  year={2015},
  publisher={National Academy of Sciences}
}

@phdthesis{gimenez2024theoretical,
  title={Theoretical and data-driven models in Ecology},
  author={Gim{\'e}nez-Romero, {\`A}lex},
  year={2024},
  school={University of the Balearic Islands (UIB); Institute for Cross-Disciplinary Physics and Complex Systems}
}

@article{song2014application,
  title={Application of dynamic data driven application system in environmental science},
  author={Song, Jingwei and Xiang, Bo and Wang, Xinyuan and Wu, Li and Chang, Chun},
  journal={Environmental Reviews},
  volume={22},
  number={3},
  pages={287--297},
  year={2014},
  publisher={NRC Research Press}
}

@article{hussain2018dynamic,
  title={A dynamic neural network architecture with immunology inspired optimization for weather data forecasting},
  author={Hussain, Abir Jaafar and Liatsis, Panos and Khalaf, Mohammed and Tawfik, Hissam and Al-Asker, Haya},
  journal={Big data research},
  volume={14},
  pages={81--92},
  year={2018},
  publisher={Elsevier}
}

@inproceedings{nino2021data,
  title={Data-Driven Methods for Weather Forecast},
  author={Nino-Ruiz, Elias David and Acevedo Garc{\'\i}a, Felipe J},
  booktitle={International Conference on Computational Science},
  pages={326--336},
  year={2021},
  organization={Springer}
}

@article{christensen2019reliable,
  title={From reliable weather forecasts to skilful climate response: A dynamical systems approach},
  author={Christensen, Hannah M and Berner, Judith},
  journal={Quarterly Journal of the Royal Meteorological Society},
  volume={145},
  number={720},
  pages={1052--1069},
  year={2019},
  publisher={Wiley Online Library}
}

@article{song2025learning,
  title={Learning solution operator of dynamical systems with diffusion maps kernel ridge regression},
  author={Song, Jiwoo and Huang, Daning and Harlim, John},
  journal={arXiv preprint arXiv:2512.17203},
  year={2025}
}

@article{messenger2021weak2,
  title={Weak {SINDy}: Galerkin-based data-driven model selection},
  author={Messenger, Daniel A and Bortz, David M},
  journal={Multiscale Modeling \& Simulation},
  volume={19},
  number={3},
  pages={1474--1497},
  year={2021},
  publisher={SIAM}
}

@article{li2025weak,
  title={A Weak Penalty Neural {ODE} for Learning Chaotic Dynamics from Noisy Time Series},
  author={Li, Xuyang and Harlim, John and Chakraborty, Dibyajyoti and Maulik, Romit},
  journal={arXiv preprint arXiv:2511.06609},
  year={2025}
}

@article{messenger2021weak,
  title={Weak {SINDy} for partial differential equations},
  author={Messenger, Daniel A and Bortz, David M},
  journal={Journal of Computational Physics},
  volume={443},
  pages={110525},
  year={2021},
  publisher={Elsevier}
}

@article{hochreiter1997long,
  title={Long short-term memory},
  author={Hochreiter, Sepp and Schmidhuber, J{\"u}rgen},
  journal={Neural computation},
  volume={9},
  number={8},
  pages={1735--1780},
  year={1997},
  publisher={MIT press}
}

@article{harlim2026diffusion,
  title={Diffusion maps kernel ridge regression},
  author={Harlim, John and Huang, Daning and Song, Jiwoo and Townsend, Alex},
  journal={arXiv preprint, in preparation},
  year={2026}
}

@article{schmidt2026data,
  title={Data-Driven Reduced-Complexity Modeling of Fluid Flows: A Community Challenge},
  author={Schmidt, Oliver T and Towne, Aaron and Lozano-Duran, Adrian and Dawson, Scott and Vinuesa, Ricardo},
  journal={arXiv preprint arXiv:2601.06183},
  year={2026}
}

@article{gottwald2021combining,
  title={Combining machine learning and data assimilation to forecast dynamical systems from noisy partial observations},
  author={Gottwald, Georg A and Reich, Sebastian},
  journal={Chaos: An Interdisciplinary Journal of Nonlinear Science},
  volume={31},
  number={10},
  year={2021},
  publisher={AIP Publishing}
}

@inproceedings{huang2025learning,
  title={Learning vector fields of differential equations on manifolds with geometrically constrained operator-valued kernels},
  author={Huang, Daning and He, Hanyang and Harlim, John and Li, Yan},
  booktitle={The Thirteenth International Conference on Learning Representations},
  year={2025}
}

@article{cl:2006,
  title={Diffusion maps},
  author={Coifman, Ronald R and Lafon, St{\'e}phane},
  journal={Applied and computational harmonic analysis},
  volume={21},
  number={1},
  pages={5--30},
  year={2006},
  publisher={Elsevier}
}

@article{vlachas2018data,
  title={Data-driven forecasting of high-dimensional chaotic systems with long short-term memory networks},
  author={Vlachas, Pantelis R and Byeon, Wonmin and Wan, Zhong Y and Sapsis, Themistoklis P and Koumoutsakos, Petros},
  journal={Proceedings of the Royal Society A: Mathematical, Physical and Engineering Sciences},
  volume={474},
  number={2213},
  pages={20170844},
  year={2018},
  publisher={The Royal Society Publishing}
}

@article{Yu2024gkbf,
  title = {Learning Networked Dynamical System Models with Weak Form and Graph Neural Networks},
  author = {Yu, Yin and Huang, Daning and Park, Seho and Pangborn, Herschel},
  year = {2026},
  doi = {10.48550/arXiv.2407.16779},
  journal = {Journal of Guidance Control and Dynamics}
}

\end{document}